\documentclass{bmvc2k}

\usepackage{times}
\usepackage{epsfig}
\usepackage{graphicx}
\usepackage{amsmath}
\usepackage{amssymb}
\usepackage{caption}
\usepackage[title]{appendix}
\usepackage{booktabs} % For formal tables
\usepackage{multirow}
\usepackage{bm}
\usepackage{soul}
\title{Boosting up Scene Text Detectors with Guided CNN}
\addauthor{Xiaoyu Yue}{yuexiaoyu@sensetime.com}{1}
\addauthor{Zhanghui Kuang}{kuangzhanghui@sensetime.com}{1}
\addauthor{Zhaoyang Zhang}{zhangzhaoyang@sensetime.com}{1}
\addauthor{Zhenfang Chen}{zfchen@cs.hku.hk}{2}
\addauthor{Pan He}{pan.he@ufl.edu}{3}
\addauthor{Yu Qiao}{yu.qiao@siat.ac.cn}{4}
\addauthor{Wei Zhang}{wayne.zhang@sensetime.com}{1}

\addinstitution{
 SenseTime Group Limited
}
\addinstitution{
 The University of Hong Kong
}
\addinstitution{
 University of Florida
}
\addinstitution{
 Chinese Academy of Sciences
}
\runninghead{Yue et al}{Boosting up Scene Text Detectors with Guided CNN}

\def\eg{\emph{e.g}\bmvaOneDot}

\def\etal{\emph{et al}\bmvaOneDot}

\begin{document}

\maketitle
%{\color{red}{Boosting Scene Text Detectors with Guided Background Synthesis}}
\begin{abstract}
   Deep CNNs have achieved great success in text detection.
   Most of existing methods attempt to improve accuracy with sophisticated network design, while paying less attention on speed.
   In this paper, we propose a general framework for text detection called Guided CNN to achieve the two goals simultaneously. 
%    It mimics the procedure of glancing and focusing of the human visual system. 
   The proposed model consists of one guidance subnetwork, where a guidance mask is learned from the input image itself,
   and one primary text detector, where every convolution and non-linear operation are conducted only in the guidance mask.
%    On the one hand,   the guidance subnetwork filters out non-text regions coarsely, and greatly reduces the computation complexity.
  On the one hand,   the guidance subnetwork filters out non-text regions coarsely, greatly reduces the computation complexity.
   On the other hand, the primary text detector focuses on distinguishing between text and hard non-text regions and regressing text bounding boxes, achieves a better detection accuracy. 
   A training strategy, called background-aware block-wise random synthesis, is proposed to further boost up the performance. 
%    We propose one training strategy, called background-aware block-wise random synthesis, to further boost up the performance. 
   We demonstrate that the proposed Guided CNN is not only effective but also efficient
   with two state-of-the-art methods, CTPN~\cite{Tian2016} and EAST~\cite{Zhou2017}, as backbones. On the challenging benchmark ICDAR 2013, it speeds up CTPN by $2.9$ times on average, while improving the F-measure by $1.5\%$. On ICDAR 2015, it speeds up EAST by $2.0$ times while improving the F-measure by $1.0\%$.
\end{abstract}

%-------------------------------------------------------------------------
\section{Introduction}
\label{sec:intro}
Reading text in natural images in the wild has attracted increasing attention recently, as shown in~\cite{Yao2012,Neumann2012,Yin2013,Neumann2015, he2016reading, Gupta2016,Tian2016,Li2017,Liu2017,Zhou2017,He2017a, He2017b,Ma2017,Jiang2017,Dai2017}. Large variance of text patterns and highly cluttered background pose the main challenge for text detection.

% Traditional methods~\cite{Epshtein2010,Yao2012,Neumann2012,Yin2013,Huang2013,Tian2015,Neumann2015} 
% follow the bottom-up pipeline, including steps of character or text component detection, feature extraction,
% component classification or filtering, text line construction and word splitting. Multi-step systems 
% are highly complicated, and errors occurred in the earlier steps are easily accumulated in the later steps.
% Therefore, their final performance heavily depends on sophisticated design of each step, which greatly prohibits
% them from being used in various scenarios.

%Deep learning~\cite{lecun1998gradient} has achieved great successes in fundamental vision tasks such as image classification~\cite{krizhevsky2012imagenet,Christian_cvpr2015,he2016deep}, object detection~\cite{Girshick2014,girshickICCV15fastrcnn,Ren2017} and image segmentation~\cite{chen2014semantic, Shelhamer2016,He2017}.
% edited by yuexy A number of CNN based text detection methods~\cite{Wu2012, Jaderberg2014, Jaderberg2016,He2016a,Gupta2016,Tian2016,He2017a,Li2017,Liu2017,Zhou2017,He2017b,Ma2017,Jiang2017,Dai2017,Jiang2017a,Hu2017} have been proposed.

% A number of CNN based text detection methods~\cite{Wu2012, Jaderberg2014, Jaderberg2016,He2016a,Gupta2016,Tian2016,He2017a,Li2017,Liu2017,Zhou2017,He2017b,Ma2017,Jiang2017,Dai2017,Jiang2017a,Hu2017} have been proposed for text detection.
Inspired by recent advances in general object detection and semantic segmentation, such as Faster R-CNN~\cite{Ren2017}, SSD~\cite{Liu2016} and FCN~\cite{Shelhamer2016},
recent text detection approaches ~\cite{He2016a,Gupta2016,Tian2016,He2017a,Li2017,Liu2017,Zhou2017,He2017b,Ma2017,Jiang2017,Dai2017,Jiang2017a,Hu2017} directly 
predict the bounding boxes of text, 
and improve text detection accuracy considerably. They focus on designing of better network architecture and objective function while paying less attention on the detection speed, which prevents text detection from being deployed in practical applications. Although
there exist methods~\cite{Soudry2014, Denton2014a,Zhang2016a,Han2016,Wu2016, Li2017a} to accelerate general CNNs, they often achieve speedup at the cost of sacrificing accuracy, and are not tailored for text detection.

\begin{figure}
\begin{minipage}[c]{0.4\textwidth}
\includegraphics[width=5cm, height=2.5cm]{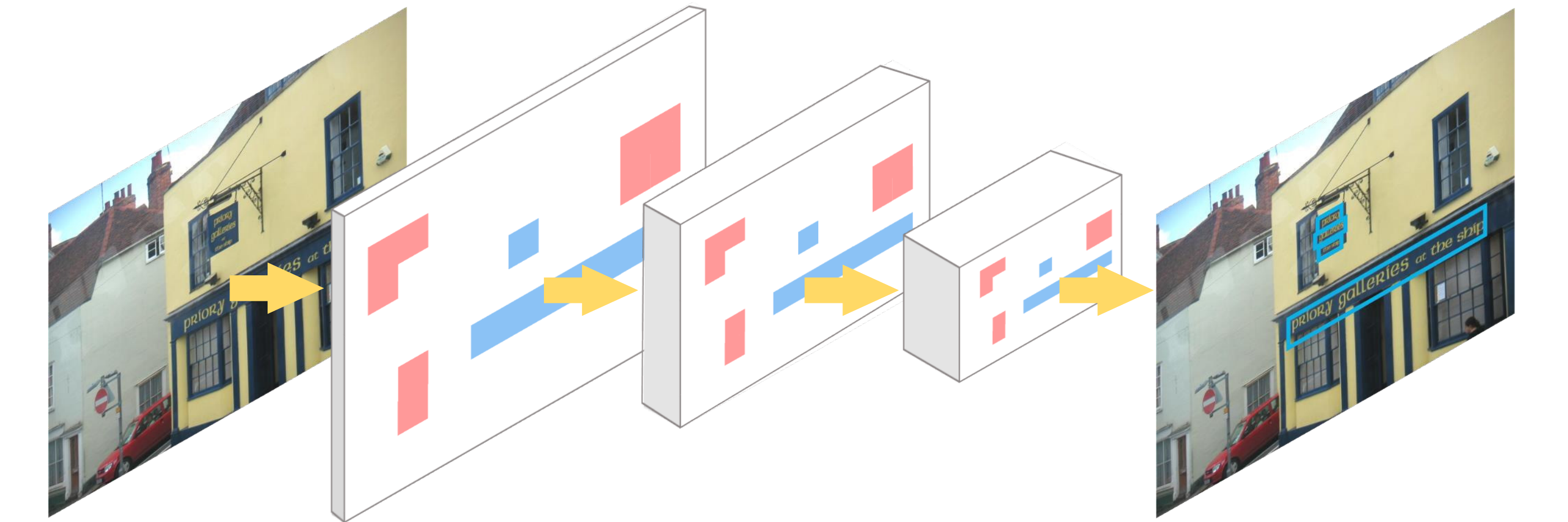}
\end{minipage}\hfill
\begin{minipage}[c]{0.6\textwidth}
\captionsetup{font=small}
\caption{
Illustration of guiding the primary text detector. Convolutions and non-linear operations are conducted only in  the guidance mask indicated by the red and blue rectangles. The guidance mask (the blue) is extended by background-aware block-wise random synthesis (the red) during training. When testing, the guidance mask is not extended.
} 
\label{fig:fig_text_detector}
\end{minipage}
\vspace{-5mm}
\end{figure}

\begin{figure}
\begin{minipage}[c]{0.5\textwidth}
\captionsetup{font=small}
\caption{
Text appears very sparsely in scene images. The left shows one example image. The right shows the text area ratio composition of ICDAR 2013 test set. Images with (0\%,10\%], (10\%,20\%], (20\%,30\%], and (30\%,40\%] text region account for 57\%, 21\%, 11\%, and 6\% respectively. Only 5 \% images have more than 40\% text region.
} 
\label{fig:fig_motivation}
\end{minipage}\hfill
\begin{minipage}[c]{0.5\textwidth}
\begin{tabular}{cc}
 {\includegraphics[width=2.6cm]{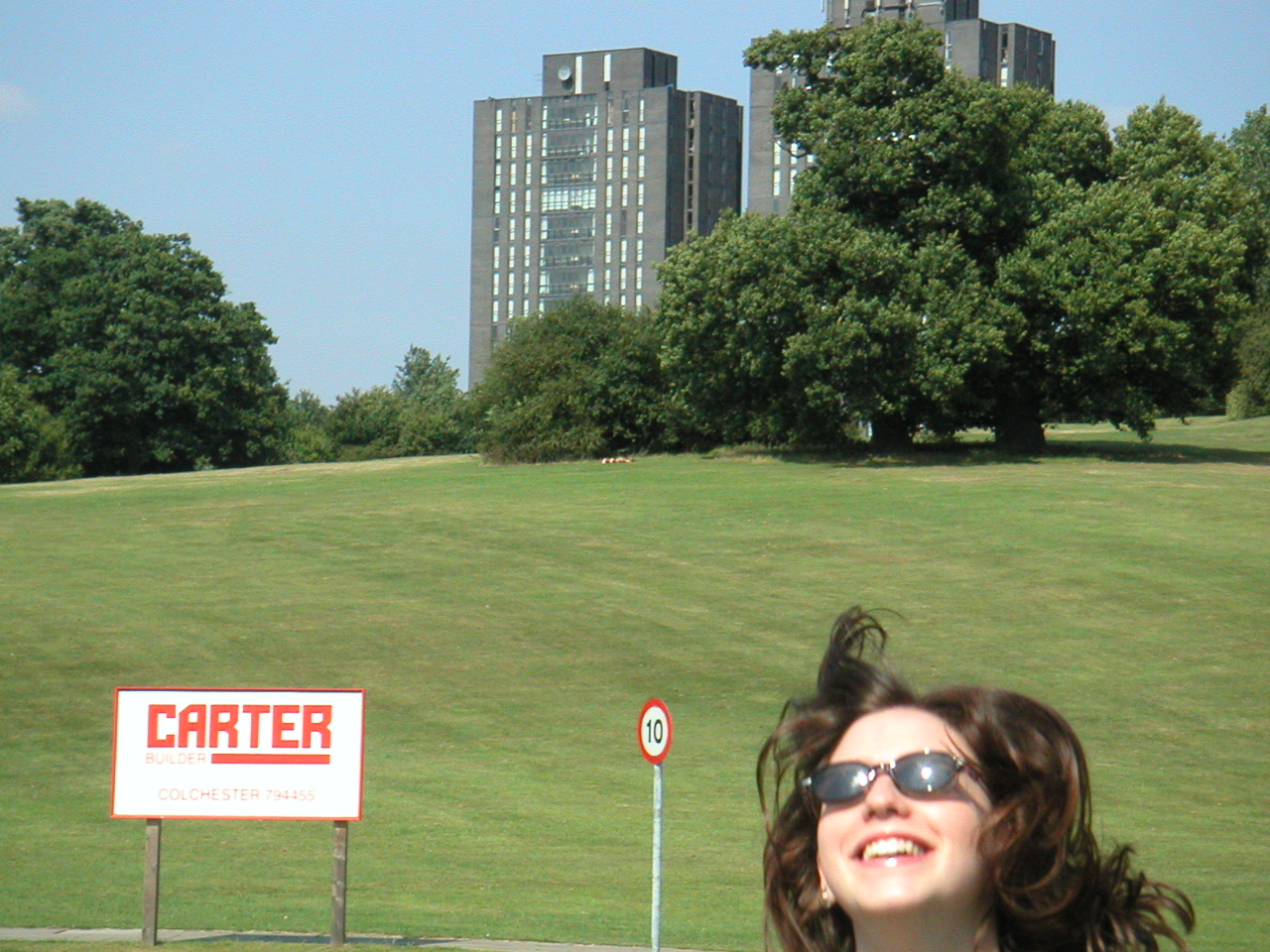}}
 & {\includegraphics[width=3.3cm]{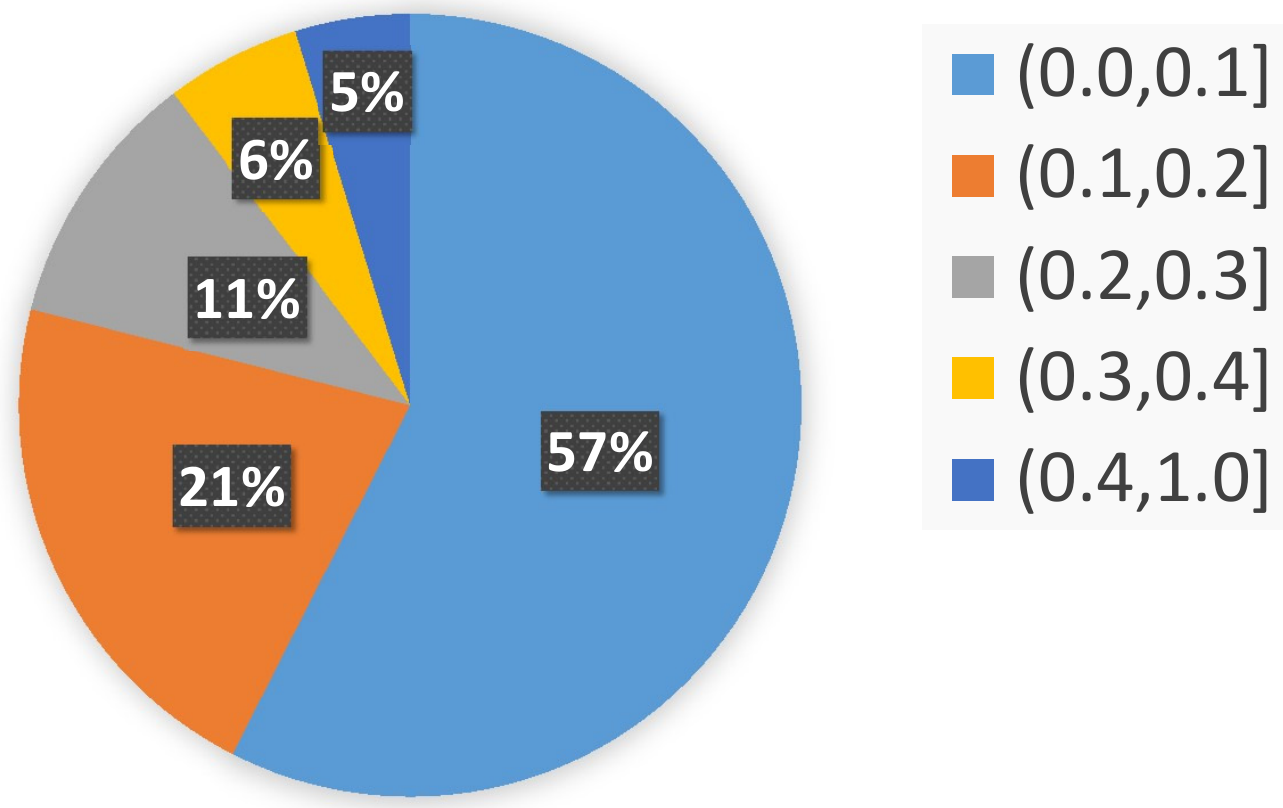}}
\end{tabular}
\end{minipage}
\vspace{-5mm}
\end{figure}

Text appears very sparsely in natural scene images, as shown in Figure~\ref{fig:fig_motivation}. Our statistics on ICDAR 2013 test set show that only 9.7\% of the whole region has text, and the rest  is background. Except for those background regions nearby texts, background regions are useless for text detection and text recognition, as discussed in~\cite{He2016a, He2017a}, and thus are unnecessarily processed in text detectors. Filtering out them could potentially speed up text detectors. However, filtering out background without sacrificing accuracy itself is not trivial.
% edited by yuexy, 5/6/2018
% On the one hand, this filtering operation should be fast with very little computation overhead. On the other hand, we need to achieve high recall of text regions after filtering as the text filtered out cannot be detected in the next stage. 
This filtering operation should be fast with very little computation overhead. We also need to achieve high recall of text regions after filtering as the text filtered out cannot be detected in the next stage. 

To this end, in this paper, we make full use of the characteristic of sparseness of text in the wild, and propose a simple yet effective general framework for text detection, called Guided CNN, to speed up text detectors, and improve their accuracy simultaneously. 
It mimics the procedure of the human visual system: first glancing at the full image and then focusing on salient regions. Specifically, it consists of two subnetworks: one guidance subnetwork, and one primary text detector.
The guidance subnetwork is learned from the input images with text regions as supervision signals.
It targets at finding text regions and filtering out most of non-text regions coarsely.
The primary text detector uses any existing single forward CNN based text detection method as backbones,
where every convolution, and nonlinear operation are conducted only in the guidance mask predicted by the guidance subnetwork (See Figure~\ref{fig:fig_text_detector}).
In this way, the primary text detector focuses on distinguishing between text and hard non-text, and predicting bounding boxes of text.
% edited by yuexy
% Note that the SSTD~\cite{He2017a} proposed the regional attention to suppress background interference, but still detected text on the whole image. Our proposed method shares the similar spirit but in the meantime further reduces much computation. It can also enhance and speed up SSTD. 

Improving speed and accuracy simultaneously is challenging. In order to achieve good trade-off of guidance performance and computational load, we introduce a \textit{context module} with pyramid pooling layers, to capture global context information in the guidance subnetwork. 
% edited by yuexy, 5/6/2018
Training the primary text detector, only with the guidance mask derived by the ground truth,  would lead to high false alarm rate, as it does not learn enough background patterns.
% Training the primary text detector, only with the guidance mask predicted by the guidance subnetwork,  would lead to high false alarm rate, as it does not learn enough background patterns.
To resolve this issue, we propose \textit{background-aware block-wise random synthesis} to extend the guidance mask, by randomly synthesizing foreground blocks in background , with a fixed probability during training (See Figure~\ref{fig:fig_text_detector}).
% To resolve this issue, we propose \textit{background-aware block-wise random synthesis} to extend the predicted guidance mask, by randomly synthesizing foreground blocks in background , with a fixed probability during training (See Figure~\ref{fig:fig_intro}).
It can be interpreted as a special kind of \textit{dropout}. %We present comprehensive experiments to verify its effectiveness on various benchmarks.

The proposed Guided CNN is  a general framework, which can be easily plugged into existing single forward CNN based text detection methods. 
%The proposed method significantly speed up the recent state-of-the-art methods~\cite{Tian2016,Zhao2017}, while achieving a higher accuracy on average. 
We conduct experiments on standard benchmarks, namely SWT~\cite{Epshtein2010}, ICDAR 2011~\cite{Minetto2010}, ICDAR  2013~\cite{Karatzas2013} 
and ICDAR 2015~\cite{Karatzas2015}, with various backbones. The proposed method significantly speeds up the recent state-of-the-art methods~\cite{Tian2016,Zhou2017}, while achieving a higher accuracy on average. Specifically, on ICDAR 2013, it speeds up CTPN~\cite{Tian2016} by $2.9$ times, while improving accuracy by 1.5\%  on average. On ICDAR 2015, it speeds up EAST~\cite{Zhou2017} by $2.0$ times, while improving accuracy by 1.0\%. It achieves an F-measure of 0.901 on ICDAR 2013, and 0.823 on ICDAR 2015 with a single scale test without bells and whistles.

% edited by yuexy
% The rest of this paper is organized as follows. Section 2 gives an
% overview of related work. Section 3 details our proposed method.
% Section 4 summarizes our experimental results, with concluding
% remarks in Section 5.

%-------------------------------------------------------------------------
\section{Related Work}
\label{sec:rel_work}
Existing methods in the literature for text detection can
roughly be categorized into two approaches, namely traditional
feature based approaches and deep learning based approaches.

\textbf{Traditional feature based approaches.} 
Hand-crafted features such as wavelet, gradient, and HOG, 
which are widely used in other computer vision applications, 
were introduced for text detection in~\cite{Li2000,Lienhart2002,Chen2004,Shi2013,Tian2015}. 
Later on, Stroke Width Transform (SWT) was proposed 
to find the value of stroke width for each image pixel~\cite{Epshtein2010,Yao2012}.
It was extended by incorporating color cues of text pixels, leading to significantly enhanced performance on inter-component separation and intra-component 
connection~\cite{Huang2013}. 
Extremal Regions (ER) were employed to generate character candidates~\cite{Neumann2012,Neumann2016}, which was then improved by 
Maximally Stable Extremal Regions (MSERs)~\cite{Yin2013,Huang2014a,Zamberletti2015a}.
% edited by yuexy, 5/6/18
% Recently, FASText was proposed to detect strokes based on an efficient pixel intensity comparison to surrounding pixels~\cite{Buta2015}. 
% Zhang \etal~\cite{Zhang2015} proposed to extract text lines directly from natural images based on the symmetry property of character groups.
% Cho, Sung, Jun~\cite{Cho2016} proposed the Canny text detector by making full use of similar properties such as spatial location, size, color, and stroke width between words regardless of language.  
Different from all the above methods which use traditional hand-crafted features, the proposed general framework unifies feature extraction and text detection in a deep neural network, and both the feature extractor and detector are trained in an end-to-end fashion.

\textbf{Deep learning based approaches.}  Earlier work~\cite{Wu2012, Jaderberg2014, Jaderberg2016,He2016a} in this stream detected text with CNNs on sliding windows or region proposals.
They didn't share computation between sliding windows or region proposals, and thus leaded to slow training and testing.
Inspired by the state-of-the-art object detection, and segmentation techniques, such as Faster R-CNN~\cite{Ren2017},  SSD~\cite{Liu2016} and FCN~\cite{Shelhamer2016}, a number of text detection methods with a single forward pass were proposed.  
Yao \etal~\cite{Yao2016} and Zhang \etal~\cite{Zhang2016}  estimated text regions, individual characters and their relationship with FCN. 
% edited by yuexy
% Gupta, Vedaldi and Zisserman~\cite{Gupta2016} directly predicted one textness score and six text bounding box parameters on convolutional feature maps. 
Different from general objects, words tend to have large aspect ratios, and multiple orientations. 
To this end, TextBoxes~\cite{Liao2017} employed default boxes with big range aspect ratios and vertical offsets.
CTPN~\cite{Tian2016} fixed the width of default boxes and only predicted height of each default box. 
% Regional attention and multi-scale inception architecture were designed to encode strong context information~\cite{He2017a}. 
% Li, Wang and Shen~\cite{Li2017} re-sampled text proposal regions according to their respective aspect ratios, and then used RNNs to encode the resulting feature maps of different lengths into fixed length vectors.
Deep matching prior network~\cite{Liu2017} used quadrilateral sliding windows, and EAST~\cite{Zhou2017} directly predicted quadrilateral shapes on convolutional feature maps.
% He \etal~\cite{He2017b} directly performed the text bounding box regression without any default boxes.
Ma \etal~\cite{Ma2017} proposed  rotation proposals to detect arbitrary-oriented scene text while Jiang \etal~\cite{Jiang2017} predicted axis-aligned bounding box and inclined bounding box as multi-task learning.  
% Dai \etal~\cite{Dai2017} detected and segmented the text instance jointly and simultaneously as done in Mask-RCNN~\cite{He2017}. 
% Jiang, Hao and Liu~\cite{Jiang2017a} combined segmentation and detection results, and obtained impressive results. 
% Other work include WordSup~\cite{Hu2017} which exploited word annotations to character based text detection. 
\textit{
Different from all the above methods, the proposed Guided CNN is a general framework to improve both the accuracy and speed based on existing text detectors with a single forward pass. 
% We demonstrate its effectiveness based CTPN and EAST.
}

Our work is related to the work of network acceleration. Linear decomposition was proposed to accelerate layers of CNNs~\cite{Denton2014a,Lebedev2015,Jaderberg2014a}, which was further extended by considering nonlinear approximation in ~\cite{Zhang2016a}.
Quantization, pruning, and Huffman coding were proposed to speed up CNNs and compress its weights in~\cite{Han2016,Wu2016}. 
Weights of CNNs were binarized  while keeping the structure of network unchanged in~\cite{Rastegari2016,Courbariaux2015,Esser2015,Soudry2014}. 
% Convolution feature was mimicked for training very efficient deep networks~\cite{Li2017a}.
\textit{Our proposed Guided CNN is orthogonal to those methods and can be combined with them for further acceleration.}

Our work is also related to cascaded CNNs. 
\cite{Zhang2016b,Qin2016,Zhang2017,LiHaoxiangandLinZheandShenXiaohuiandBrandtJonathanandHua2015} first used a sliding window method, a fully convolution network, or a region proposal network to generate face candidates, and then one by one fed each cropped face candidate into a series of cascaded CNNs for further filtering and bounding box refinement. 
\cite{li2017layercascade} treated a single deep model as a cascade of several sub-models and processed harder regions progressively.  
% \cite{Yang2016} conducted cascaded refinement on scale dependent pooled feature maps.
\textit{
Different from the above methods, our Guided CNN has no object (i.e., text) candidates, but predicts a guidance mask which can be of any shape to tell the primary text detector where its convolution and nonlinear operations are conducted. 
% All isolated regions of the guidance mask are handled by the text detector with one pass but not one by one. 
% Besides, it targets at improving the speed and accuracy of existing methods at the same time.
}

%-------------------------------------------------------------------------
\section{Guided CNN}
\label{sec:guided_cnn}
\begin{figure}
\begin{minipage}{1\linewidth}
\centerline{ \includegraphics[width =0.93\textwidth, height=3.3cm]{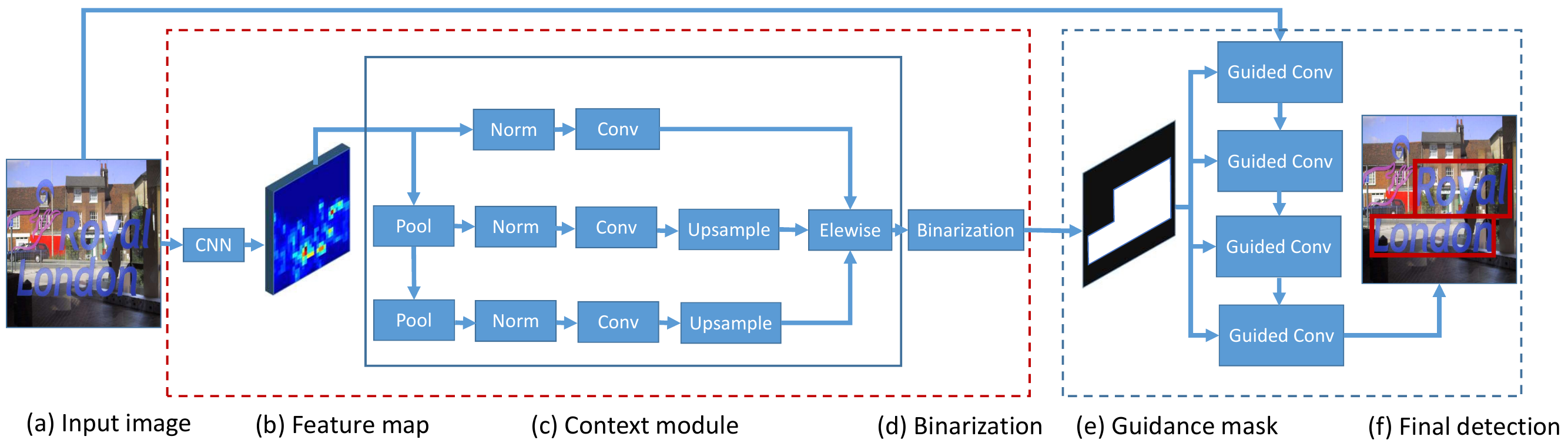}}
%\subcaption{  \label{figa} }
\end{minipage}
\begin{minipage}{1\linewidth}
\captionsetup{font=small}
\caption{
The Guided CNN framework for text detection. It consists of one guidance subnetwork indicated by the red dash rectangle, one primary text detector indicated by the blue dash rectangle. Given an input image (a), we first use CNN to get the convolutional feature map (b), then a context module (c) is used to capture multi-scale context  and predict one guidance map, followed by binarizing (d) the guidance map to one guidance mask (e). 
%The guidance mask is further extended by background-aware block-wise random synthesis (e) during training, resulting in extended guidance mask.
Finally, the (extended) guidance mask is used to guide the primary text detector to detect text (during training), resulting in the final detection (f).}
\label{fig:fig_framework}
\end{minipage}
\vspace{-4mm}
\end{figure}

Our proposed Guided CNN is based on the observation that text appears very sparsely in natural scene images as shown in Figure~\ref{fig:fig_motivation}. We target at improving the speed and accuracy of existing CNN base text detectors by filtering out most of the background. We achieve these two goals by proposing Guided CNN with a carefully-designed network architecture and its corresponding training procedure.
%We first introduce the overall framework of Guided CNN, and then presents its two components, namely, the guidance subnetwork, and the primary text detector. We finally discuss about the training procedure with background-aware block-wise random synthesis and its connection with dropout.
% \subsection{Overall Framework}
%\label{subsec:subsec_overfall_framework}
An overview of our framework is illustrated in Figure~\ref{fig:fig_framework}.  It consists of a guidance subnetwork, and a primary text detector. 
The guidance subnetwork is designed to coarsely predict the guidance mask, i.e., the text regions. 
The primary text detector conducts convolution and non-linear operations only in the  guidance mask. 
Our Guided CNN is a general framework for text detection. Its primary text detector can use any existing single forward CNN based text detector as backbone. 
% edited by yuexy
% To demonstrate the general applicability   of our framework, we instantiate Guided CNN with multiple architectures, such CTPN~\cite{Tian2016}, and EAST~\cite{Zhou2017}. 
We denote Guided CNN with backbone using the nomenclature \textit{Guided backbone}. Guided CNN with CTPN as backbone, for example, is called Guided CTPN. 

\subsection{Guidance Subnetwork} 
\label{subsec:subsec_guidance_subnetwork}
The guidance subnetwork uses the pretrained PVANET~\cite{Kim2016} to extract the convolutional feature map as shown in Figure~\ref{fig:fig_framework} (b).
The final feature map, i.e., $conv5\_4$, is of $\frac{1}{32}$ size of the input image.
On the top of the map, we use a context module to extract multiple scale context and predict the guidance as shown in Figure~\ref{fig:fig_framework} (c).
Its output keeps $\frac{1}{32}$ of the input image, and is fed into a cross entropy loss layer during training.
%The context module is followed by a convolution layer to generate one prediction map for each context scale .
%The prediction maps of multiple scale are added together, resulting the final prediction map which is fed into a cross entropy loss layer.
We generate the ground truth guidance mask using text bounding boxes of training data. We denote a rectangle by a four-tuple $(r, c, h, w)$ that specifies its top-left
corner $(r, c)$ and its height and width $(h, w)$. For high recall, we set the ground truth label of the location $(y,x)$ on the final
prediction guidance map to be $1$ as long as its corresponding rectangle $(32y-16,32x-16, 32, 32)$ in the input image has intersection with any text bounding box. The predicted guidance map is binarized (Figure~\ref{fig:fig_framework} (d)) with a threshold 
$\tau$ to a guidance mask as shown in Figure~\ref{fig:fig_framework} (e).

Inspired by the recent advancement of image segmentation~\cite{Liu2015},
we design the \textit{context module} with pyramid pooling.
As shown in Figure~\ref{fig:fig_framework} (c),
the context module consists of 3 levels. 
At the first level (the first row), feature vectors
at each location are $l2$ normalized, followed by a convolutional layer to predict the guidance map.
The latter two levels follow the same procedure,
except that they use one average  pooling layer with stride $2$ to down-sample the feature map at the beginning and up-sample back after predicting the guidance map at the end. 
%The second scale  is generated by max-pooling the feature maps of the first scale with stride $2$
%while the third scale in red is further generated from the second scale with the same operation.
The normalization layer at each level is introduced to avoid the unbalance magnitude of the feature map. The predicted guidance maps of all levels are element-wise added together before feeding them to the loss layer.

\begin{figure}
\begin{minipage}[c]{0.4\textwidth}
\begin{tabular}{cc}
{\includegraphics[width=3.5cm]{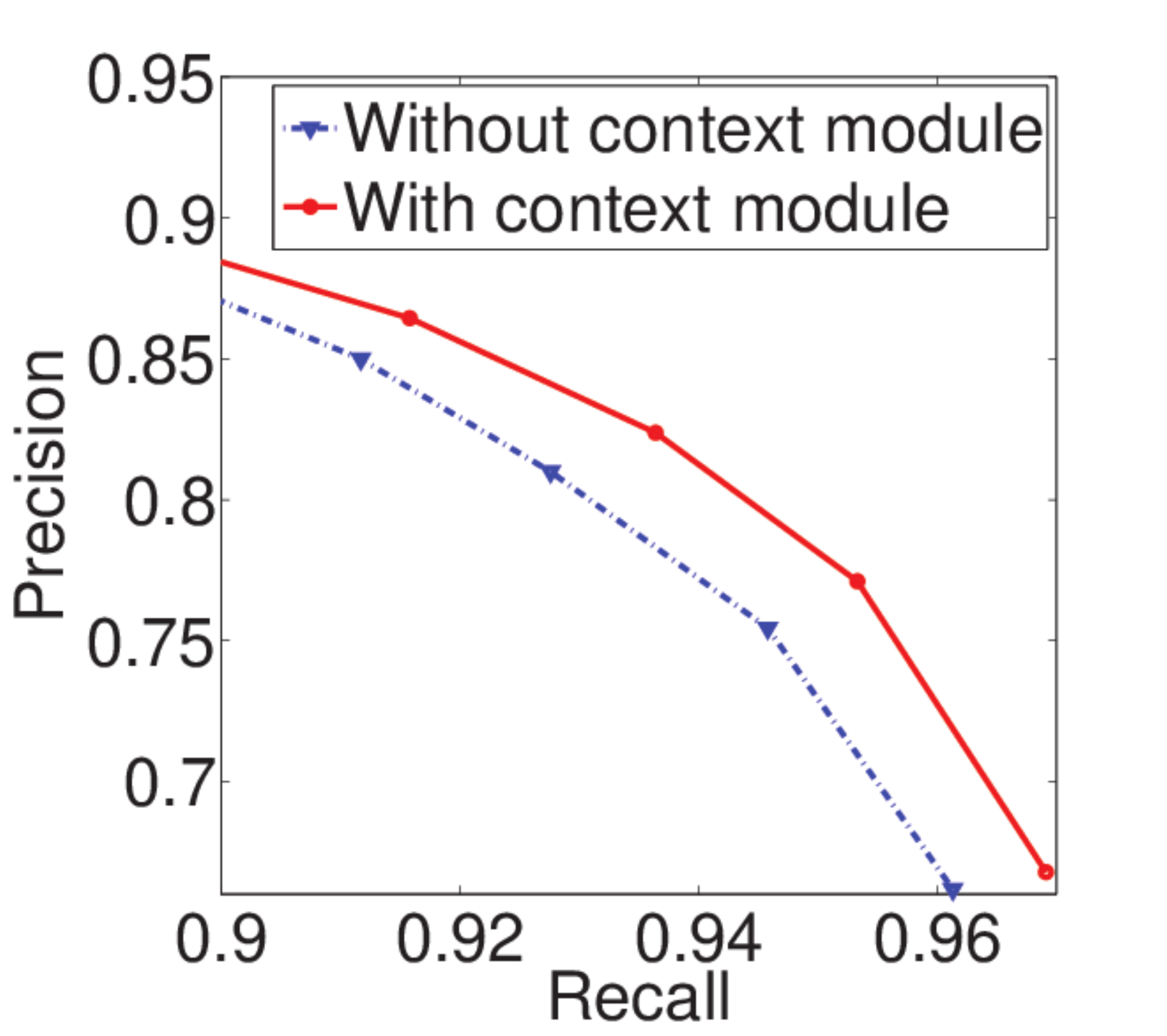}} &
{\includegraphics[width=3cm, height=2.9cm]{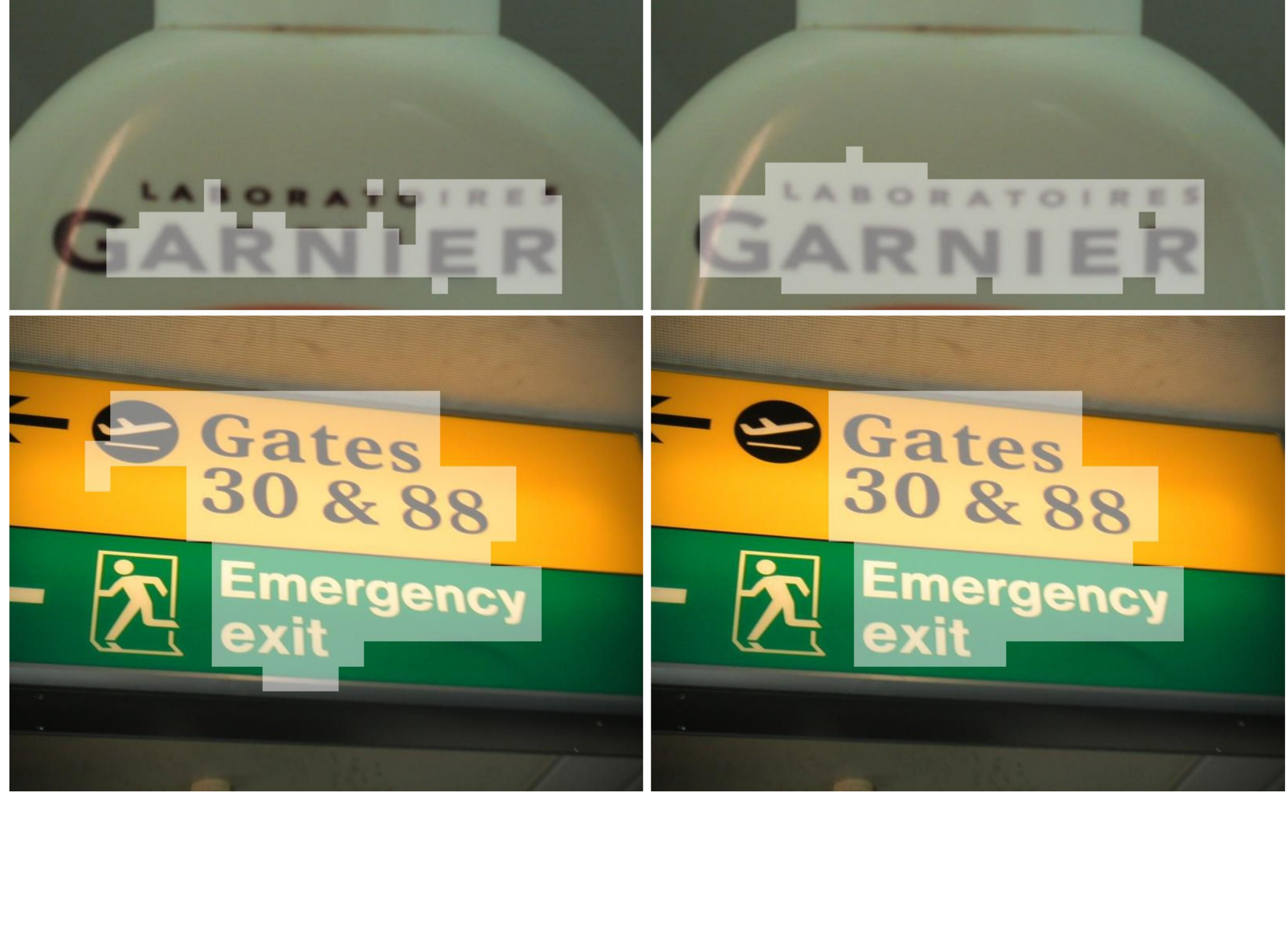}} \\
(a) & (b)
\end{tabular}
\end{minipage}\hfill
\begin{minipage}[c]{0.42\textwidth}
\captionsetup{font=small}
\caption{Quantitative and qualitative comparisons of the guidance subnetwork with and without the context module. (a) Recall precision curves of the guidance subnetwork with or without the context module.
% Different recall and precision are obtained by varying the threshold $\tau$. 
(b)
% A visualization of the predicted guidance mask by the guidance subnetwork with or without the context module.
The left and right columns show the input images superposed with the predicted guidance mask without and with the context module, respectively.}
\label{fig:fig_context_module}
\end{minipage}
\vspace{-4mm}
\end{figure}

% \begin{figure}[tpb]
% \begin{tabular}{cc}
% {\includegraphics[width=0.3\linewidth]{images/fig_recall_precision.eps}} 
%  & {\includegraphics[width=0.2\linewidth, height=0.2\linewidth]
%  {images/example.eps}} \\
%  (a) & (b)
% \end{tabular}
% \caption{Quantitative and qualitative comparisons of the guidance subnetwork with and without the context module. (a) Recall precision curves of the guidance subnetwork with or without the context module. Different recall and precision are obtained by varying the threshold $\tau$. (b) A visualization of the predicted guidance mask by the guidance subnetwork with or without the context module. The left and right columns show the input images superposed with the predicted guidance mask without and with the context module, respectively.}
% \label{fig:fig_context_module}
% \end{figure}

The context module is essential to guarantee that the guidance subnetwork gets high recall and high precision. 
Figure~\ref{fig:fig_context_module} (a) compares the recall and precision of the guidance subnetwork with
context module and those without it (i.e., the first level prediction)
on ICDAR 2013.
% {\color{red} on which dataset??}.
It has been shown that the guidance subnetwork with context module obtains much more accurate guidance
prediction than that without it. 
Figure~\ref{fig:fig_context_module} (b) visualizes
the guidance mask (binarized guidance map) predicted with context module and that without it. Obviously, 
context module can reduce false alarms greatly and increase recall. We emphasize that the cost of context module can be ignorable compared with that of the whole guidance subnetwork since the feature map size of context module is small enough (i.e., $\frac{1}{32}$ of the input image size).

\subsection{Primary Text Detector}  
The primary text detector can use any existing
single forward CNN based text detector as backbone. 
In existing text detectors,  each convolution layer has only one input, convolution operation is done
in the whole input feature map. Instead, in our primary text detector,
each convolution layer has two inputs, namely, the input feature map and the guidance mask predicted by the guidance subnetwork,
and convolution operation is done only in  the guidance mask.

\begin{figure}
\begin{minipage}[c]{0.7\textwidth}
\centerline{ \includegraphics[width=8cm]{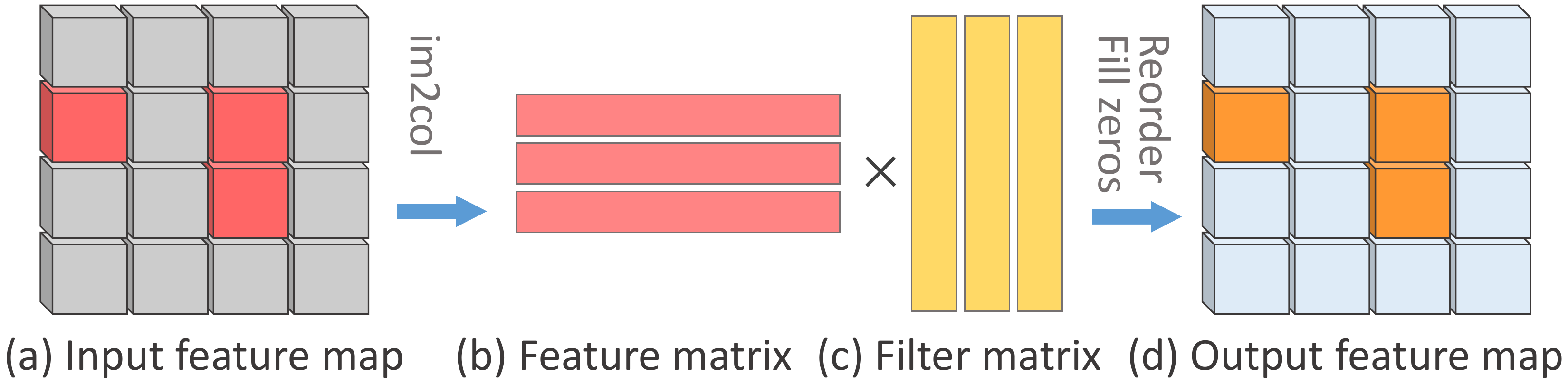}}
\end{minipage}\hfill
\begin{minipage}[c]{0.28\textwidth}
\captionsetup{font=small}
\caption{Illustration of the guided convolution. The red indicates the feature vectors in  the guidance mask. }
\label{fig:fig_im2col}
\end{minipage}
\vspace{-5mm}
\end{figure}

% \begin{figure}
% \begin{minipage}{\linewidth}
% \centerline{ \includegraphics[width =1\textwidth]{images/fig_im2col.pdf}}
% %\subcaption{  \label{figa} }
% \end{minipage}
% \caption{Illustration of the guided convolution. The red indicates the feature vectors in  the guidance mask. }
% \label{fig:fig_im2col}
% \end{figure}

All the regions %(possibly with multiple isolated regions)
in the guidance mask are handled in the primary text detector with one pass but not one by one. To this end, we implement a new layer called \textit{guided convolution} by modifying the ``im2col'' in caffe~\cite{Jia2014} as shown in Figure~\ref{fig:fig_im2col}. Different from the original ``im2col'' which
reorders the input feature map into a big matrix, 
our guided convolution layer reorders the input feature vectors only in 
the guidance mask (Figure~\ref{fig:fig_im2col} (a)) and generates a much smaller matrix (Figure~\ref{fig:fig_im2col} (b)).  After matrix multiplication with the convolution filter matrix (Figure~\ref{fig:fig_im2col} (c)), 
its output is inserted back into a big output feature map  with zero filling in background (Figure~\ref{fig:fig_im2col} (d)). From Figure~\ref{fig:fig_im2col}, it can be concluded that Guided CNN can speed up its backbone by $r$ times with ignoring the overhead of the guidance subnetwork if the guidance mask accounts for the ratio of $\frac{1}{r}$ over the whole feature map area.

\subsection{Training with Background-aware Block-wise Random Synthesis}
 \label{subsec:subsec_training}
Training the primary text detector with guidance is not trivial. Directly training the primary text detector
leads to a high false alarm rate. It is reasonable since the primary text detector fails to learn many background pattern variations during training.

To this end, we extend the predicted guidance mask by randomly synthesizing mask blocks with size $32\times 32$ (since each location represents $32\times 32$ rectangle in the input 
image) in the background with a probability $p$ during training.
We experimentally find that this simple approach is very effective, and can improve the final text detection performance greatly. 
When $p=0$, the primary text detector is trained without randomly synthesis. 
When $p=1$, it is degraded to its backbone without the guidance mask where convolution and other non-linear operations are conducted in the whole feature map.

\begin{figure}
\begin{minipage}[c]{0.37\textwidth}
\centerline{ \includegraphics[width=4.7cm]{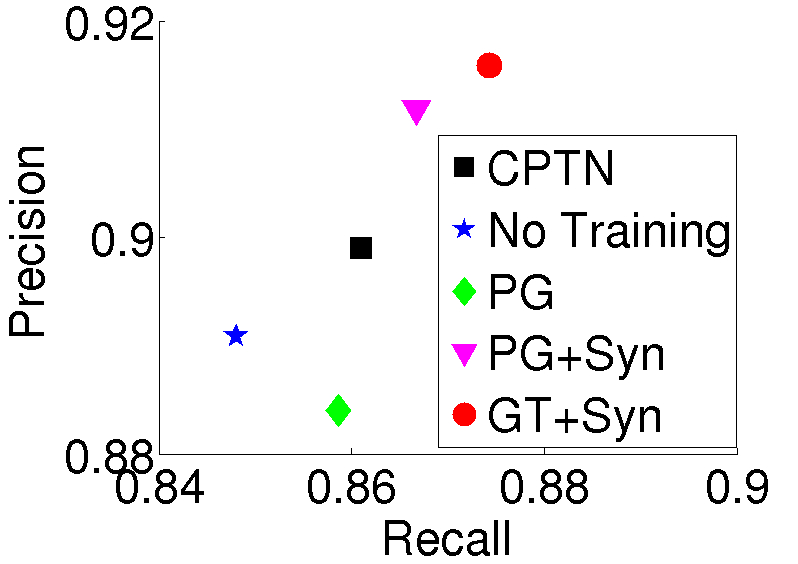}}
\end{minipage}\hfill
\begin{minipage}[c]{0.63\textwidth}
\captionsetup{font=small}
\caption{Performance improved by background-aware block-wise random synthesis. $\blacksquare$ indicates CTPN, {\color{blue}$\star$} indicates applying the predicted guidance mask on CTPN during testing without retraining, {\color{green}$\blacklozenge$} indicates Guided CTPN trained with the predicted guidance mask, {\color{magenta}{$\blacktriangledown$}} indicates Guided CTPN trained with the predicted guidance mask and random synthesis, and  {\color{red} {$\large\bullet$}} indicates Guided CTPN trained with the ground truth mask and random synthesis.}
\label{fig:fig_randomly_synthesis}
\end{minipage}
\vspace{-6mm}
\end{figure}

% \begin{figure}
% \begin{minipage}{0.9\linewidth}
% \centerline{ \includegraphics[width =0.6\textwidth]{images/fig_training.jpg}}
% %\subcaption{  \label{figa} }
% \end{minipage}
% \caption{TODO2}
% \label{fig:fig_randomly_synthesis}
% \end{figure}

Figure~\ref{fig:fig_randomly_synthesis} compares the results on ICDAR 2013 with different training strategy. Firstly, we directly apply our predicted guidance mask on the original CTPN without retraining. It performs slightly worse than the original CTPN. It is straightforward since the contour patterns on the feature maps caused by predicted mask are never
learned during training.
Therefore, we then 
retrain Guided CTPN with the Predicted Guidance (PG). Its performance is improved but still worse than CTPN. Experimental results show that it has many false alarms. As we discussed before, totally dropping background patterns causes this issue. It is further validated in our third strategy, where the primary text detector is trained with both the predicted guidance mask and background-aware block-wise random Synthesis (Syn) ($p=0.4$). Training with this strategy performs even better than the original CTPN, while speeding up considerably. Although the predicted guidance mask is with high recall, it might miss some small or big-scale texts. To resolve this issue, we instead choose the forth strategy, where the Ground Truth guidance mask (GT), together with random Synthesis (Syn),
are used to train the primary text detector. We find that this training strategy further gives a better performance, in terms of recall and precision.
The rest of experimental results is reported using this training strategy except as otherwise noted.

\textbf{Advantages.} Training the primary text detector with the ground truth guidance and random synthesis achieves two obvious advantages. 
First, the primary text detector can obtain better performance than its backbone while speeding up it greatly. 
Second, the training of guidance subnetwork, and that of the primary text detector are not correlated.  Therefore, 
they can be trained individually and in parallel, leading to convenient deployment.

\textbf{Connection with Dropout}. 
\label{subsec:subsec_dropout}
One possible explanation for our background-aware block-wise random synthesis is to dropout~\cite{Srivastava2014} feature map with probability $1-p$ in background. Following standard test procedure of dropout, 
we multiply the feature map in the primary text detector with a scale of $p$ but not $0$ in background regions during testing,
the primary text detector can be further improved.
For clarify, we called this testing method as \textit{Guided backbone+}. For example, Guided CNN with CTPN as its backbone is called Guided CTPN+ if
 it follows the dropout test procedure when testing.  We emphasize that
 Guided backbone+ would not accelerate its backbone. It can be used in the case when accuracy but not speed is most important. 

The background-aware block-wise random synthesis is different from the standard dropout. First, the standard dropout drops feature activations independently 
while our method drop all activations of one block once. Second the standard dropout drops activation everywhere while our method only in the background (non masked regions). We did experiments by inserting one dropout layer before the detection layer, and didn't observe any improvement.

%-------------------------------------------------------------------------
\section{Experiments}
\label{sec:experiments}
We conduct comprehensive ablation experiments along with a thorough comparison of 
the proposed Guided CNN and the state-of-the-art text detection methods. We instantiate our Guided CNN with CTPN~\cite{Tian2016} and EAST~\cite{Zhou2017} since they are 
the state-of-the-art text detection methods for near horizontal text and multi-orientation text respectively. 

\subsection{Datasets and Evaluation}
The proposed method are evaluated on four text detection benchmarks, namely, SWT~\cite{Epshtein2010}, ICDAR 2011~\cite{Minetto2010}, ICDAR 2013~\cite{Karatzas2013},
and ICDAR 2015~\cite{Karatzas2015}. 
% ICDAR 2011, ICDAR 2013, and ICDAR 2015 are from the Robust Reading Competition held 
% in 2011, 2013, and 2015 respectively.
ICDAR datasets consist of 229 images for training and 255 for testing, 229 for training and 233 for testing, and 1,000 for training and 500 for testing respectively. All the images are labeled in word level.
ICDAR 2011, and ICDAR 2013 focuses on horizontal or near-horizontal text while ICDAR 2015 targets at incidental scene text with multiple orientations. %, which are collected by Google Glass.
The SWT datasets contains 307 color images with sizes ranging from $1024\times 368$ to $1024\times 768$. The SWT dataset is more challenging than ICDAR 2011 and ICDAR 2013  because of extremely smaller texts, repeating patterns, various plants, etc. We follow previous work by using standard evaluation protocols which are provided by the dataset creators or competition organizers.  Our training dataset contains the training sets of the ICDAR 2013 and ICDAR 2015. We also collect images from Internet as the training data as done in~\cite{Tian2016,He2017a}. We have $4000$ training images in total. 

\subsection{Implementation Details}
We use caffe~\cite{Jia2014} to implement all our experiments 
except the training 
related with~EAST~\cite{Zhou2017} which is done with tensorflow. 

For the guidance subnetwork, its backbone is pretrained on ImageNet. We preprocess the input image as its corresponding primary text detector. 
We employ multi-step learning rate policy with the base learning rate being  $0.01$. 
The learning rate decays by $10$ times every $6000$ iterations,
and the maximum iteration size is set to  $30000$. For stable training, we set $iter\_size$ to  $10$ to obtain stable gradients. 

For the primary text detector, we finetune its corresponding backbone model with
the ground truth guidance extended by random synthesis. CTPN~\cite{Tian2016} is reimplemented by ourselves as its model and  source code released by its original authors without side refinement\footnote{https://github.com/tianzhi0549/CTPN}. The source code and the model of EAST is reimplemented and released by the third party~\footnote{https://github.com/argman/EAST}. The re-implemented EAST uses ResNet-50~\cite{he2016deep} as its backbone model, while the original EAST VGG~\cite{Simonyan15}, PVANET~\cite{Kim2016} and PVANET with double channels in \cite{Zhao2017}. The input image preprocessing procedure keeps the same as that in \cite{Tian2016,Liu2017} during training and testing. 
%Again, we use multi-step
% learning rate policy with the base learning rate being $0.001$. The learning rate decays by $10$ times every $16000$ iterations,
% and the maximum iteration size is set to  $30000$. For stable training, we set $iter\_size$ to $5$ to obtain stable gradient. 

% \textbf{Training dataset.} Our training dataset contains the training sets of the ICDAR 2013 and ICDAR 2015. We 
% also collect images from Internet as the training data as done in~\cite{Tian2016,He2017a}.  
% We have $4000$ training images in total. All the models related with CTPN are trained with this dataset. For fair comparison, for models related with EAST, we follow the instruction of its open source code, and train them on the training dataset of ICDAR 2013 and ICDAR 2015.

\textbf{Runtime.} As for testing the forward speed, 
 we measure the runtime of all models  on a PC with $i7$ CPU with caffe~\cite{Jia2014} as text detection approaches are deployed on devices with CPU in most scenarios. 
  The runtime of Guided CNN (or Guided CNN+) includes that of two subnetworks. 
%  The runtime of Guided CNN (or Guided CNN+) includes the runtime of the two subnetworks. 
 For models related with EAST, we convert them to caffe models before testing.  All results including accuracy and runtime are reported with \textit{a single scale test}.

\subsection{Ablation Experiments}
We run a number of ablations to analyze the impact of hyper parameters and the effectiveness of each component of the proposed Guided CNN.

\textbf{The impact of parameters $\tau$ and $p$.}
There are only two parameters (i.e., the threshold $\tau$
to binarize the predict guidance map as described in Section~\ref{subsec:subsec_guidance_subnetwork}  and the probability $p$ of background-aware block-wise random synthesis as described in Section~\ref{subsec:subsec_training})  in Guided CNN. To demonstrate 
the impact of the parameters on the performance of the model, 
we conduct experiments on ICDAR 2013 based on CTPN under different parameter settings.

\begin{figure}
\begin{minipage}[c]{0.3\textwidth}
\centerline{ \includegraphics[width=4cm, height=3cm]{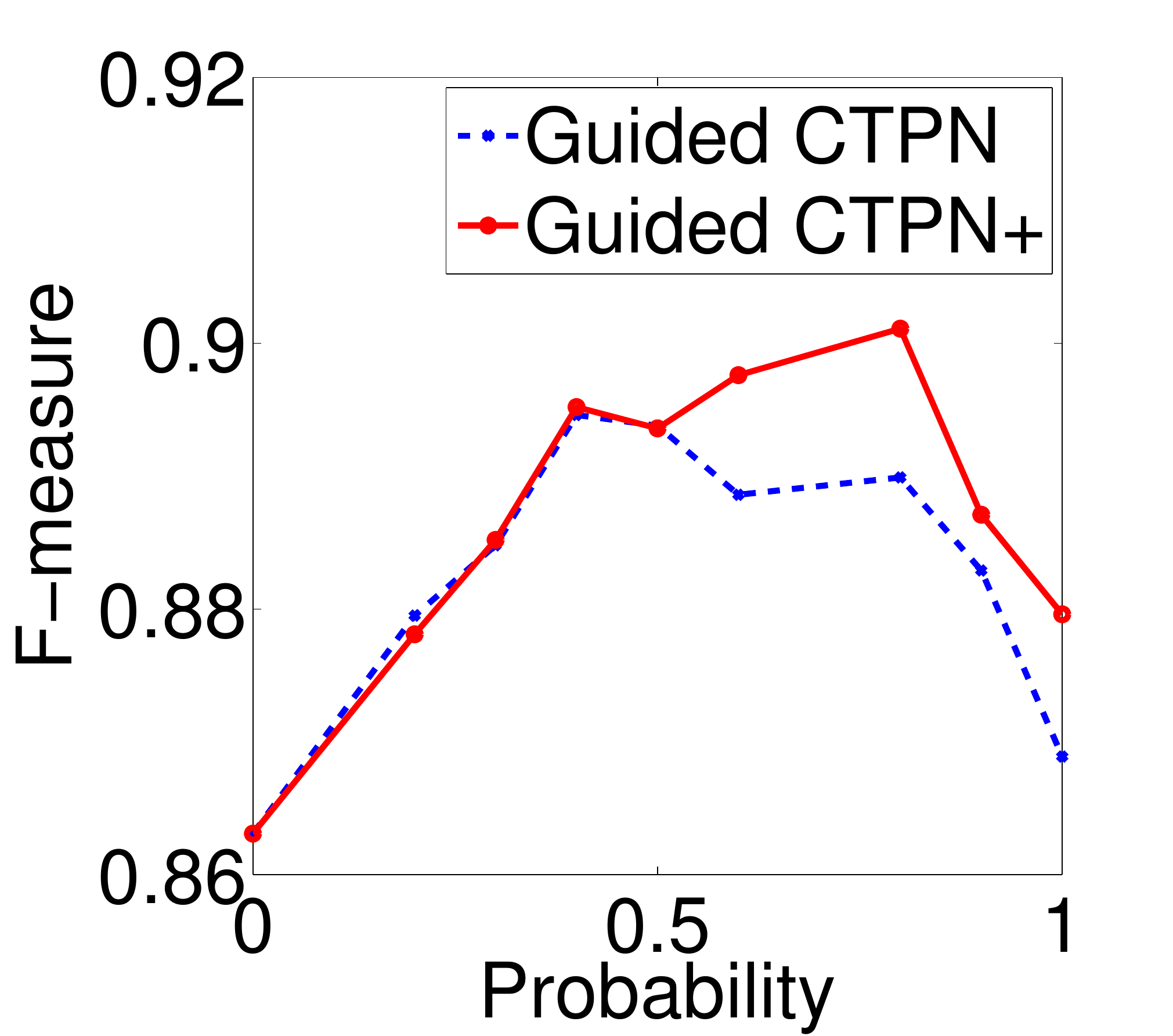}}
\end{minipage}\hfill
\begin{minipage}[c]{0.7\textwidth}
\captionsetup{font=small}
\caption{Impact of the probability $p$ of background-aware block-wise random synthesis. As discussed in Section~\ref{subsec:subsec_dropout}, if we interpret the proposed background-ware block-wise random synthesis as dropout, Guided CTPN is a kind of approximation of Guided CTPN+ with ignoring feature maps scaled by $p$ in background. as $p$ increases from 0 to 1, the approximation error  becomes unignorable. Therefore, Guided CTPN achieves best trade-off at a smaller $p$ (i.e.,  $p=0.4$).}
\label{fig:fig_noise}
\end{minipage}
\vspace{-4mm}
\end{figure}

% \begin{figure}
% \begin{minipage}{0.5\linewidth}
% \centering
% \centerline{ \includegraphics[width =0.55\textwidth]{images/fig_guidance_noise.eps}}
% %\subcaption{  \label{figa} }
% \end{minipage}
% \caption{Impact of the probability $p$ of background-aware block-wise random synthesis.}
% \label{fig:fig_noise}
% \end{figure}
\begin{table*}[]
\begin{minipage}{1\linewidth}
\begin{minipage}{0.5\linewidth}
\centering
%\footnotesize
\resizebox{2\columnwidth}{!}{
\begin{tabular}{l || l|l|l|l|| l|l|l|l|| l|l|l|l}
\hline
\multirow{2}{*}{Method}  & \multicolumn{4}{c||}{SWT} & \multicolumn{4}{c||}{ICDAR 2011} &\multicolumn{4}{c}{ICDAR 2013}    \\ \cline{2-13} 
                        &Recall&Precision& F-measure&Speedup&Recall&Precision& F-measure&Speedup&Recall& Precision &F-measure&Speedup \\ \hline \hline
CTPN &0.635& 0.688&0.660&$\times 1$ &0.858&    0.867&    0.862 &$\times 1$&  0.861&    0.899&     0.880 & $\times 1$\\ \hline 
Guided CTPN& \textbf{0.669}&0.664&0.667&$\bm{\times 4.2}$& 0.868&     0.893&    0.880&$\bm{\times 2.9}$&0.874& 0.916&0.895& $\bm{\times 2.9} $\\ 
Guided CTPN+&0.660&\textbf{0.691}&\textbf{0.675}&$\times 0.7$&\textbf{0.875}&    \textbf{0.906}&    \textbf{0.890} &$\times 0.7$ &\textbf{0.880}& \textbf{0.924}& \textbf{0.901}&$\times 0.7$ \\ \hline
\end{tabular}{}
}
\end{minipage}
\end{minipage}
\begin{minipage}{1\linewidth}
\captionsetup{font=small}
\caption{Effectiveness of guidance on SWT, ICDAR 2011 and ICDAR 2013.}
\label{tab:guided_cnn_vs_CTPN}
\end{minipage}
\vspace{-5mm}
\end{table*}
\begin{table}[!t]
\begin{minipage}[c]{1\textwidth}
\centering
\begin{minipage}[c]{0.5\textwidth}
%\footnotesize
\resizebox{\columnwidth}{!}{
\begin{tabular}{l || l|l|l|l}
\hline
Method  & Recall&Precision& F-measure&Speedup \\ \hline \hline
EAST &0.773&0.847&0.808&$\times 1$ \\ \hline 
%78.91% 	84.83% 	81.77%
Guided EAST &0.789&0.848 &0.818&$\bm{\times 2.0$} \\ 
%78.87% 	86.63% 	82.30%
Guided EAST+&\textbf{0.789}&\textbf{0.866}&\textbf{0.823}&${\times 0.8}$ \\ \hline
\end{tabular}{}
}
\end{minipage}
\end{minipage}
\begin{minipage}[c]{1\textwidth}
\captionsetup{font=small}
\caption{Effectiveness of guidance on ICDAR 2015. For fair comparison, we compare Guided EAST and Guided EAST+ with the EAST result re-implemented by the third party.}
\label{tab:guided cnn_vs_east}
\end{minipage}
\vspace{-5mm}
\end{table}

When $\tau$ is set to $0.1$, $0.2$, $0.3$ and $0.4$, the guidance subnetwork obtains  the recall of $0.97$, $0.95$, $0.94$, and $0.92$ while 
Guided CTPN achieves speedup by $2.69$, $2.90$, $3.01$, and $3.12$ times respectively. We fix $\tau$ to $0.2$ for all rest experiments since it obtains good trade-off between recall and speedup. Figure~\ref{fig:fig_noise} shows the performance of Guided CTPN and Guided CTPN+ with different $p$. 
It has been shown that Guided CTPN+ obtains its best result when $p=0.8$
while Guided CTPN obtains its best result when $p=0.4$. 
%As discussed in Section~\ref{subsec:subsec_dropout}, if we interpret the proposed background-ware block-wise random synthesis as dropout, Guided CTPN is a kind of approximation of Guided CTPN+ with ignoring feature maps scaled by $p$ in background. as $p$ increases from 0 to 1, the approximation error  becomes unignorable. Therefore, Guided CTPN achieves best trade-off at a smaller $p$ (i.e.,  $p=0.4$).
In the rest experiments, we report results of Guided CNN with $p=0.4$ while those of Guided CNN+ with $p=0.8$.

\textbf{The effectiveness of guidance}. We investigate the effectiveness of the guidance.
We employ CTPN and EAST as backbone, 
and evaluate Guided CNNs and their corresponding backbones on benchmark datasets. %SWT, ICDAR 2011, ICDAR 2013 and ICDAR 2015 by the popular online evaluation tool DetEval. CTPN is re-implemented by us.Note that the re-implemented CTPN results are  a little different from but comparable with those in~\cite{Tian2016}. We believe it is because of different training data.

% \begin{table}
% \begin{minipage}[c]{0.5\textwidth}
% \centering
% % \footnotesize
% \resizebox{2\columnwidth}{!}{
% \begin{tabular}{l || l|l|l|l|| l|l|l|l|| l|l|l|l}
% \hline
% \multirow{2}{*}{Method}  & \multicolumn{4}{c||}{SWT} & \multicolumn{4}{c||}{ICDAR 2011} &\multicolumn{4}{c}{ICDAR 2013}    \\ \cline{2-13} 
%                         &Recall&Precision& F-measure&Speedup&Recall&Precision& F-measure&Speedup&Recall& Precision &F-measure&Speedup \\ \hline \hline
% CTPN &0.635& 0.688&0.660&$\times 1$ &0.858&    0.867&    0.862 &$\times 1$&  0.861&    0.899&     0.880 & $\times 1$\\ \hline 
% Guided CTPN& \textbf{0.669}&0.664&0.667&$\bm{\times 4.2}$& 0.868&     0.893&    0.880&$\bm{\times 2.9}$&0.874& 0.916&0.895& $\bm{\times 2.9} $\\ 
% Guided CTPN+&0.660&\textbf{0.691}&\textbf{0.675}&$\times 0.7$&\textbf{0.875}&    \textbf{0.906}&    \textbf{0.890} &$\times 0.7$ &\textbf{0.880}& \textbf{0.924}& \textbf{0.901}&$\times 0.7$ \\ \hline
% \end{tabular}{}
% }
% \end{minipage}
% \caption{Effectiveness of guidance on SWT, ICDAR 2011 and ICDAR 2013.}
% \label{tab:guided_cnn_vs_CTPN}
% \end{table}

From Table~\ref{tab:guided_cnn_vs_CTPN}, Guided CTPN consistently outperforms CTPN in terms of the F-measure while achieving considerable speedup. Guided CTPN+ performs best with sacrificing some speed. 
It can be used in the case when computation resource is adequate.
Specifically, on ICDAR 2011 and ICDAR 2013, compared with CTPN, Guided CTPN improves the F-measure 
by $1.5\sim1.8\%$ while speeding up $2.9$ times. 
Guided CTPN+ improves the F-measure by $2.1\sim 2.8\%$ at the cost of $0.3$ times slowing down. 
On SWT, Guided CTPN and Guided CTPN+ achieve $0.7\%$ and $1.5\%$ F-measure gain respectively, although its ground truth
is annotated in text line level which is different from that of our training data (\eg, training set of ICDAR 2013).
Guided CTPN eventually speeds up 4.2 times.
Surprisingly, Guided CTPN obtains a higher recall rate although it filters out most of regions of input image, 
we believe that it is because  the guidance mechanism makes the text detector to 
focus on distinguishing between hard non-text and text regions, and bounding box regression, leading
to better training and generalization. We also analyze the runtime for each component of Guided CTPN on ICDAR 2013. The guidance subnetwork accounts for 15\% of the runtime while the primary text detector 85\%.

% \begin{table}[]
% \centering
% %\footnotesize
% \caption{Effectiveness of guidance on ICDAR 2015. For fair comparison, we compare Guided EAST and Guided EAST+ with the EAST result re-implemented by the third party.}
% \label{tab:guided cnn_vs_east}
% \resizebox{\columnwidth}{!}{
% \begin{tabular}{l || l|l|l|l}
% \hline
% Method  & Recall&Precision& F-measure&Speedup \\ \hline \hline
% EAST &0.773&0.847&0.808&$\times 1$ \\ \hline 
% %78.91% 	84.83% 	81.77%
% Guided EAST &0.789&0.848 &0.818&$\bm{\times 2.0$} \\ 
% %78.87% 	86.63% 	82.30%
% Guided EAST+&\textbf{0.789}&\textbf{0.866}&\textbf{0.823}&${\times 0.8}$ \\ \hline
% \end{tabular}{}
% }
% \end{table}

From Table~\ref{tab:guided cnn_vs_east}, it has been shown that Guided EAST improves the F-measure by 1.0\% while speeding up $2.0$ times compared with EAST.
Guided EAST+ improves the F-measure by $1.5\%$ while slowing down 20\%.

\begin{table}[!t]
% \begin{minipage}[c]{1\textwidth}
\centering
\begin{minipage}{1\textwidth}
% \centering
\resizebox{\columnwidth}{!}{
\begin{tabular}{l|l|l|l| l|l|l | l|c|c|c}
\hline
\multicolumn{7}{c}{ICDAR 2013} \vline & \multicolumn{4}{c}{ICDAR 2015} \\
\hline
 \multirow{2}{*}{Method}& \multicolumn{3}{c}{IC13 Eval} \vline &\multicolumn{3}{c}{DetEval} \vline & \multirow{2}{*}{Method} & \multirow{2}{*}{R} & \multirow{2}{*}{P} & \multirow{2}{*}{F}\\ 
  \cline{2-7} &R&P&F&R&P&F &  &  &  &  \\ \hline \hline
%R2CNN~\cite{Jiang2017}&-&-&-&0.826& 0.936 &  0.877  \\ 
Jiang~\cite{Jiang2017a}&-&-&-&\textbf{0.915}&0.922& \textbf{0.919} & SSTD~\cite{He2017a} &0.739&0.802&0.769\\ 
WordSup~\cite{Hu2017}&-&-&-&0.875&0.933&0.903 & SegLink~\cite{Shi2017a}&0.768 &0.731 &0.750\\
SegLink~\cite{Shi2017a}&-& -& -& 0.830&0.877&    0.853 & WordSup~\cite{Hu2017}&0.770&0.793&0.782 \\
Yao~\cite{Yao2016}&0.802&0.889&    0.843&-&-&- & R2CNN*~\cite{Jiang2017} &0.743 &0.764 & 0.753 \\
Zhang~\cite{Zhang2016}&   0.780&0.880&     0.830&-&-&- & DMPNet~\cite{Liu2017} &0.682 &0.732 & 0.706 \\

He~\cite{He2017b}&0.810&0.920 & 0.860&-&-&- & RRPN~\cite{Ma2017} &0.770 &0.840 & 0.800 \\
TextBoxes~\cite{Liao2017}&0.740& 0.860& 0.800& 0.740& 0.880& 0.810 & Original EAST*~\cite{Zhou2017} &0.735&0.836&  0.782 \\ 
SSTD~\cite{He2017a}&\textbf{0.857}& 0.884& \textbf{0.870}&0.862 & 0.893&0.877 & & & & \\
Original CTPN~\cite{Tian2016}&0.737& \textbf{0.928}& 0.822&0.830 & 0.930 & 0.877 & & & & \\ \hline 

Guided CTPN&0.846& 0.881 & 0.863 &0.874&	0.916&	0.895& Guided EAST&\textbf{0.789} &0.848 &0.818 \\ 
Guided CTPN+&0.846 &0.885&\textbf{0.870}&0.880&\textbf{0.924}& 0.901& Guided EAST+&\textbf{0.789} &\textbf{0.866}&\textbf{0.823}\\ \hline 
\end{tabular}{}
}
\end{minipage}
% \end{minipage}
\begin{minipage}[c]{1\textwidth}
\captionsetup{font=small}
\caption{Comparisons with state-of-the-art text detection methods on ICDAR 2013 and ICDAR 2015. * indicates the best result with a single scale test.}
\label{tab:com_state_of_the_art_icdar13_icdar15}
\end{minipage}
\vspace{-6mm}
\end{table}
\textbf{Comparisons with state-of-the-art results}. From Table~\ref{tab:com_state_of_the_art_icdar13_icdar15}, it has been shown that Guided CTPN and Guided  CTPN+ are comparable with~\cite{He2017a} and outperform others using the IC13 evaluation protocol. 
% From Table~\ref{tab:com_state_of_the_art_icdar13}, It has been shown that Guided CTPN and Guided  CTPN+ are comparable with~\cite{He2017a} and outperform others using the IC13 evaluation protocol. 
They also outperform others except~\cite{Jiang2017a} using the DetEval evaluation protocol. Note that our Guided CTPN is based on a simple architecture of CTPN and accelerates CTPN significantly while \cite{Jiang2017a} combined a segmentation and detection networks.
Guided EAST and Guided EAST+ outperform others. Especially, Guided EAST+ outperforms the second best method RRPN by 2.3\%.

\section{Conclusion}

We have proposed a general framework for text detection called Guided CNN to improve the accuracy and the speed of existing single forward CNN based text detectors simultaneously. %The Guided CNN consists of one guidance subnetwork, where a coarse guidance mask is predicted from the input image, and one primary text detector, where convolution and non-linear operations are conducted only in the  guidance mask.%
We have designed a context module to capture multi-scale context in the guidance subnetwork, leading  to effective guidance mask predictions. We have proposed a novel background-aware block-wise random synthesis training strategy, resulting in accuracy improvement and convenient deployment. It can be interpreted as a special kind of dropout.
The proposed Guided CNN is a general framework which can be plugged into any existing single forward CNN based text detectors. We have demonstrated its general applicability by instantiating it with CTPN and EAST as backbone.  Extensive experiments have
evaluated its effectiveness and efficiency. %It consistently improves the accuracy of its backbone, while achieving considerable speedup on various text benchmarks. Besides, it is complementary to other general CNN acceleration approaches, and  can be combined with others for further acceleration.

\bibliography{egbib}

\newpage
\appendix 

\section{Supplementary Material --- Boosting up Scene Text Detectors with Guided CNN}
In order to supplement our main submission, we provide additional materials in this document. We show visualization and qualitative examples on horizontal text dataset ICDAR2013, and multi-oriented text dataset ICDAR2015 with various backbones to demonstrate the effectiveness of the proposed Guided CNN and Guided CNN+.

\begin{figure}[htpb]
\begin{minipage}{1\linewidth}
\tabcolsep2.1pt
\renewcommand\arraystretch{1.3}
\begin{tabular}{ccccc}
{\includegraphics[width=0.175\linewidth]{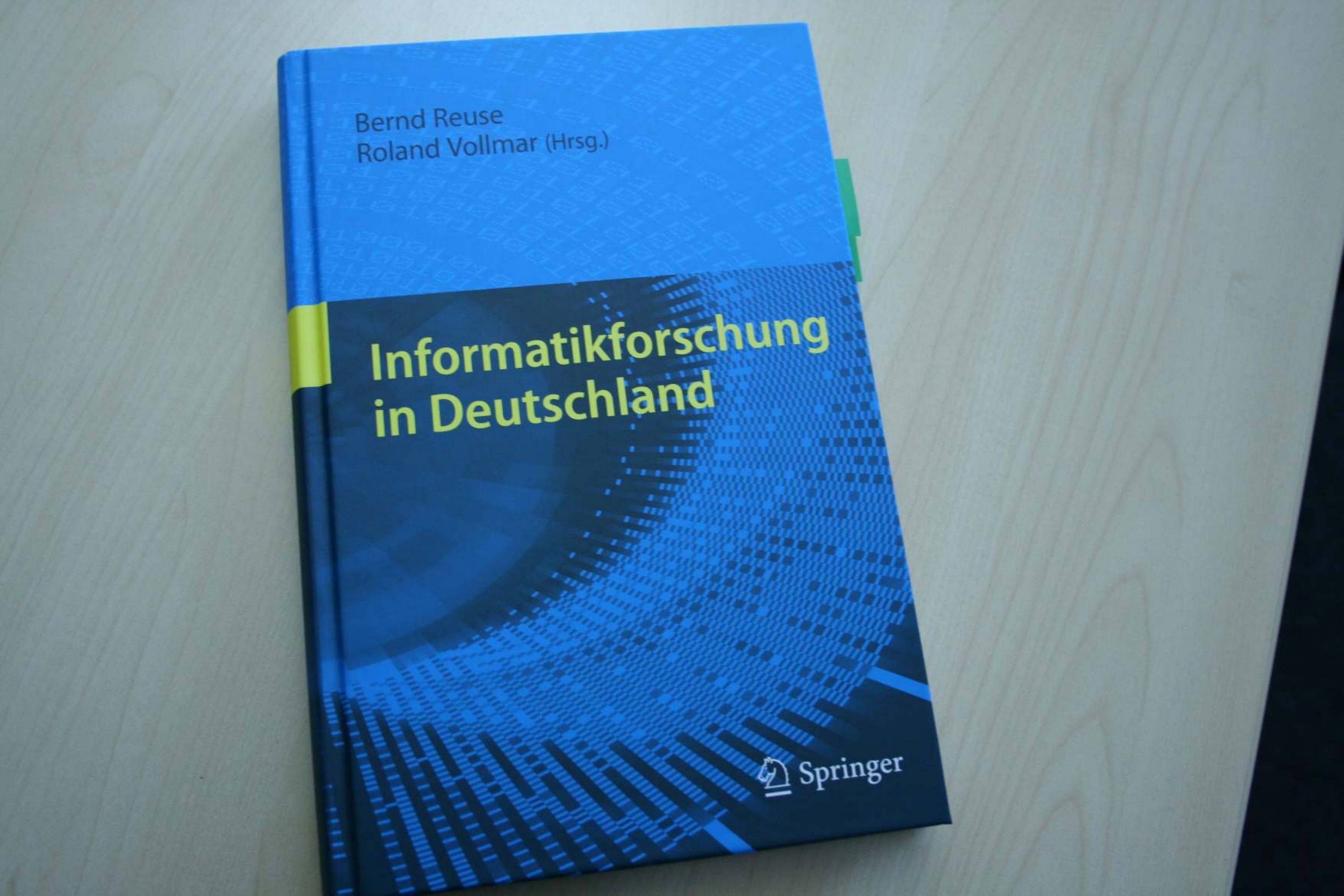}} 
& {\includegraphics[width=0.175\linewidth]{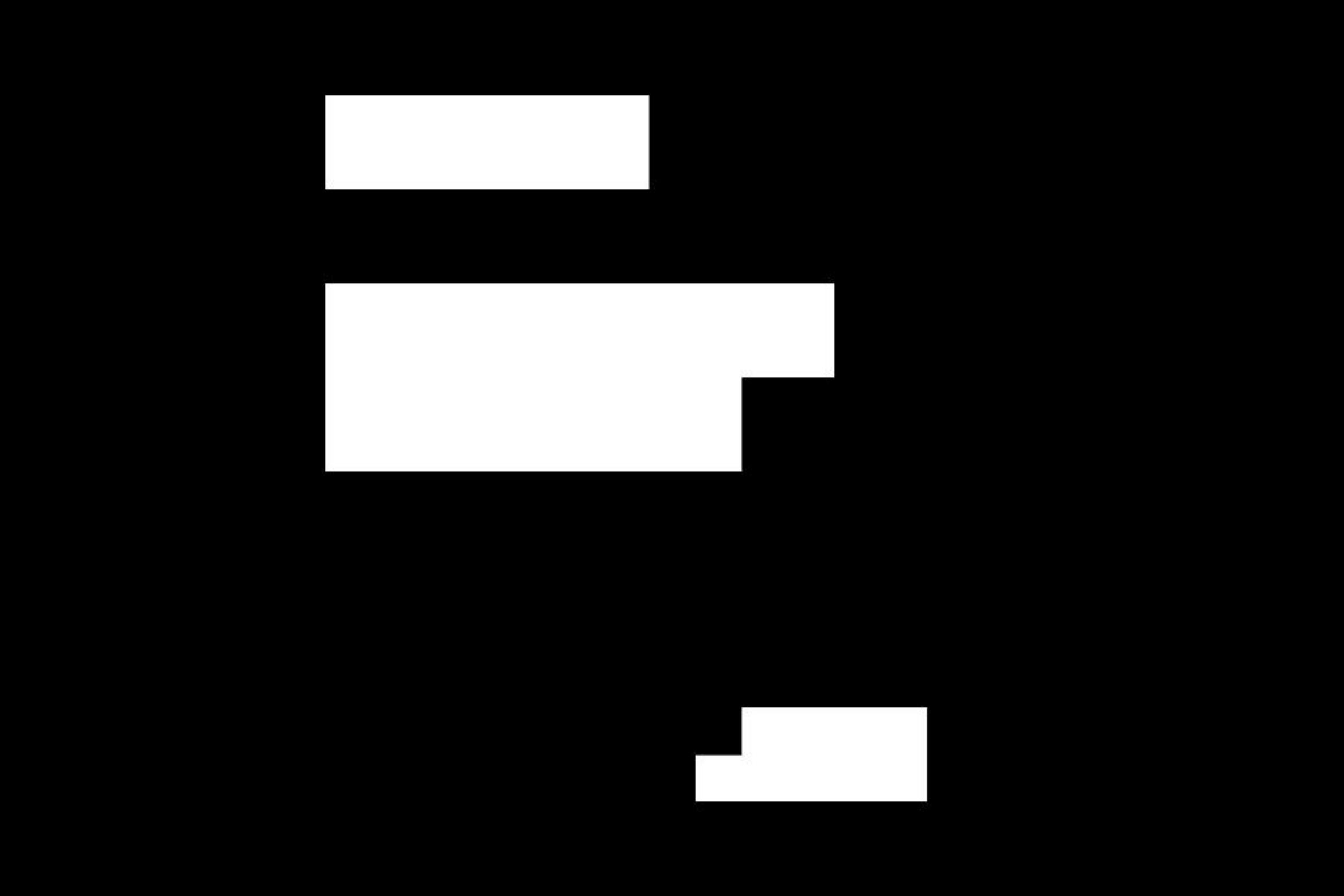}}
& {\includegraphics[width=0.175\linewidth]{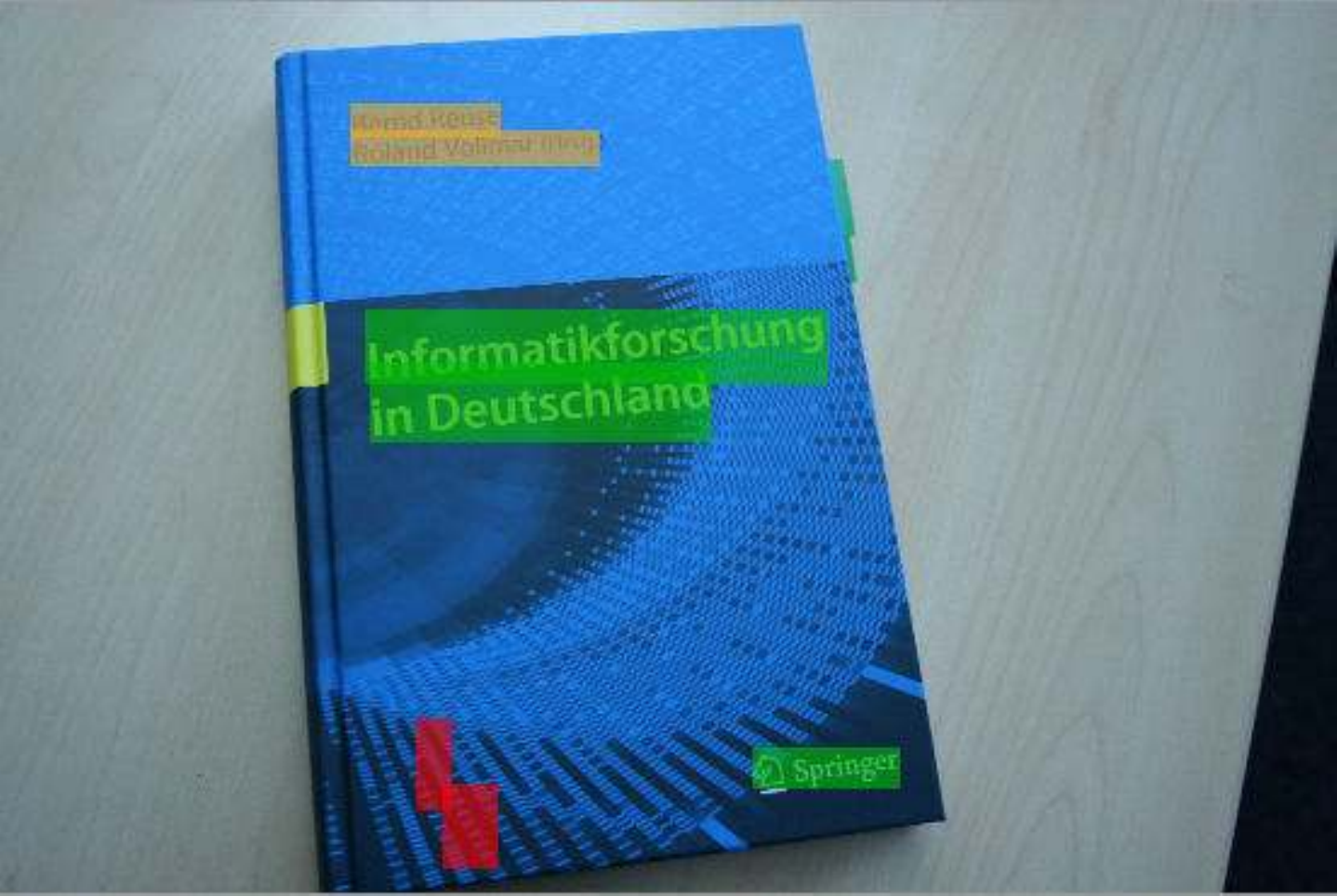}}
& {\includegraphics[width=0.175\linewidth]{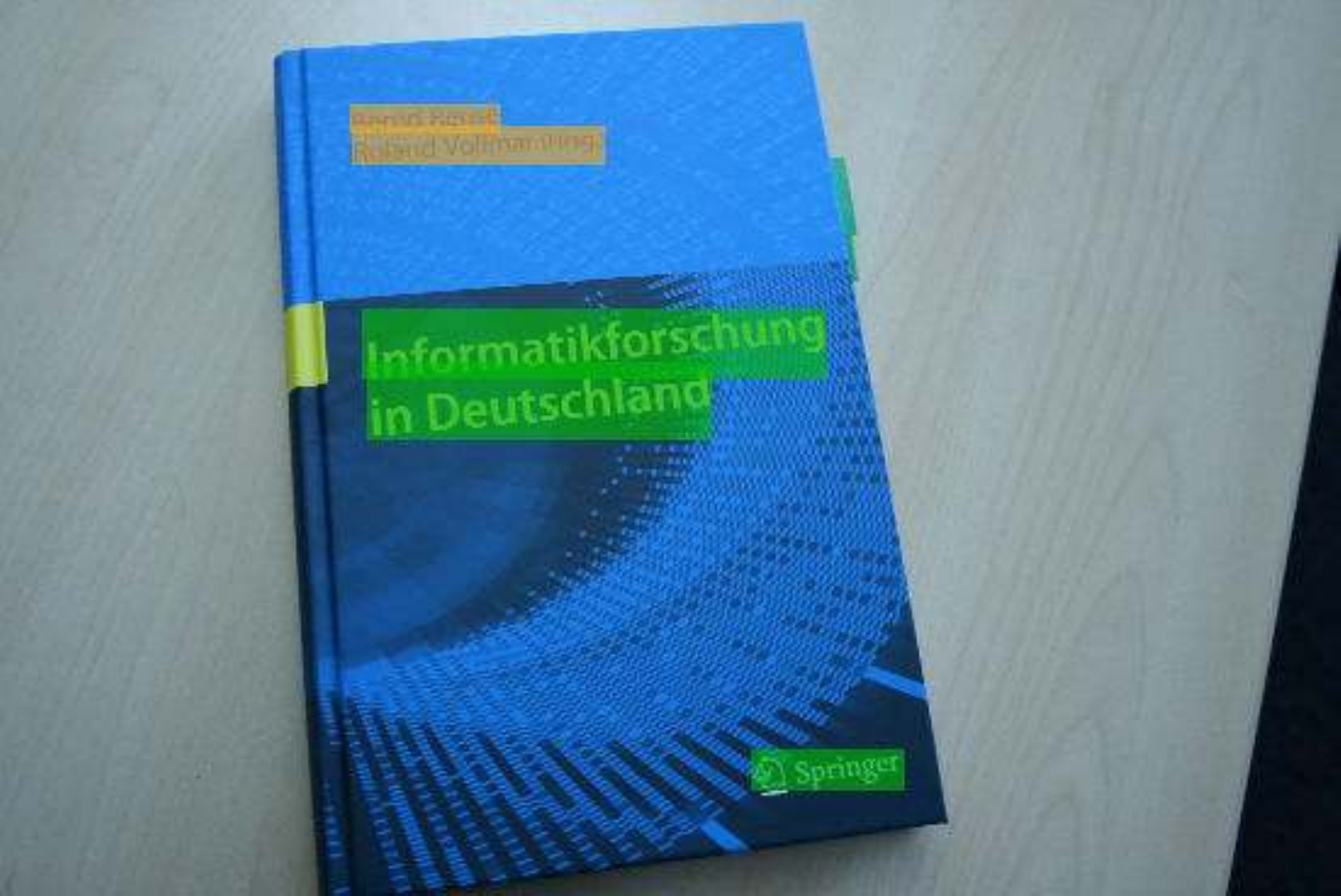}}
& {\includegraphics[width=0.175\linewidth]{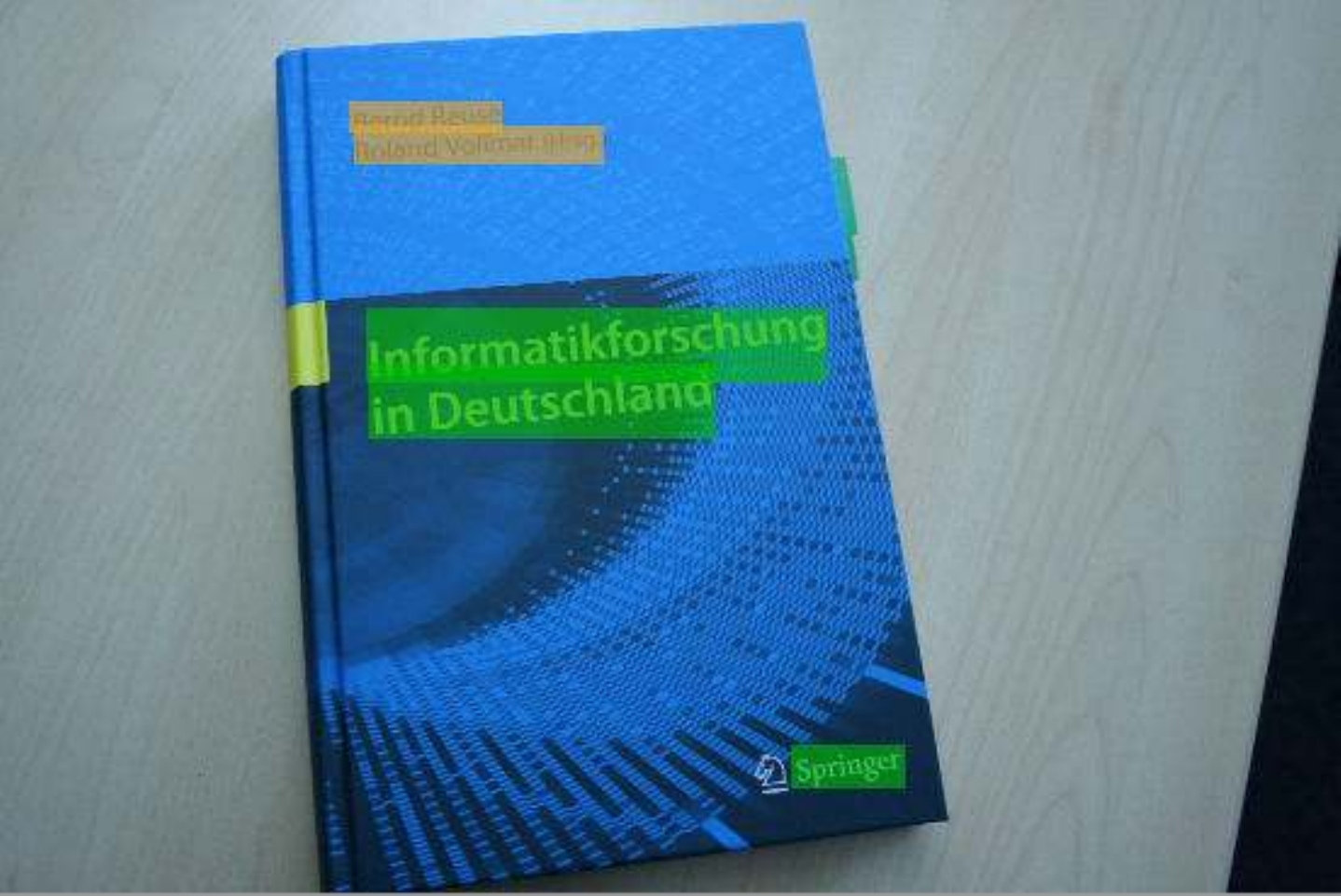}} \\
{\includegraphics[width=0.175\linewidth]{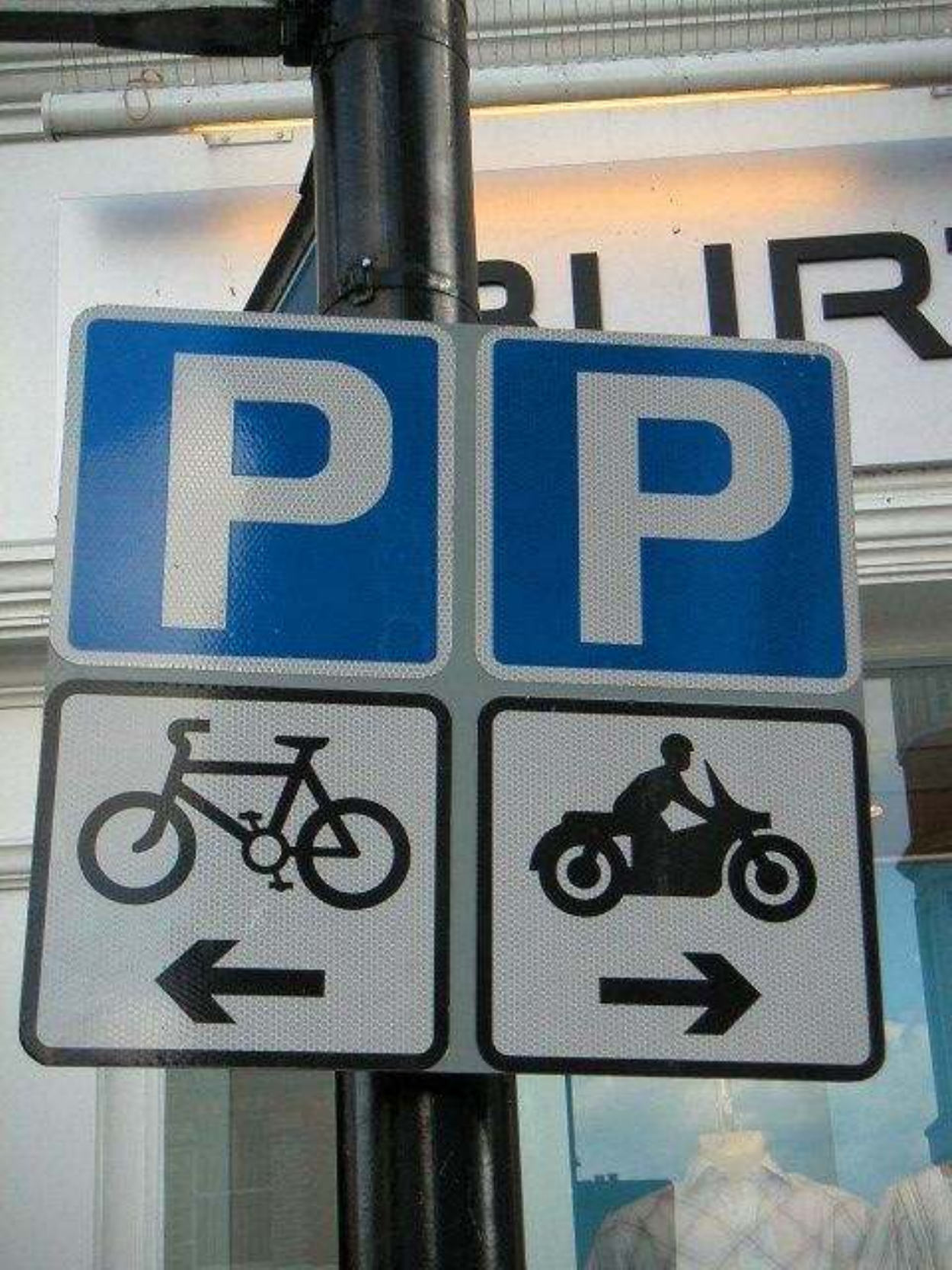}} 
& {\includegraphics[width=0.175\linewidth]{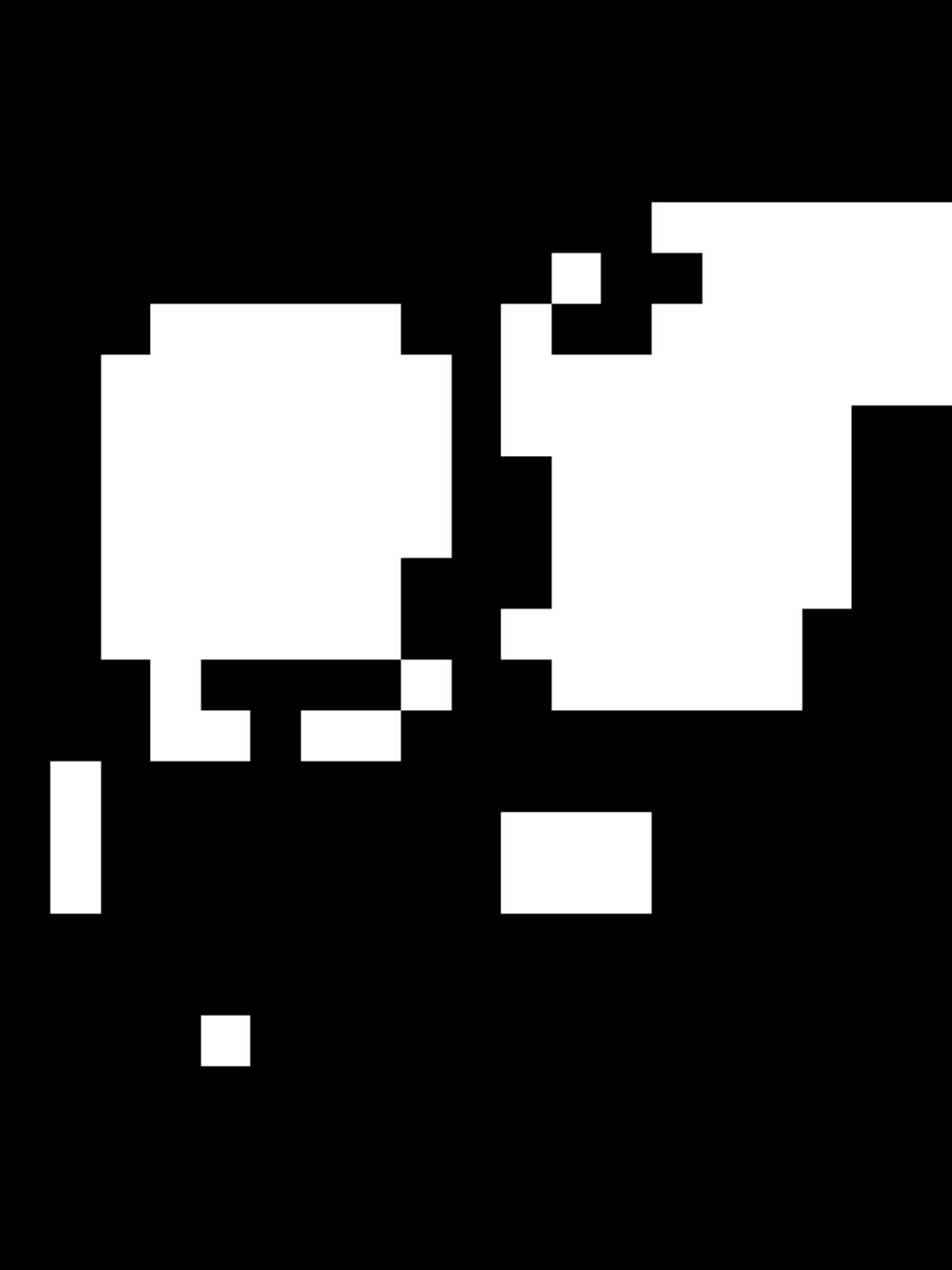}}
& {\includegraphics[width=0.175\linewidth]{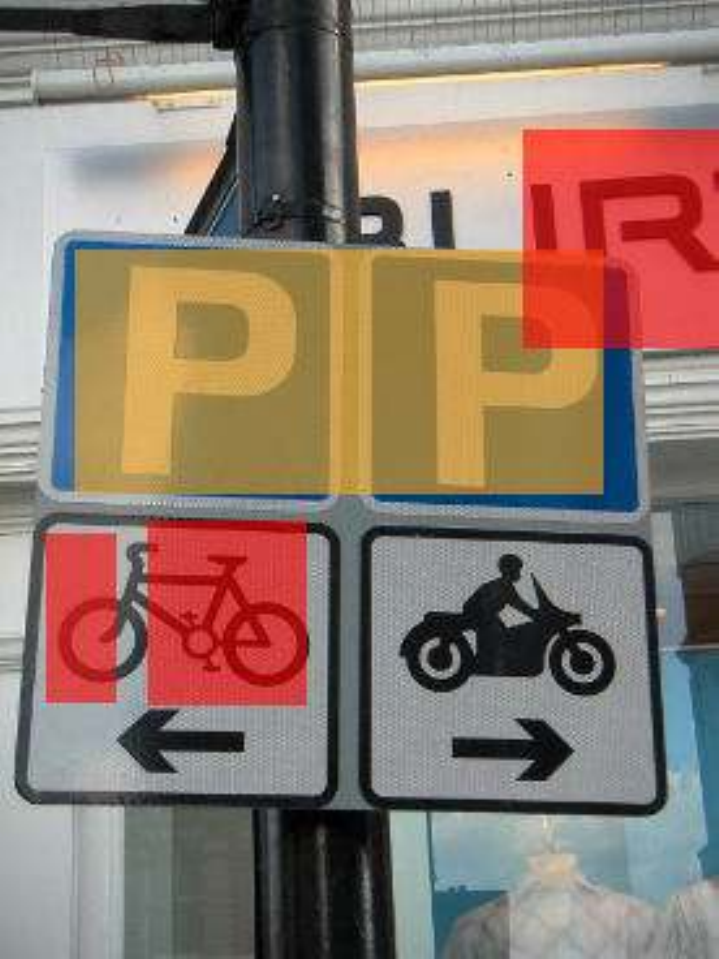}}
& {\includegraphics[width=0.175\linewidth]{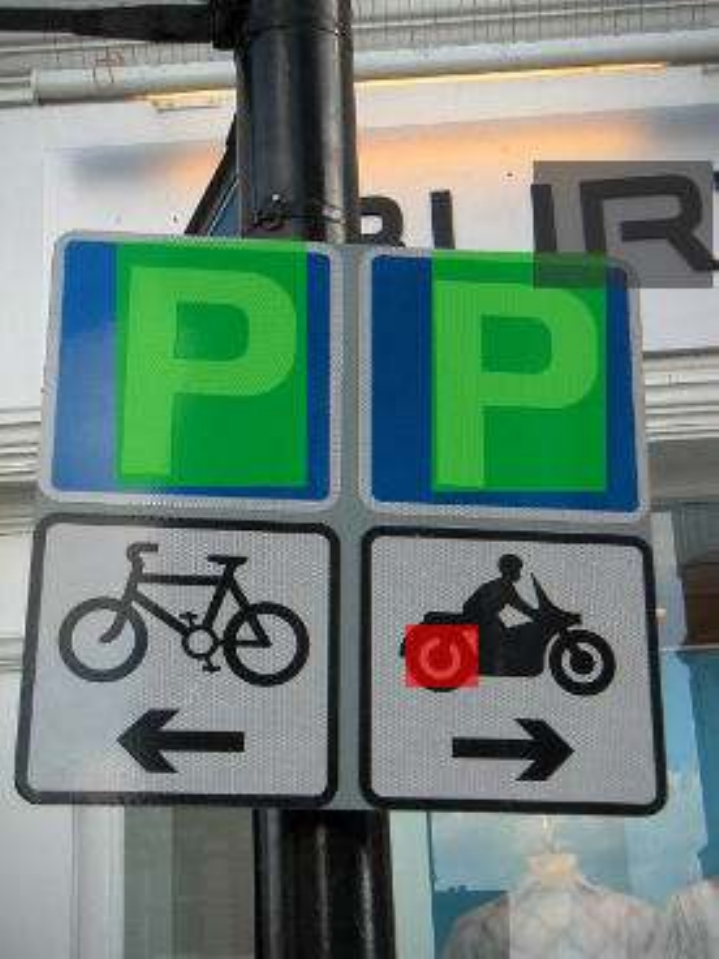}}
& {\includegraphics[width=0.175\linewidth]{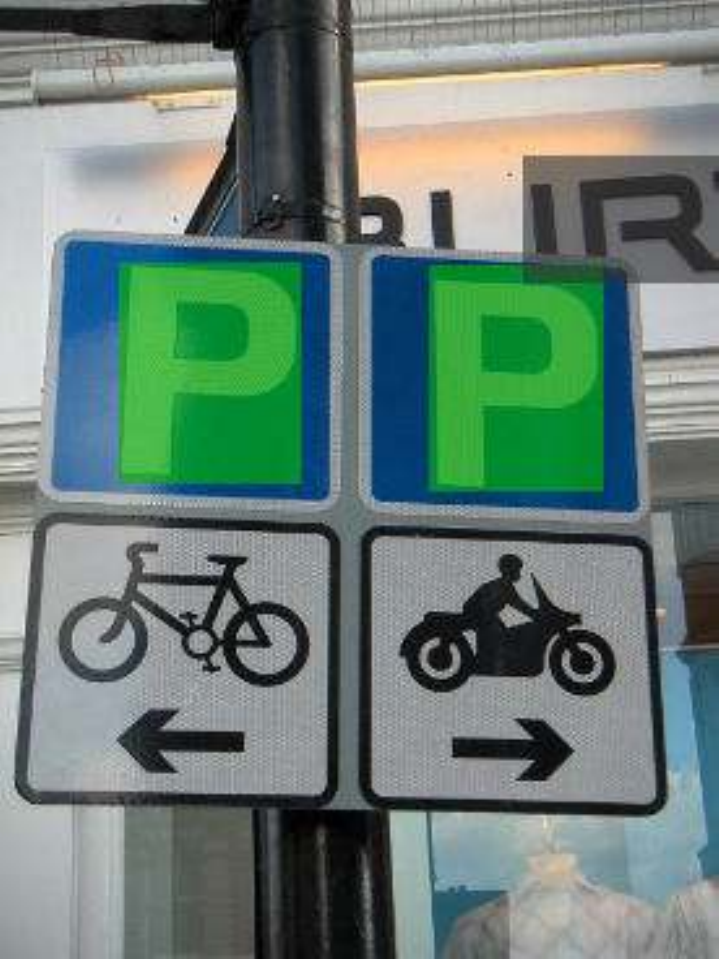}} \\
{\includegraphics[width=0.175\linewidth]{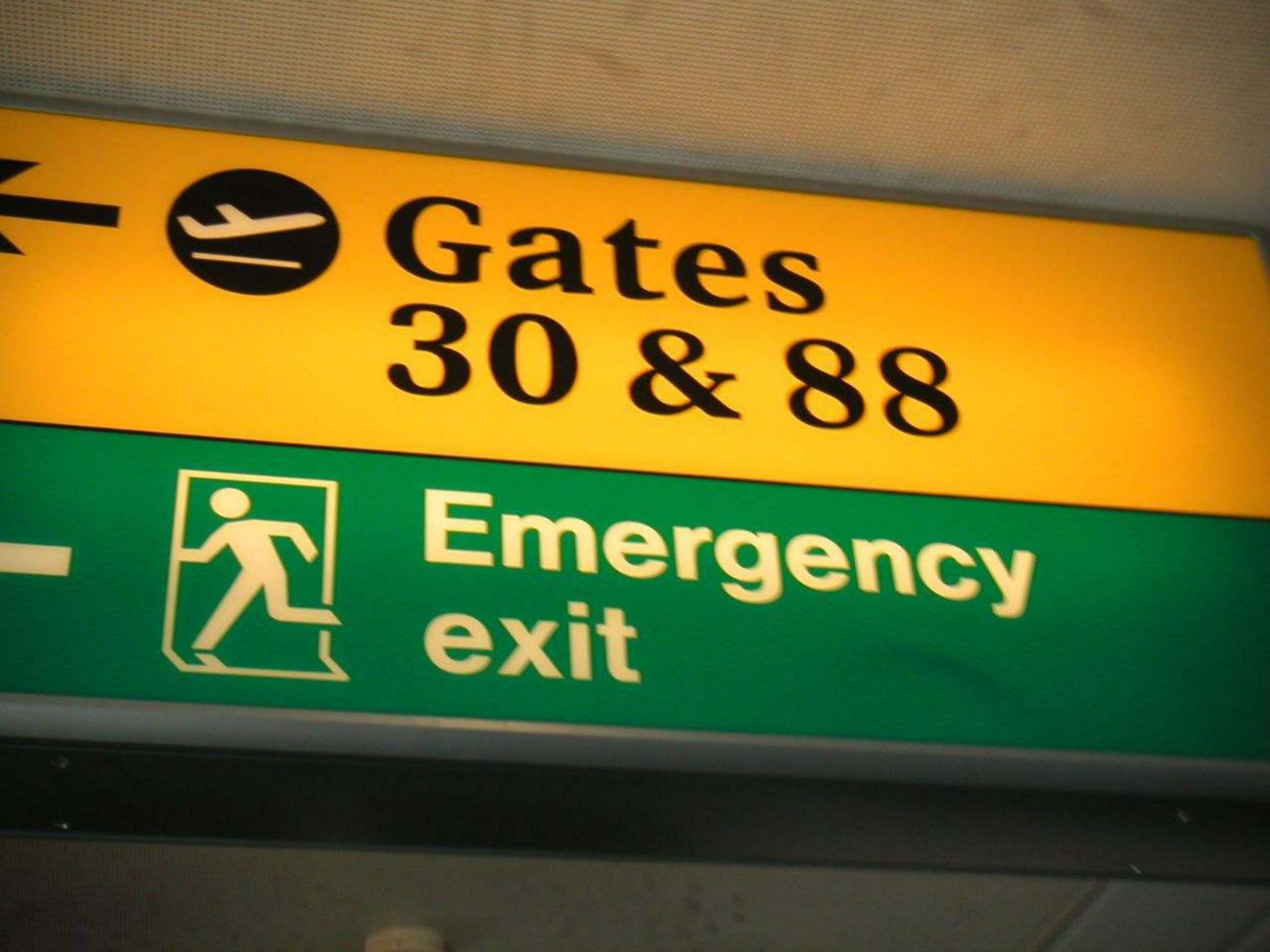}} 
& {\includegraphics[width=0.175\linewidth]{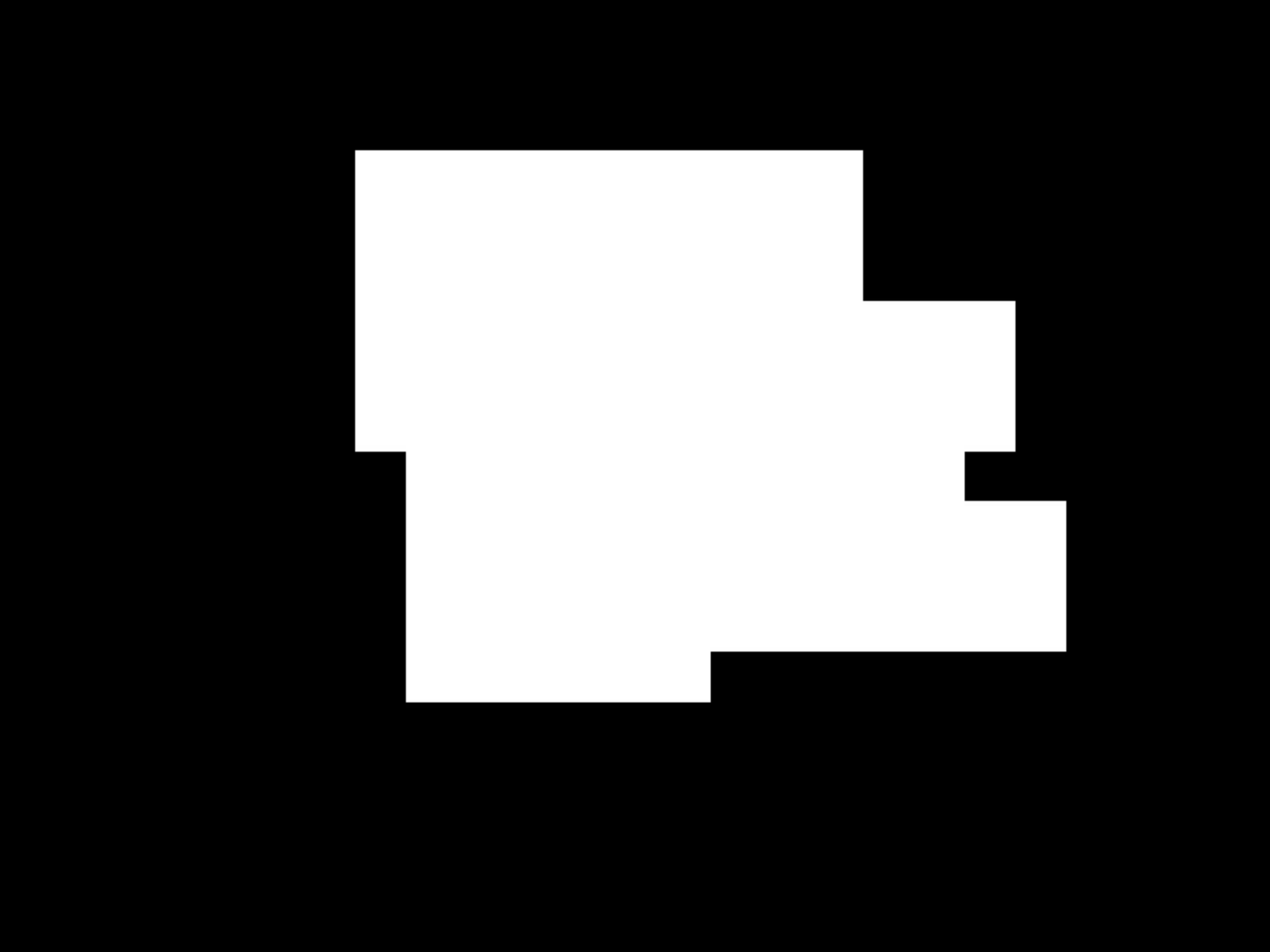}}
& {\includegraphics[width=0.175\linewidth]{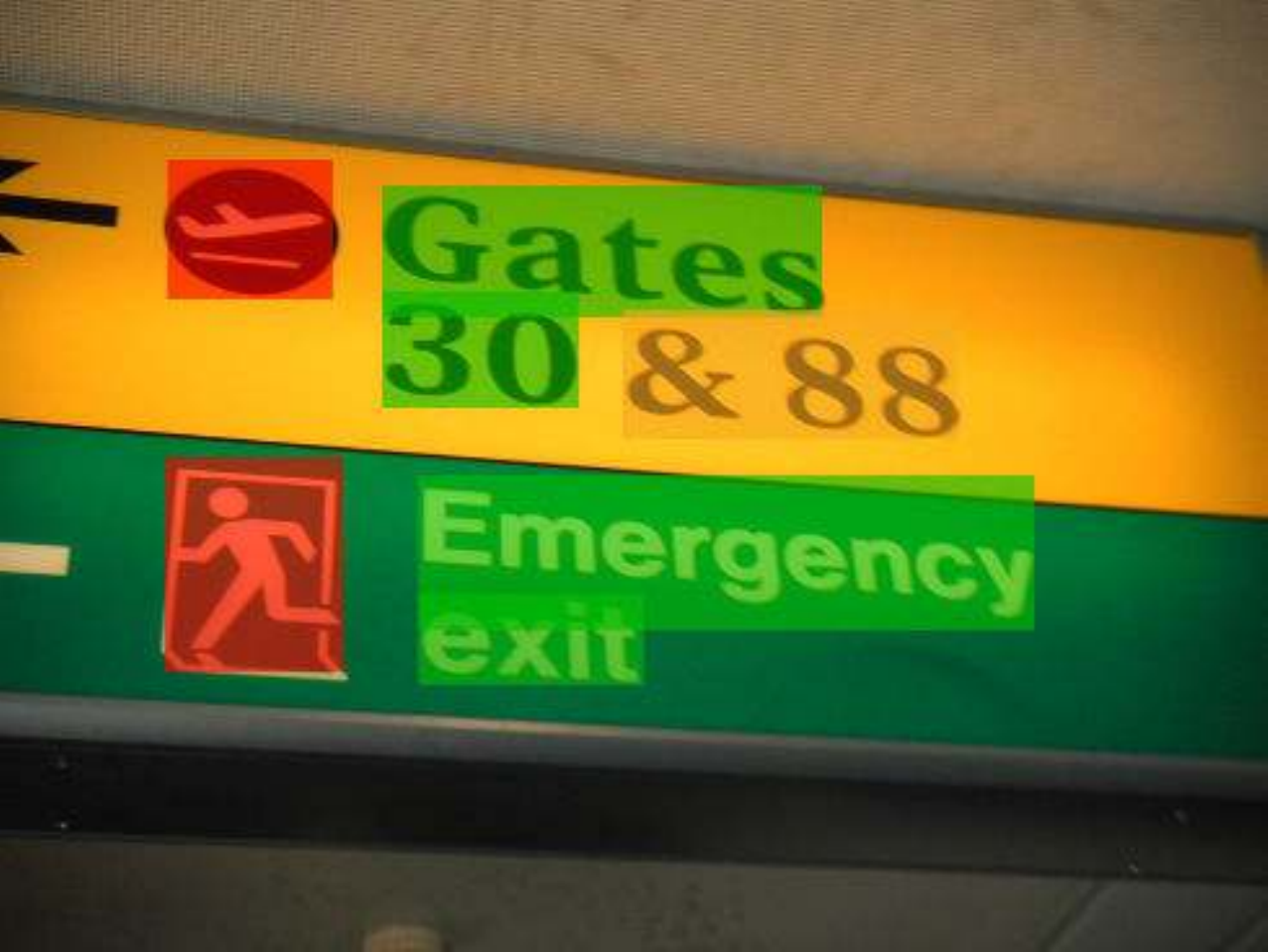}}
& {\includegraphics[width=0.175\linewidth]{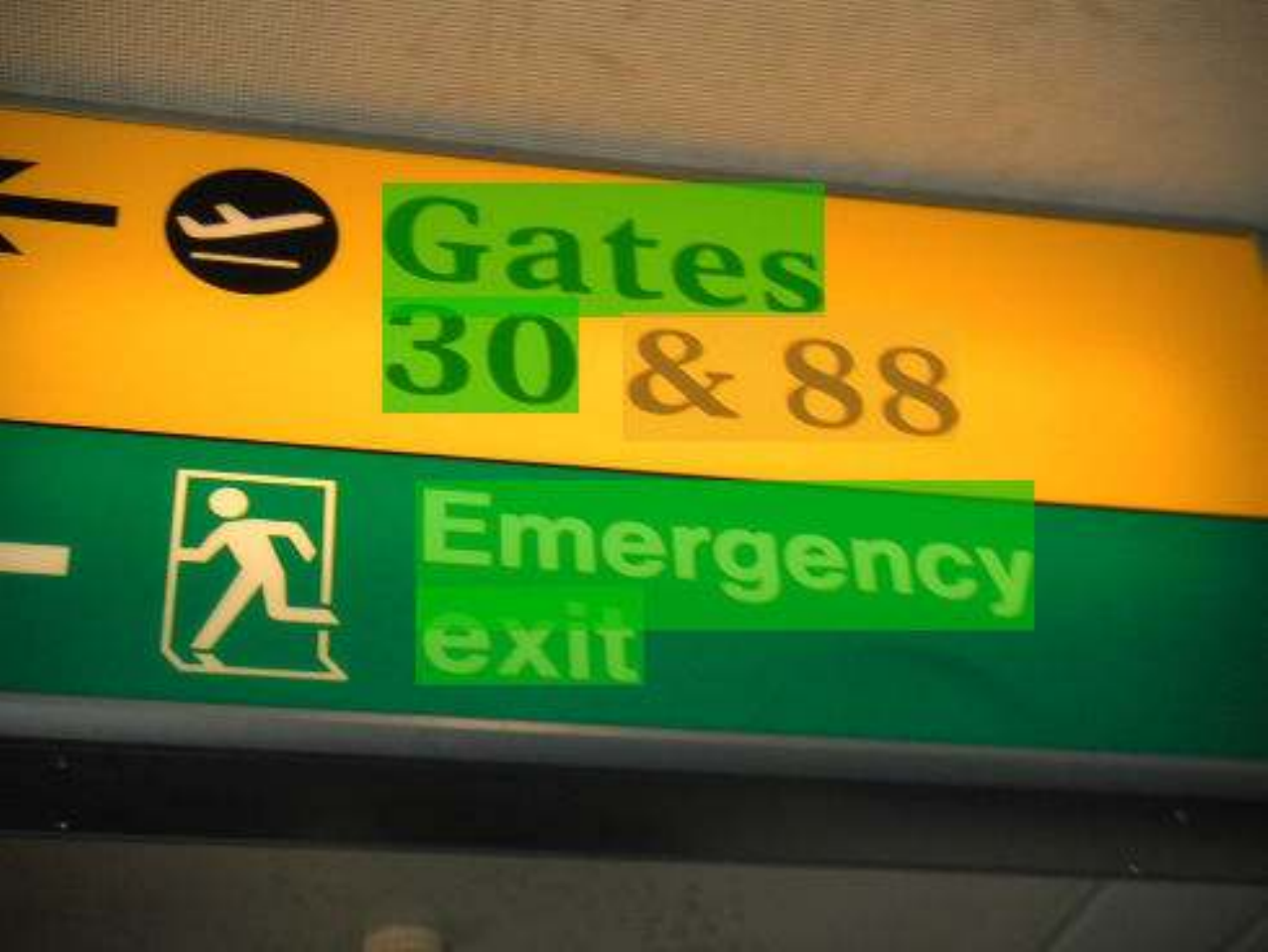}}
& {\includegraphics[width=0.175\linewidth]{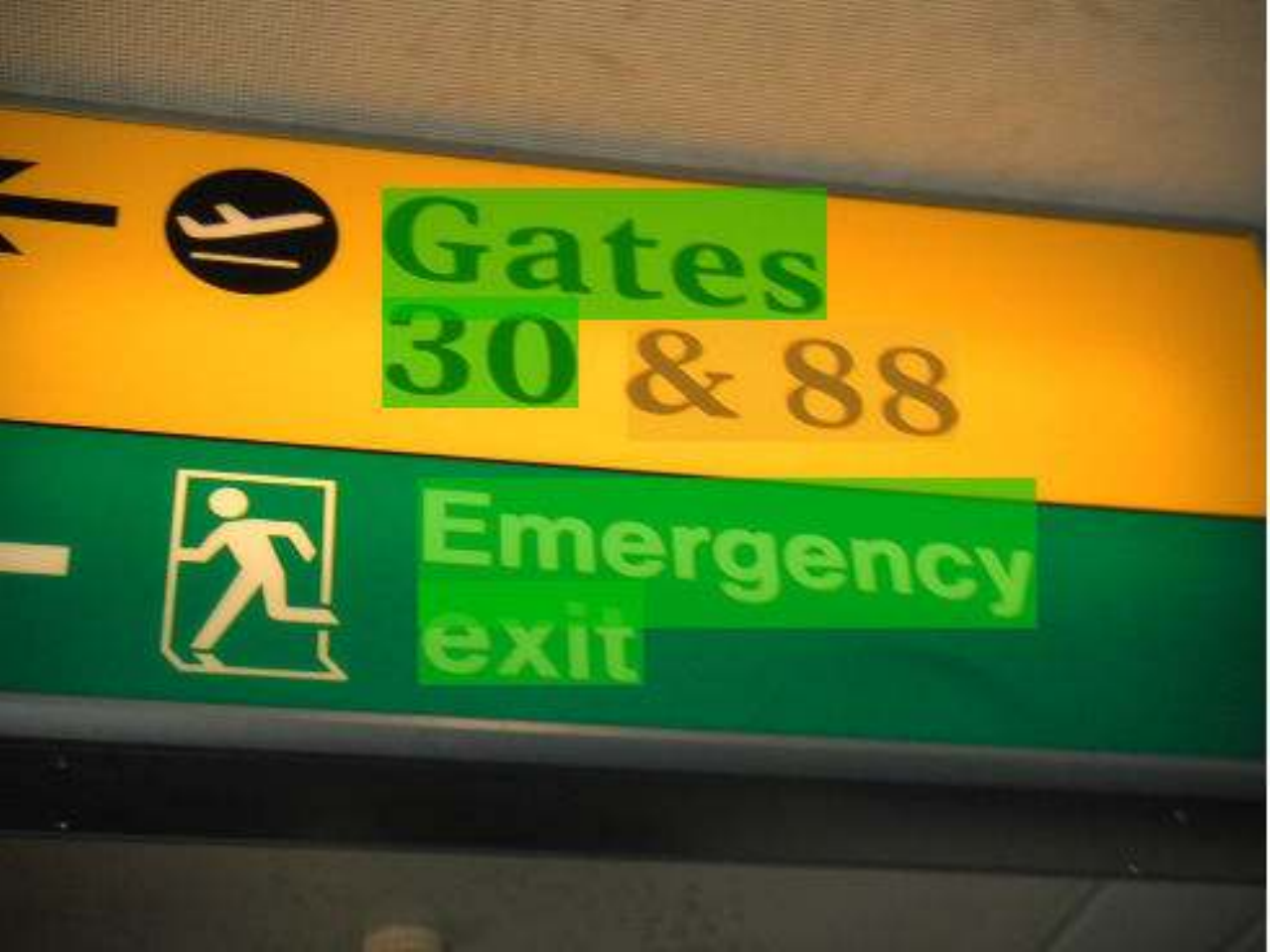}} \\
{\includegraphics[width=0.175\linewidth]{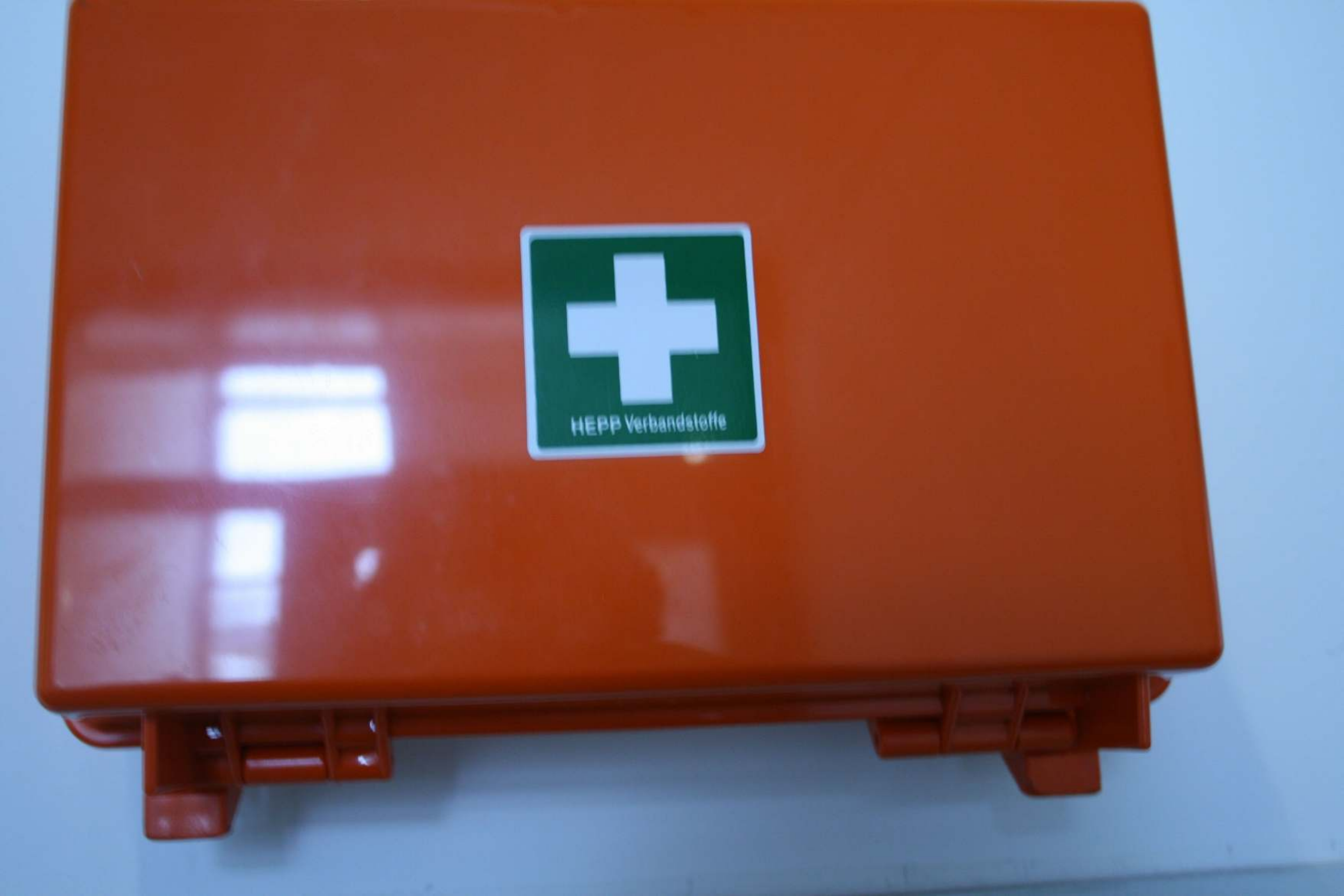}} 
& {\includegraphics[width=0.175\linewidth]{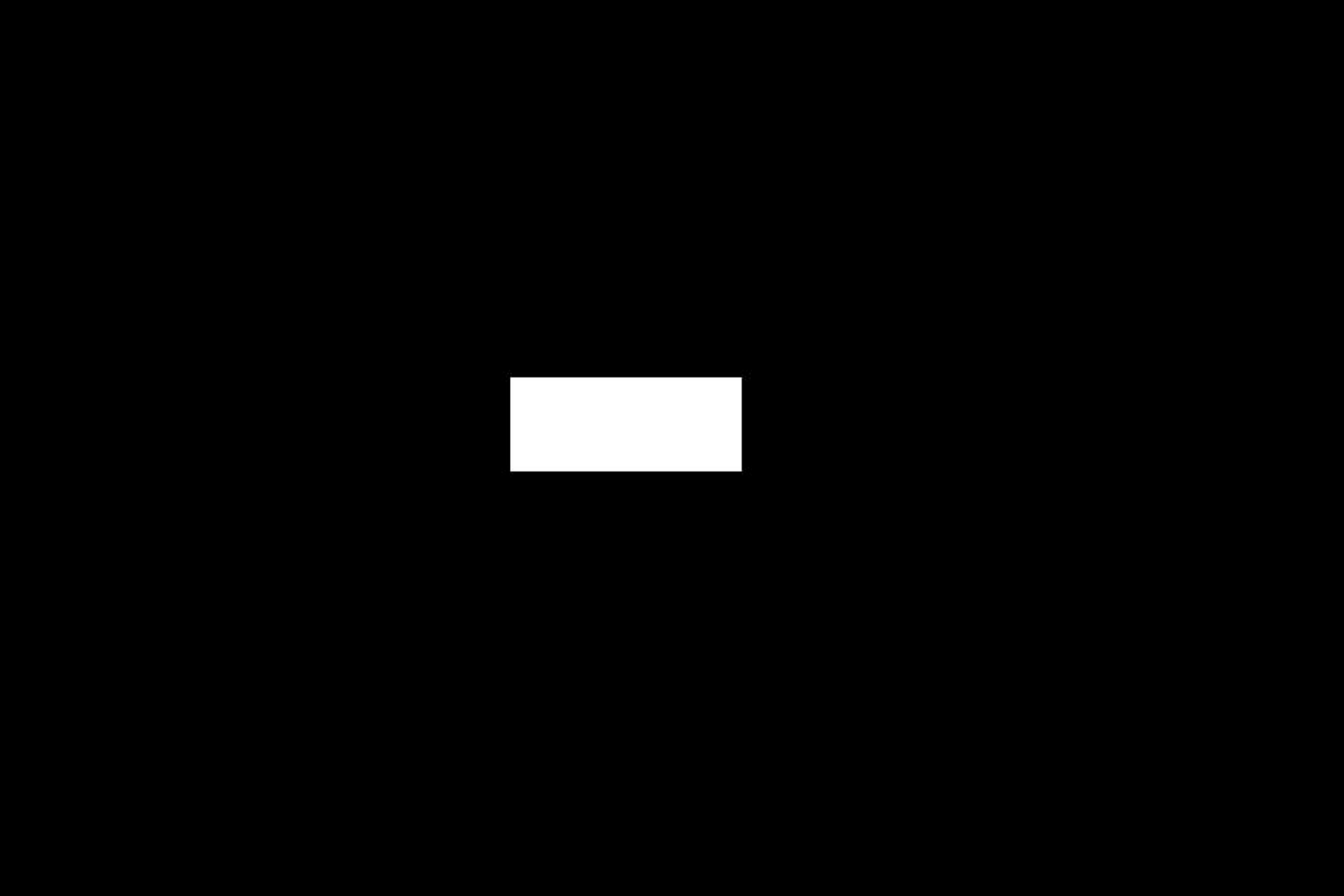}}
& {\includegraphics[width=0.175\linewidth]{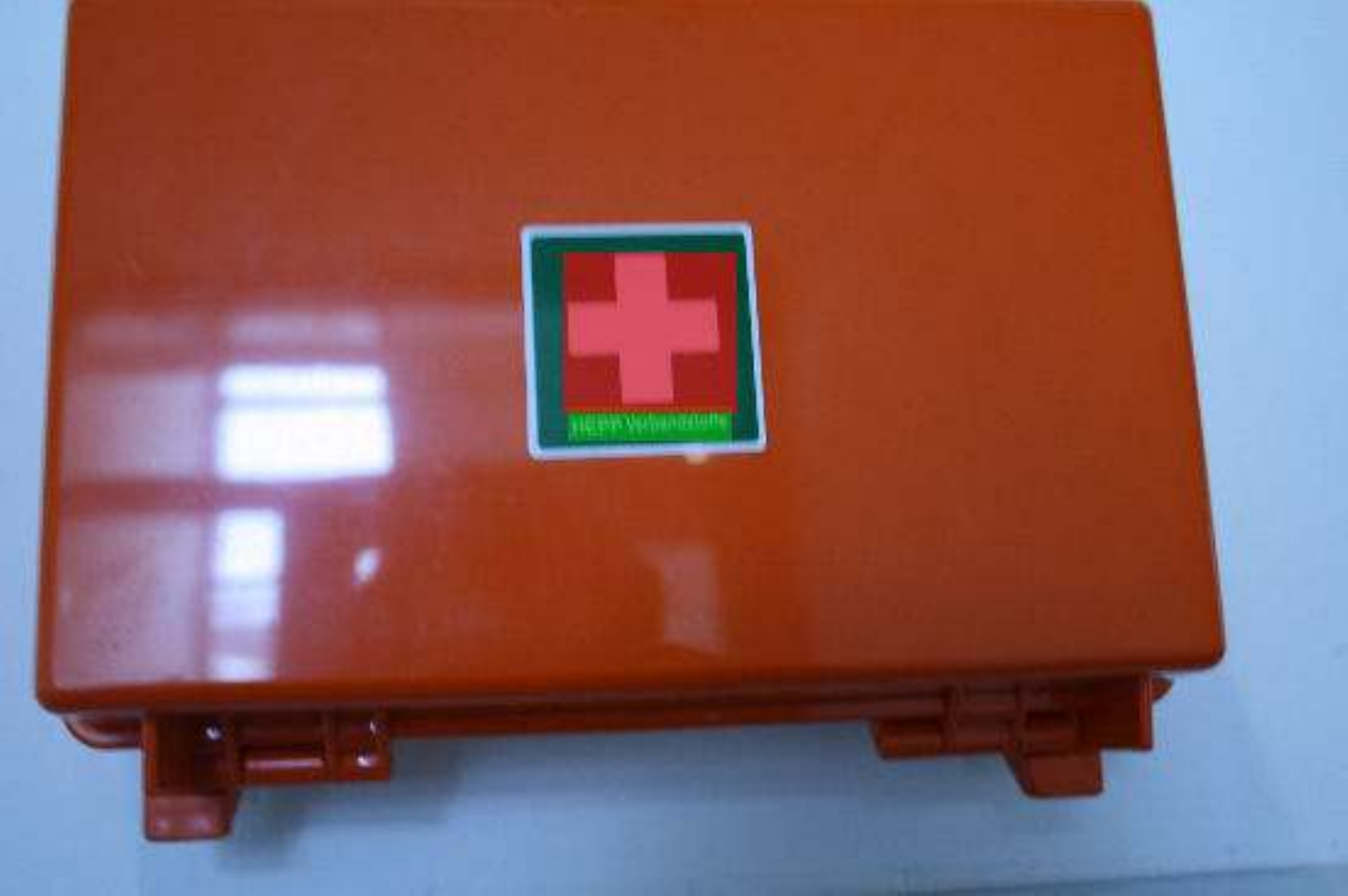}}
& {\includegraphics[width=0.175\linewidth]{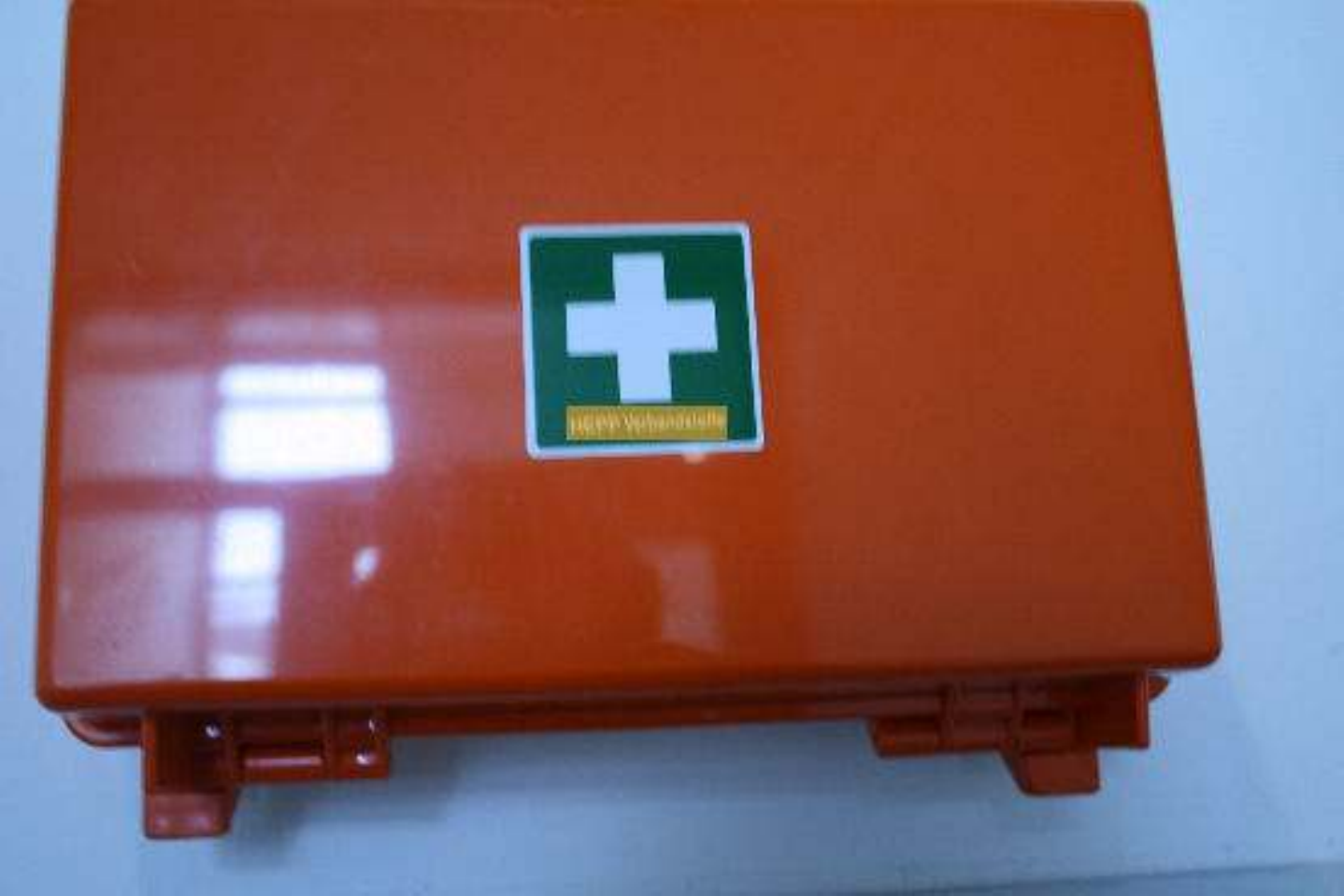}}
& {\includegraphics[width=0.175\linewidth]{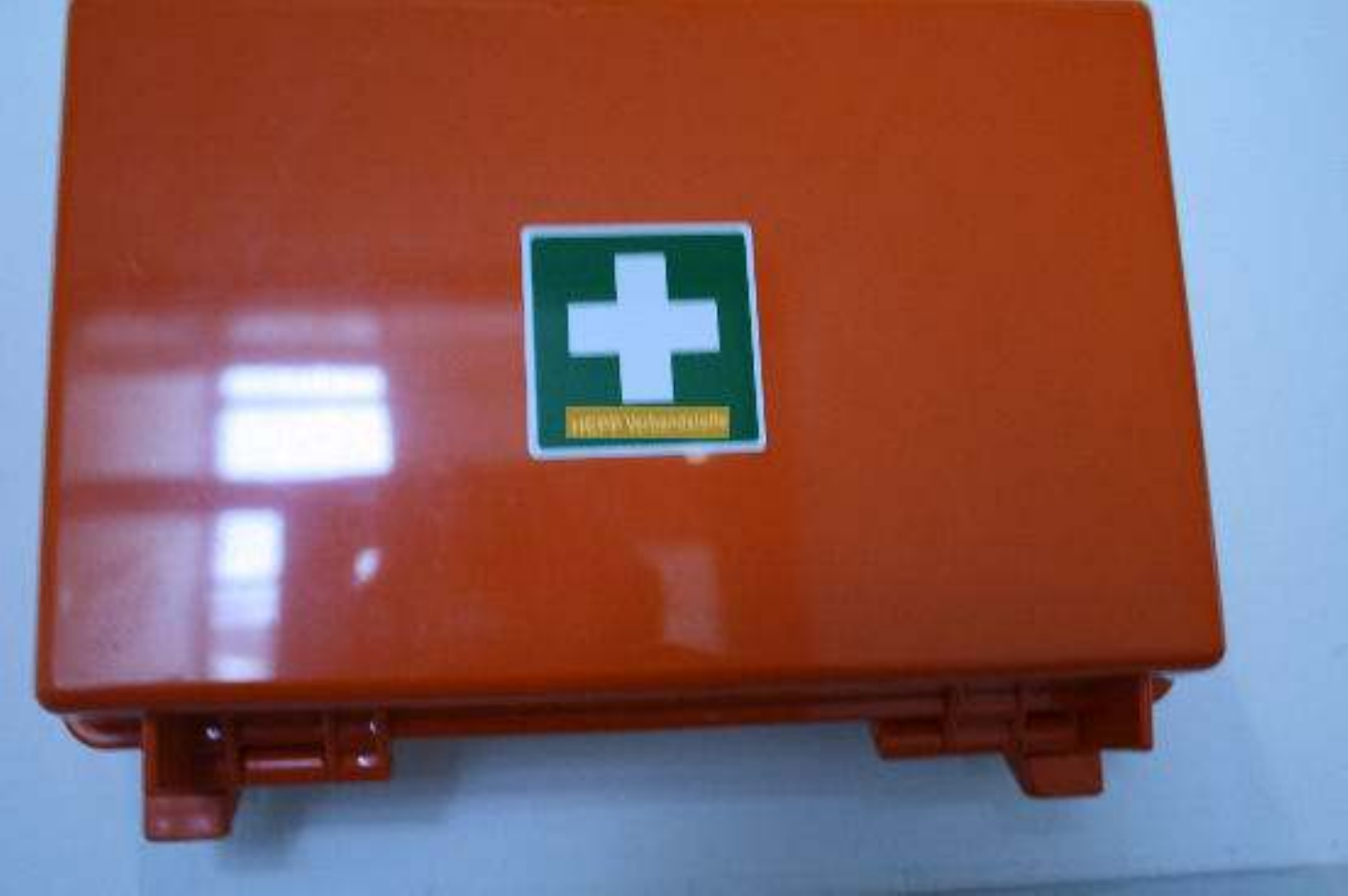}} \\
{\includegraphics[width=0.175\linewidth]{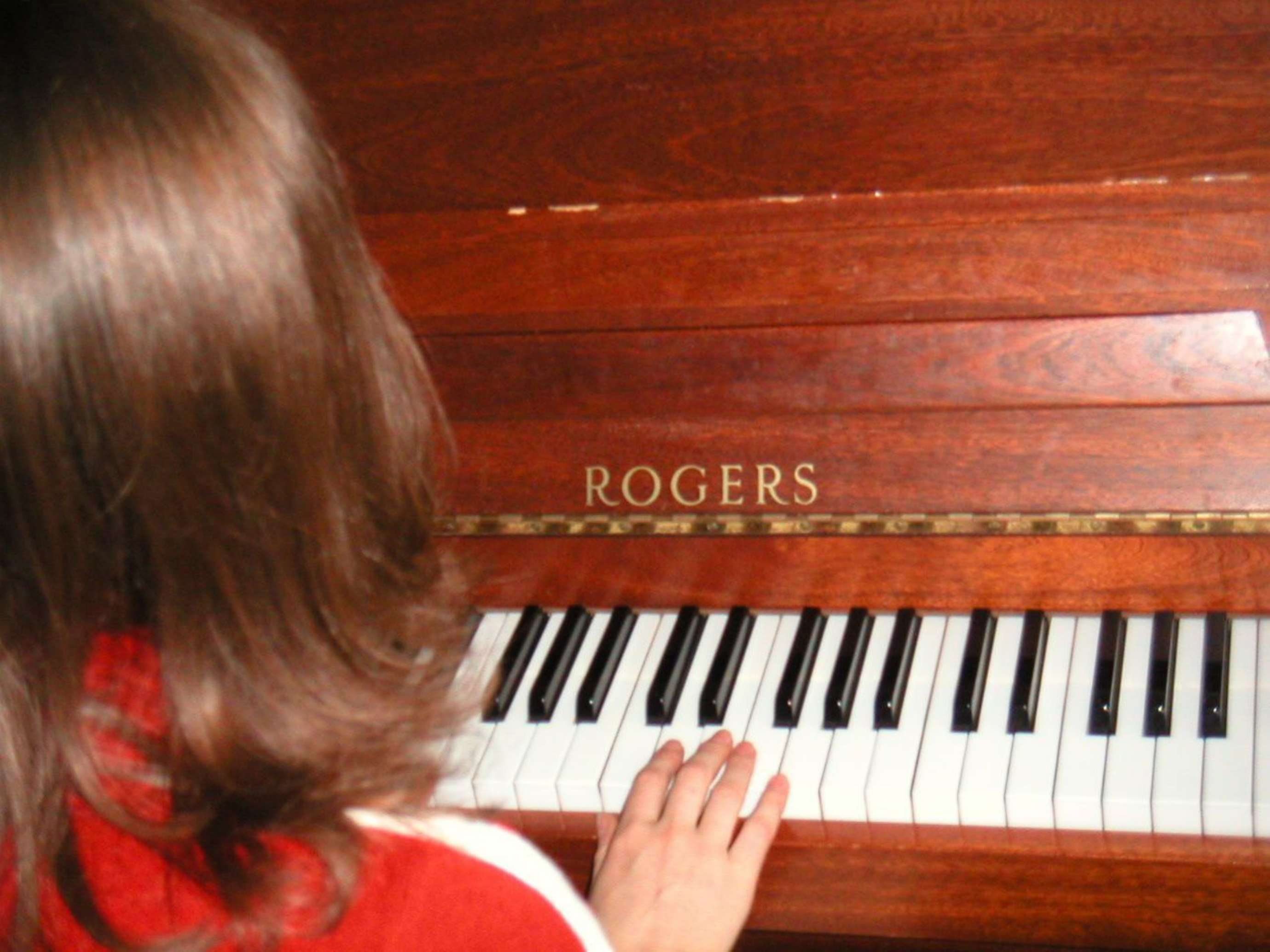}} 
& {\includegraphics[width=0.175\linewidth]{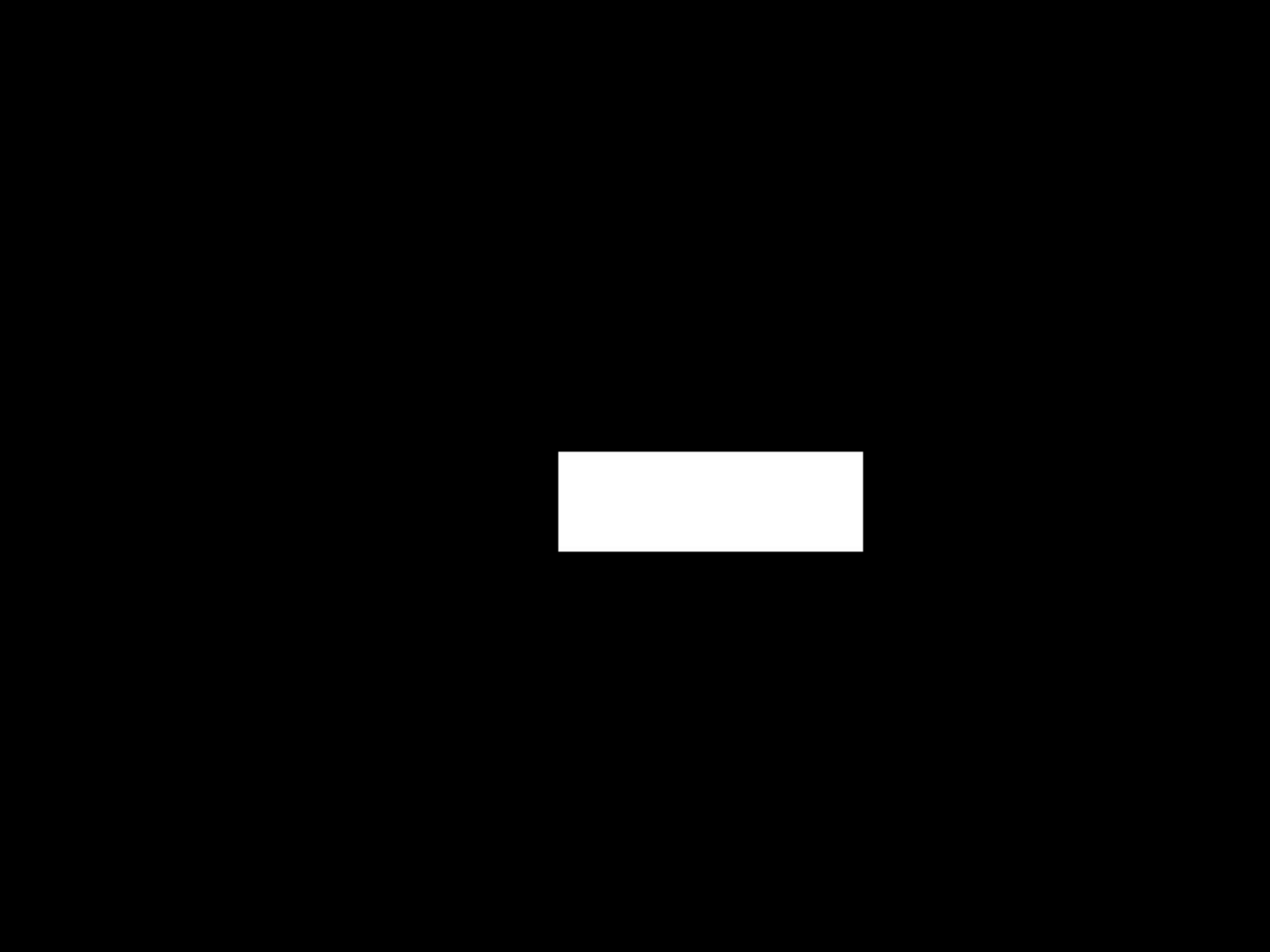}}
& {\includegraphics[width=0.175\linewidth]{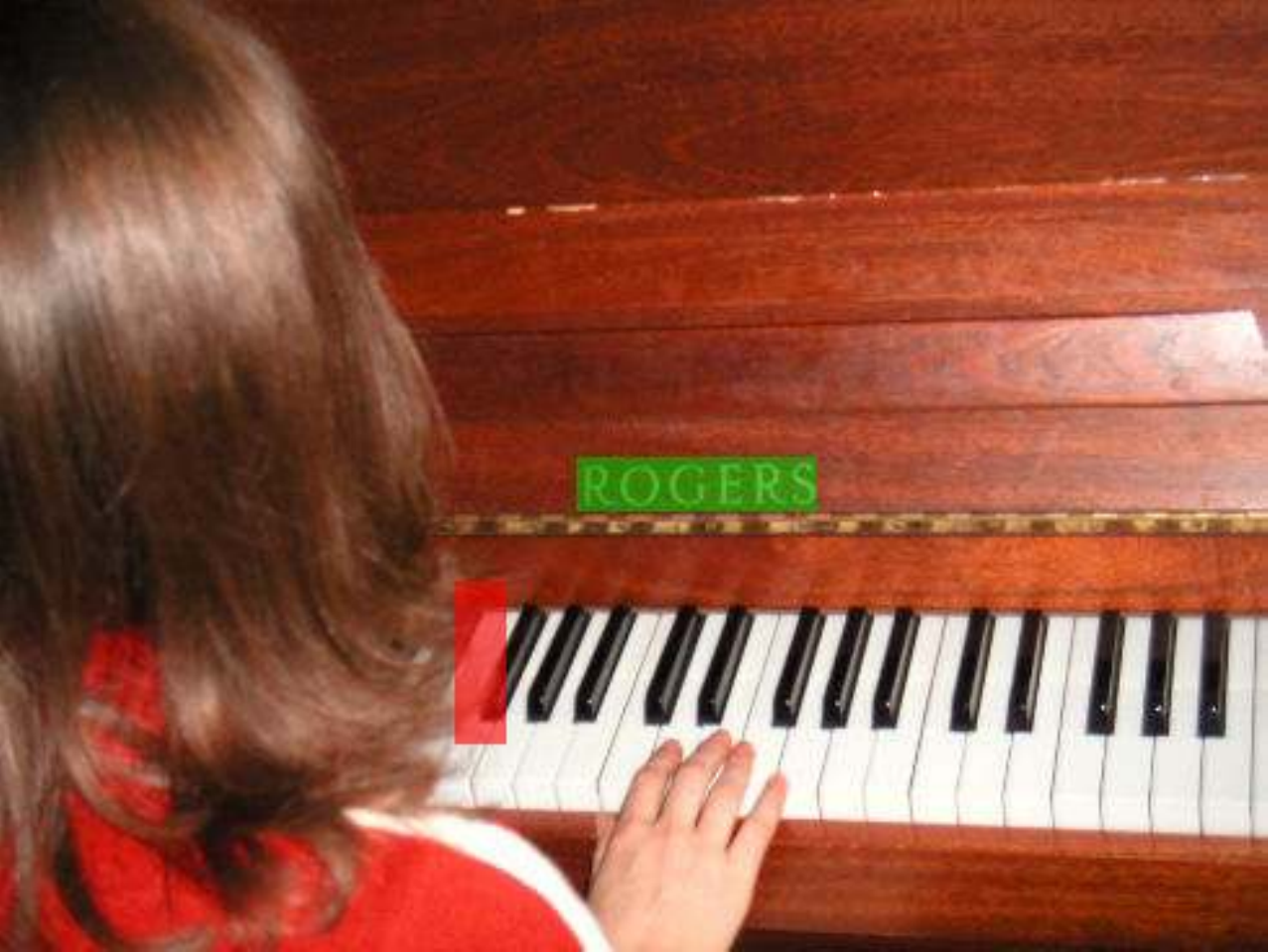}}
& {\includegraphics[width=0.175\linewidth]{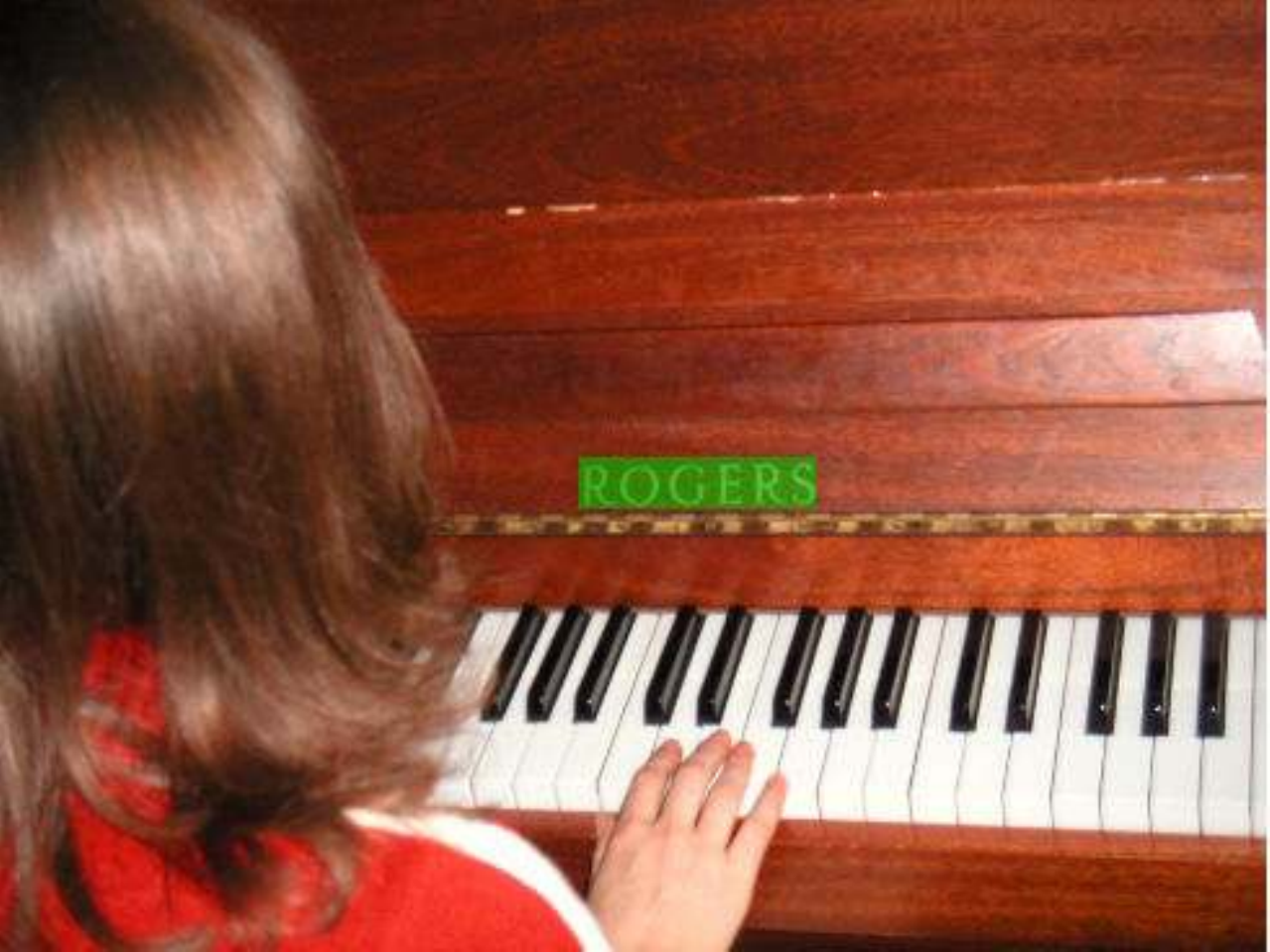}}
& {\includegraphics[width=0.175\linewidth]{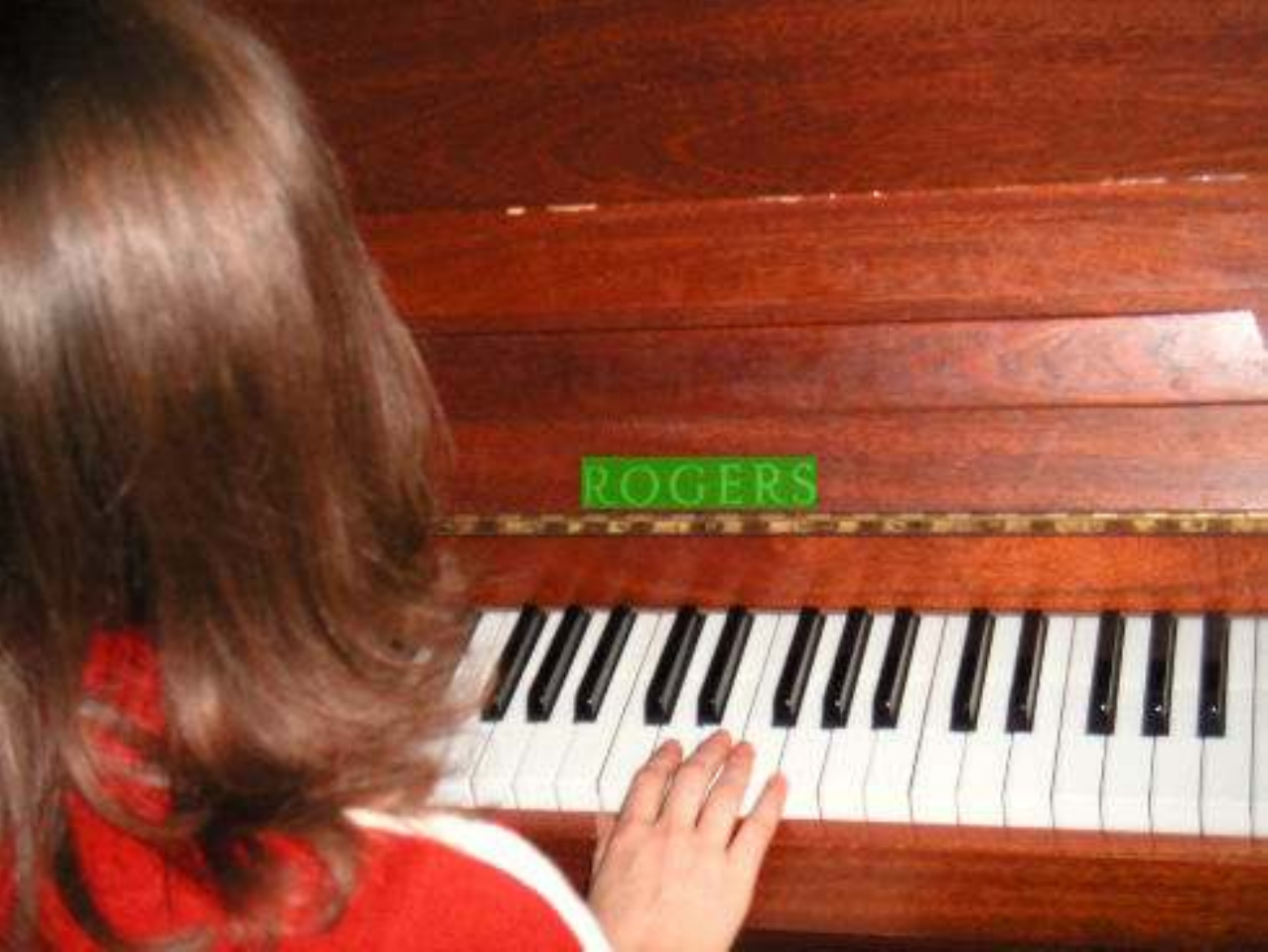}} \\
 (a) Input & (b) Predicted mask & (c) CTPN & (d) Guided CTPN & (e) Guided CTPN+
\end{tabular}
\end{minipage}
\begin{minipage}{1\linewidth}
\caption{Comparison of qualitative result on ICDAR 2013 using DetEval evaluation protocol, with Guided-CNN/Guided-CNN+ which instantiated with CTPN as backbone model. Over-merged, false alarm and true positive are highlighted in yellow, red and green, respectively. Best viewed in color.}
\label{fig:fig_1}
\end{minipage}
\end{figure}

\begin{figure}[htpb]
\begin{minipage}{1\linewidth}
\tabcolsep2.1pt
\renewcommand\arraystretch{1.3}
\begin{tabular}{ccccc}
{\includegraphics[width=0.175\linewidth]{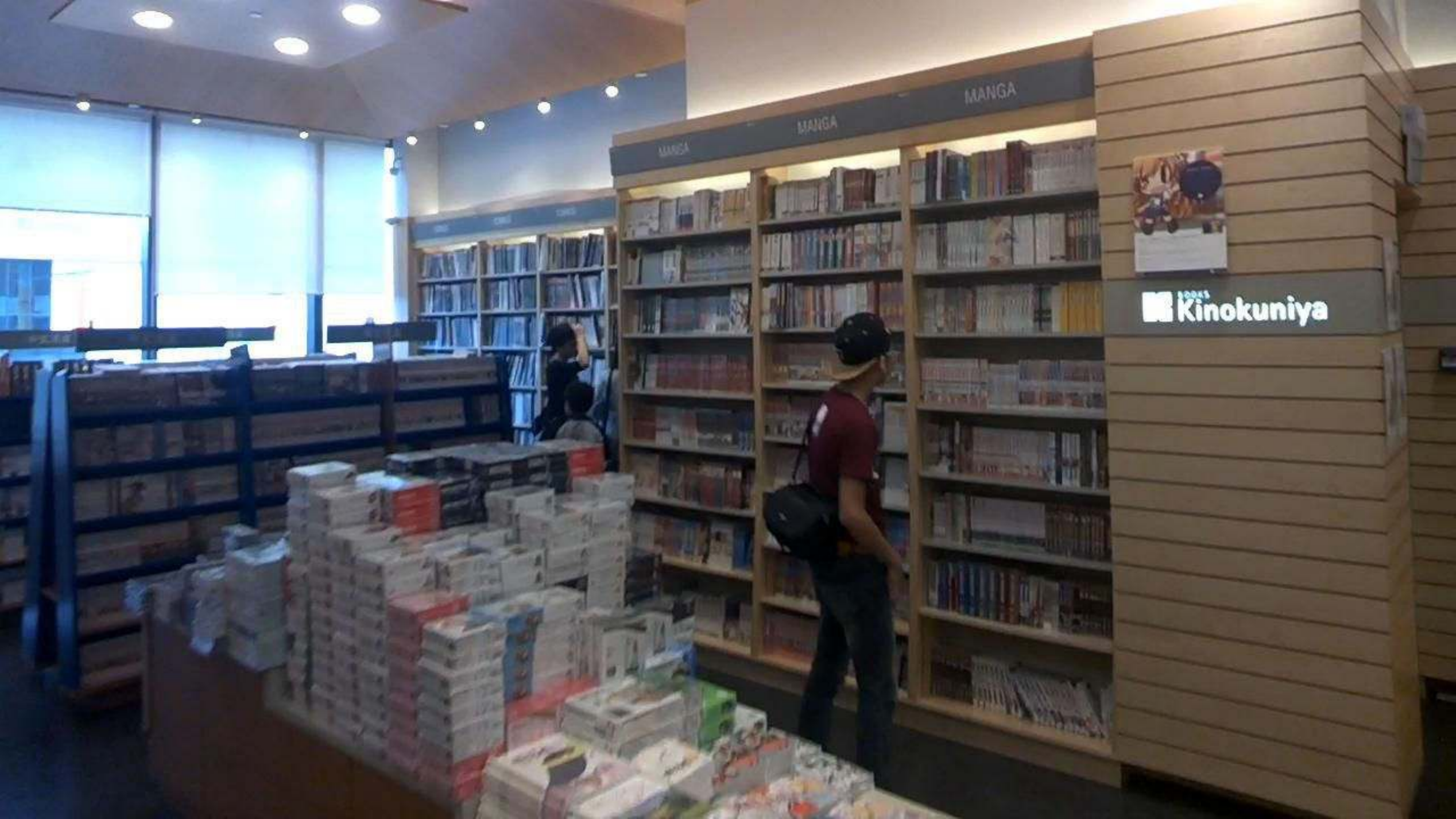}} 
& {\includegraphics[width=0.175\linewidth]{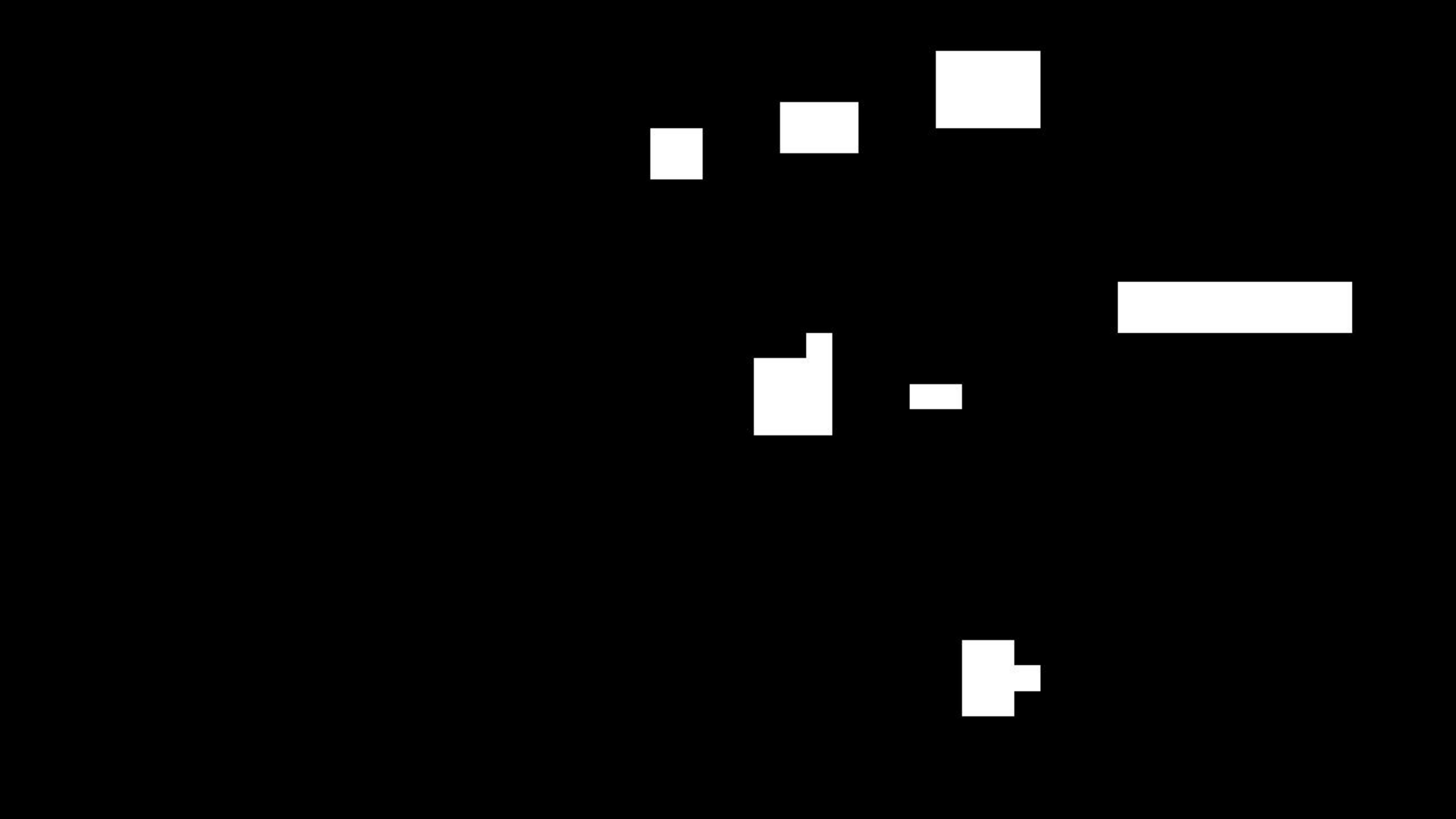}}
& {\includegraphics[width=0.175\linewidth]{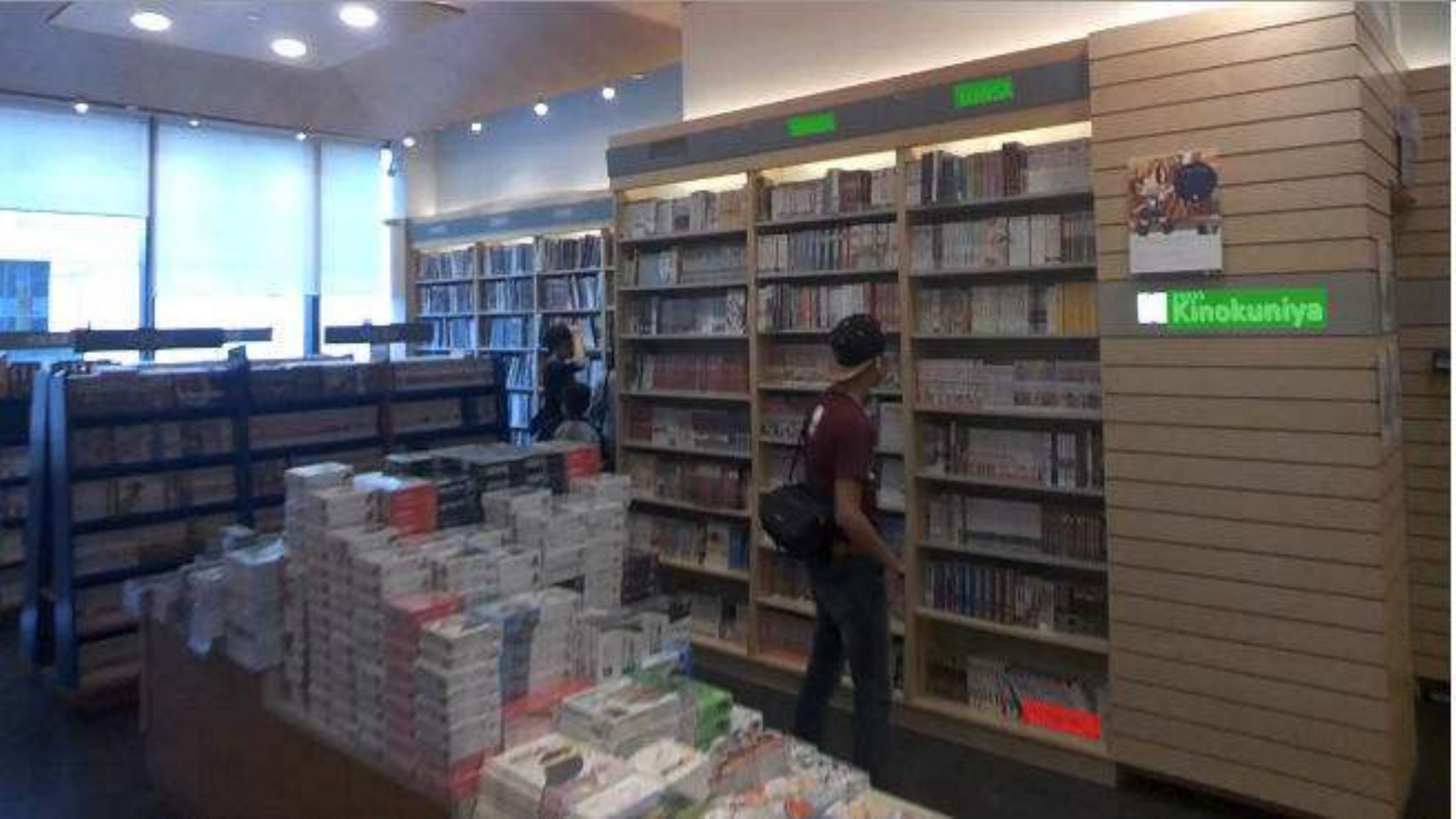}}
& {\includegraphics[width=0.175\linewidth]{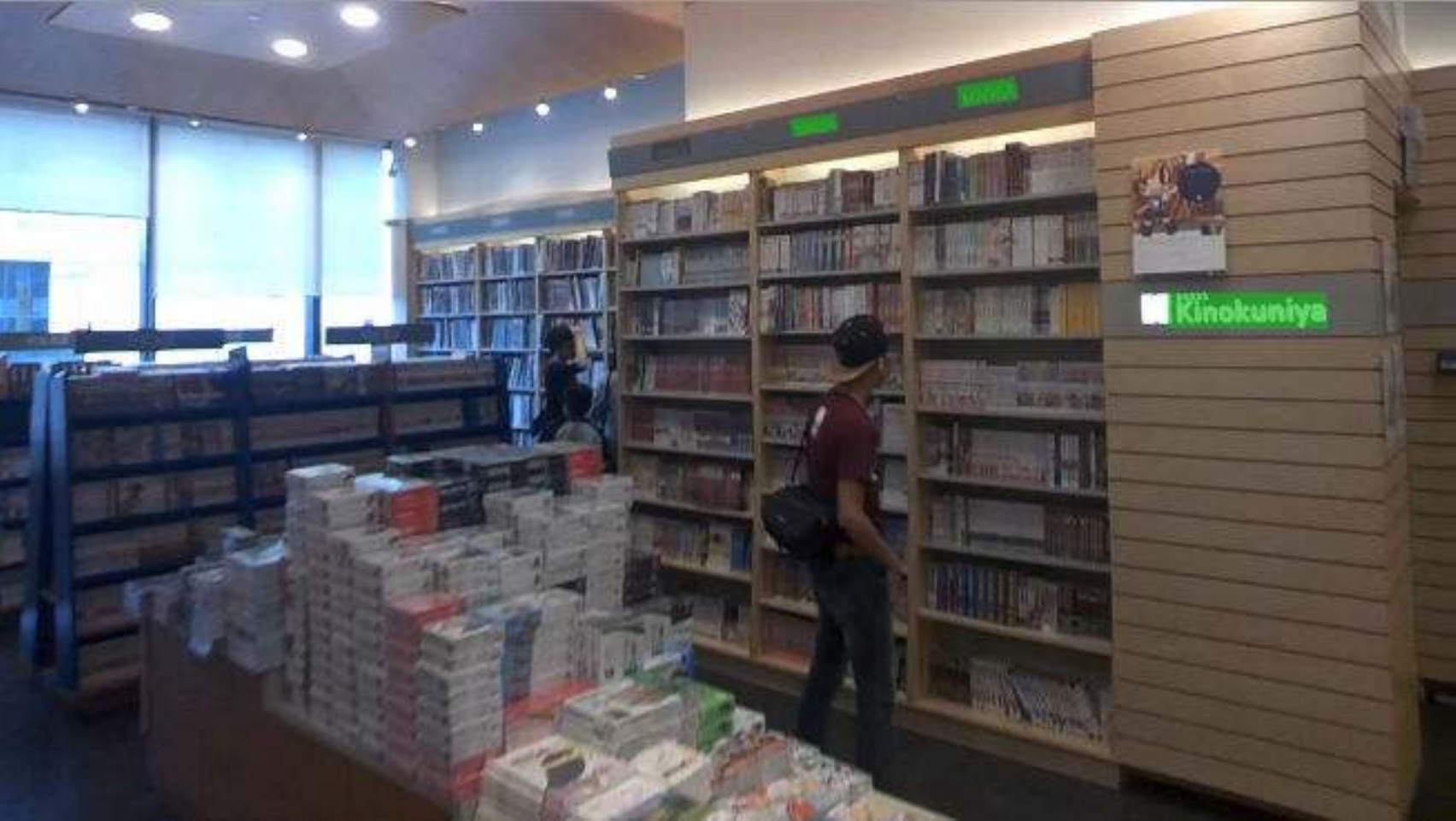}}
& {\includegraphics[width=0.175\linewidth]{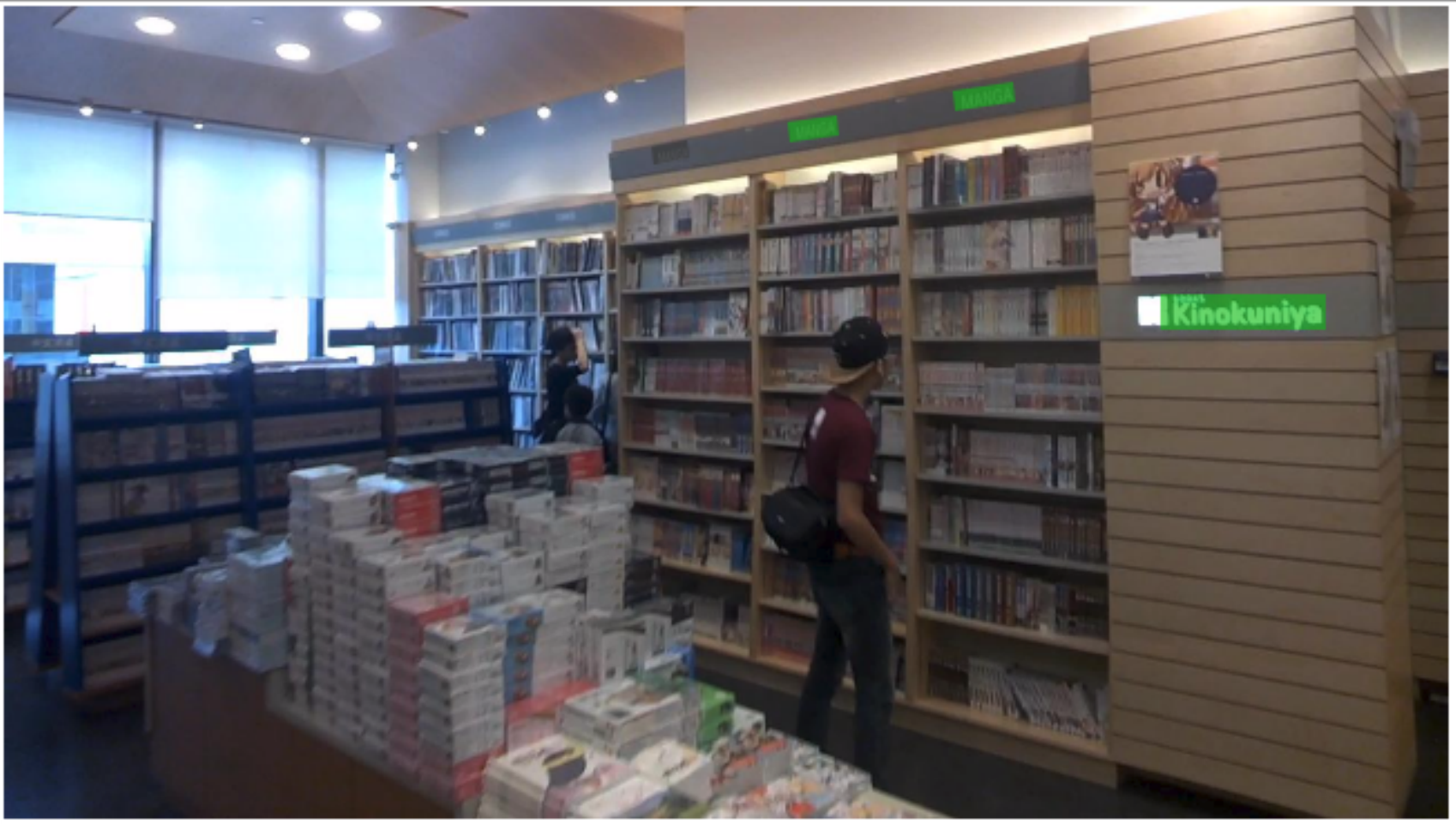}} \\
{\includegraphics[width=0.175\linewidth]{img_east/img_92.pdf}} 
& {\includegraphics[width=0.175\linewidth]{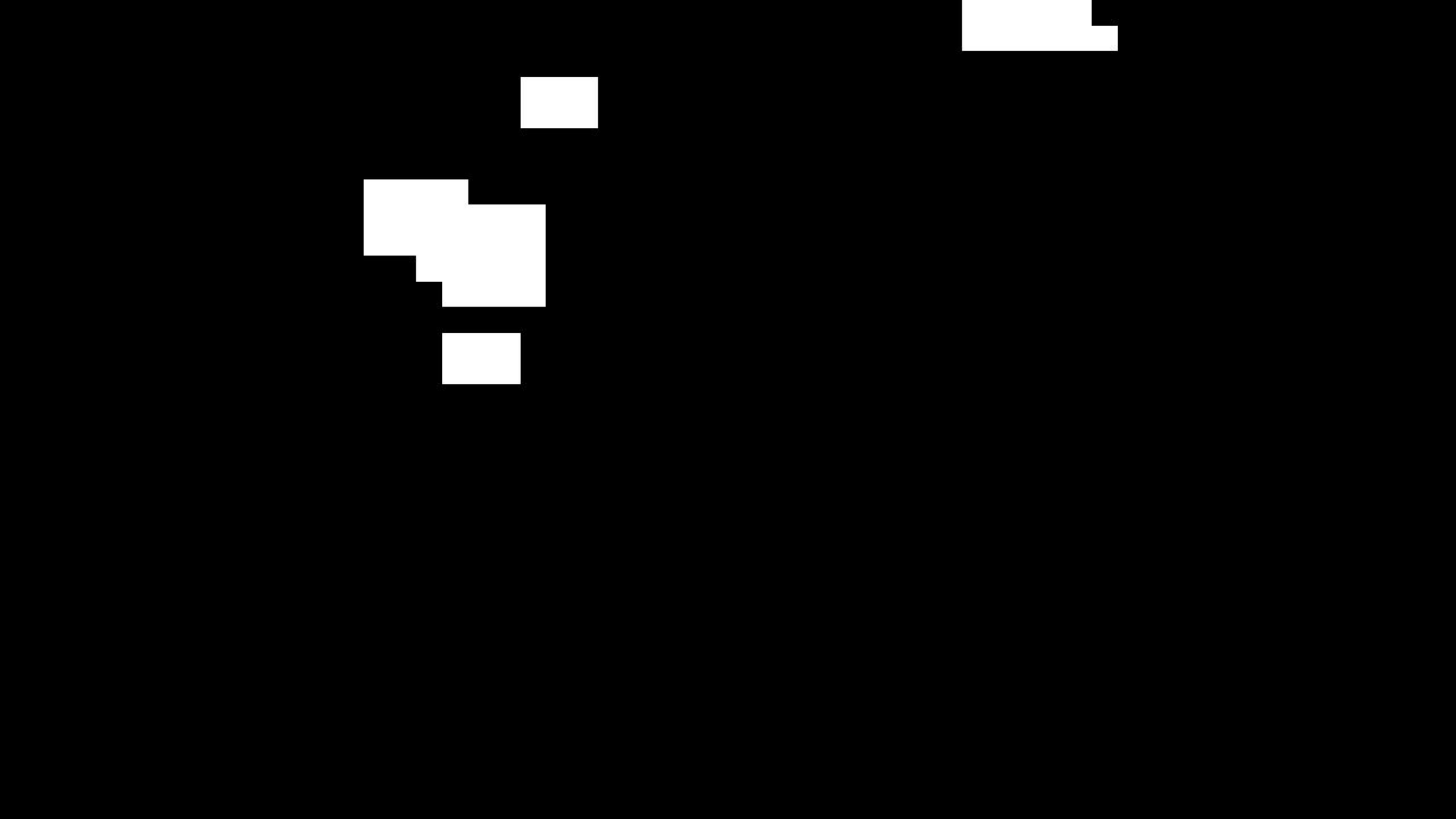}}
& {\includegraphics[width=0.175\linewidth]{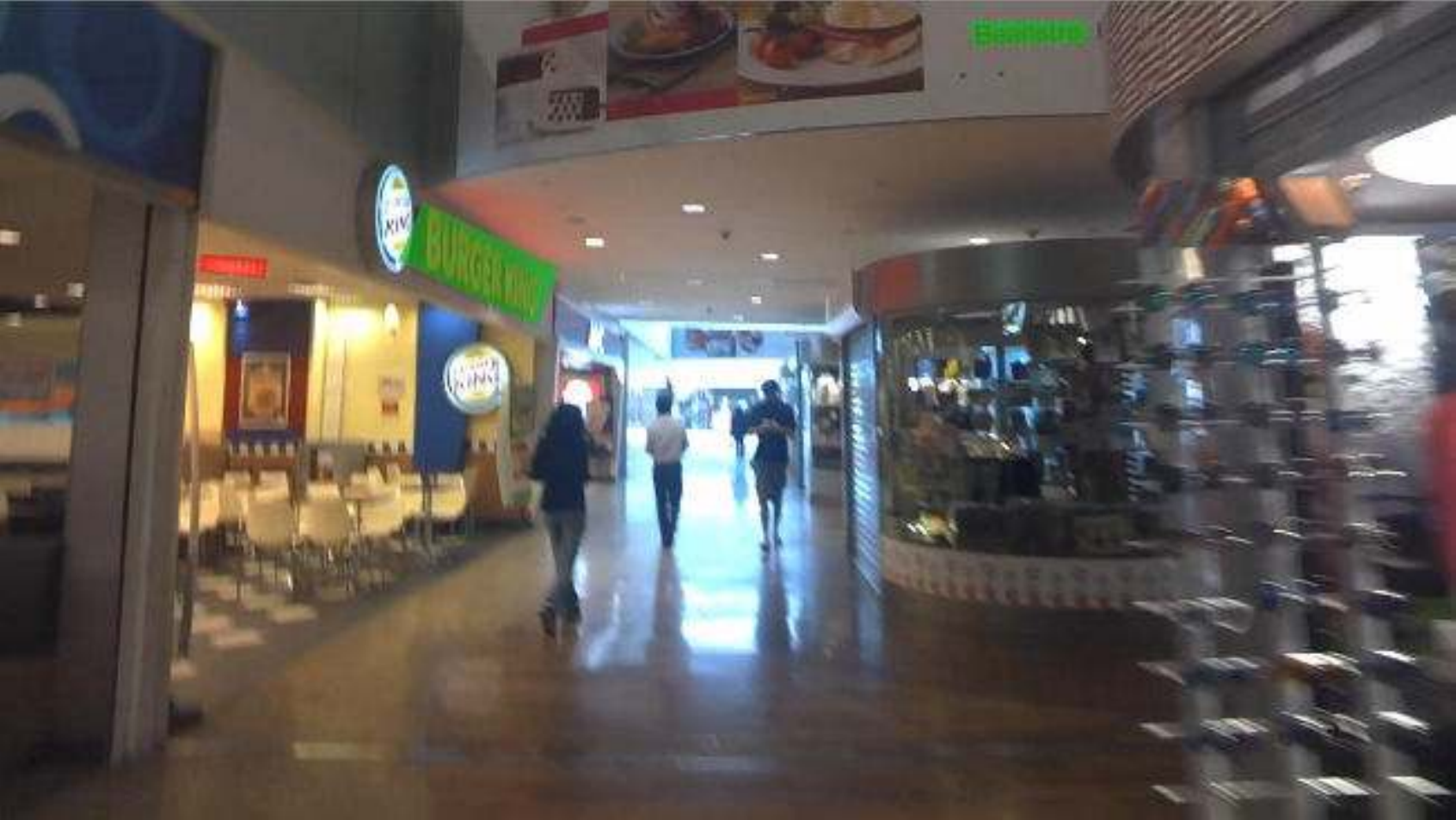}}
& {\includegraphics[width=0.175\linewidth]{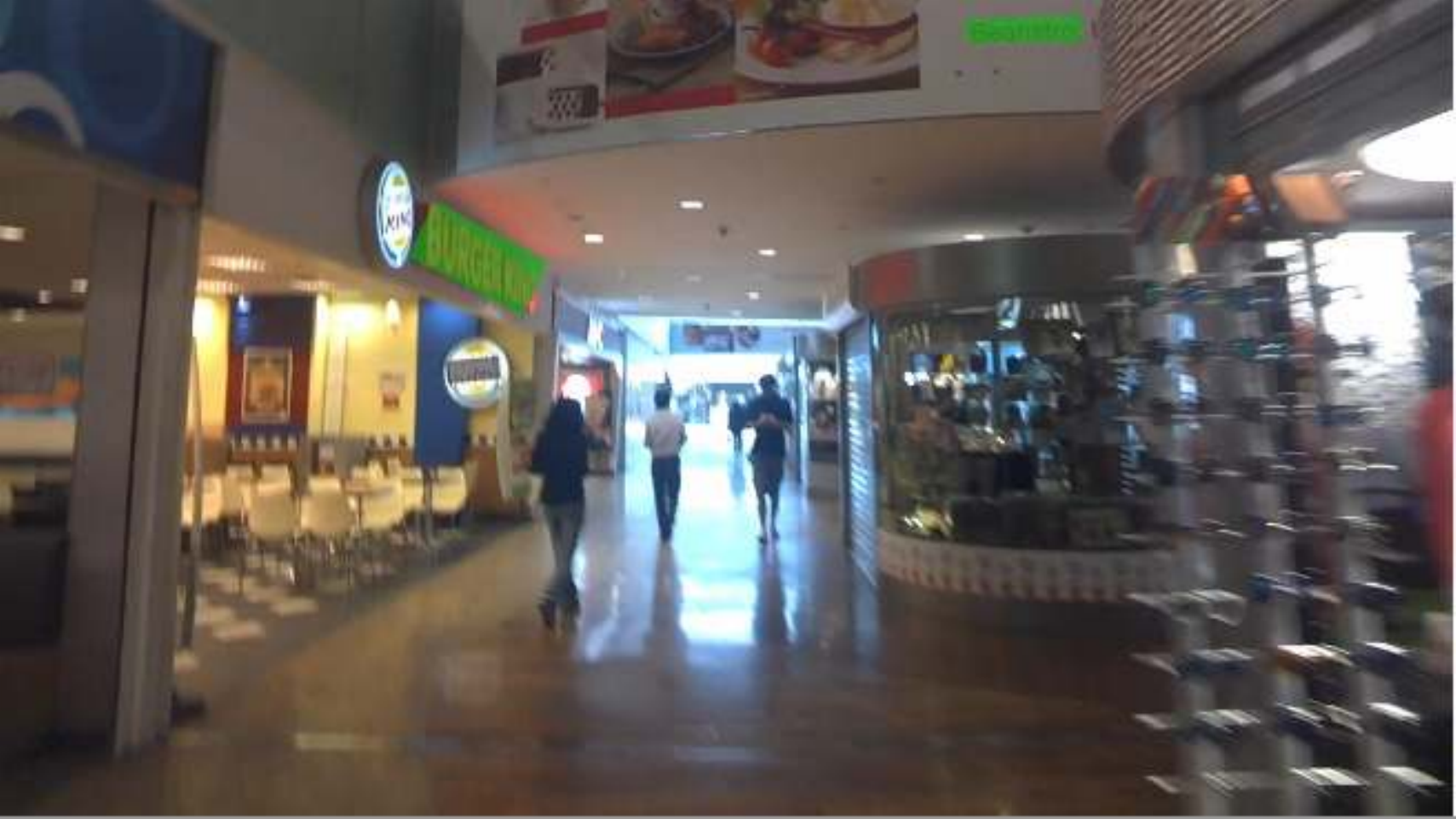}}
& {\includegraphics[width=0.175\linewidth]{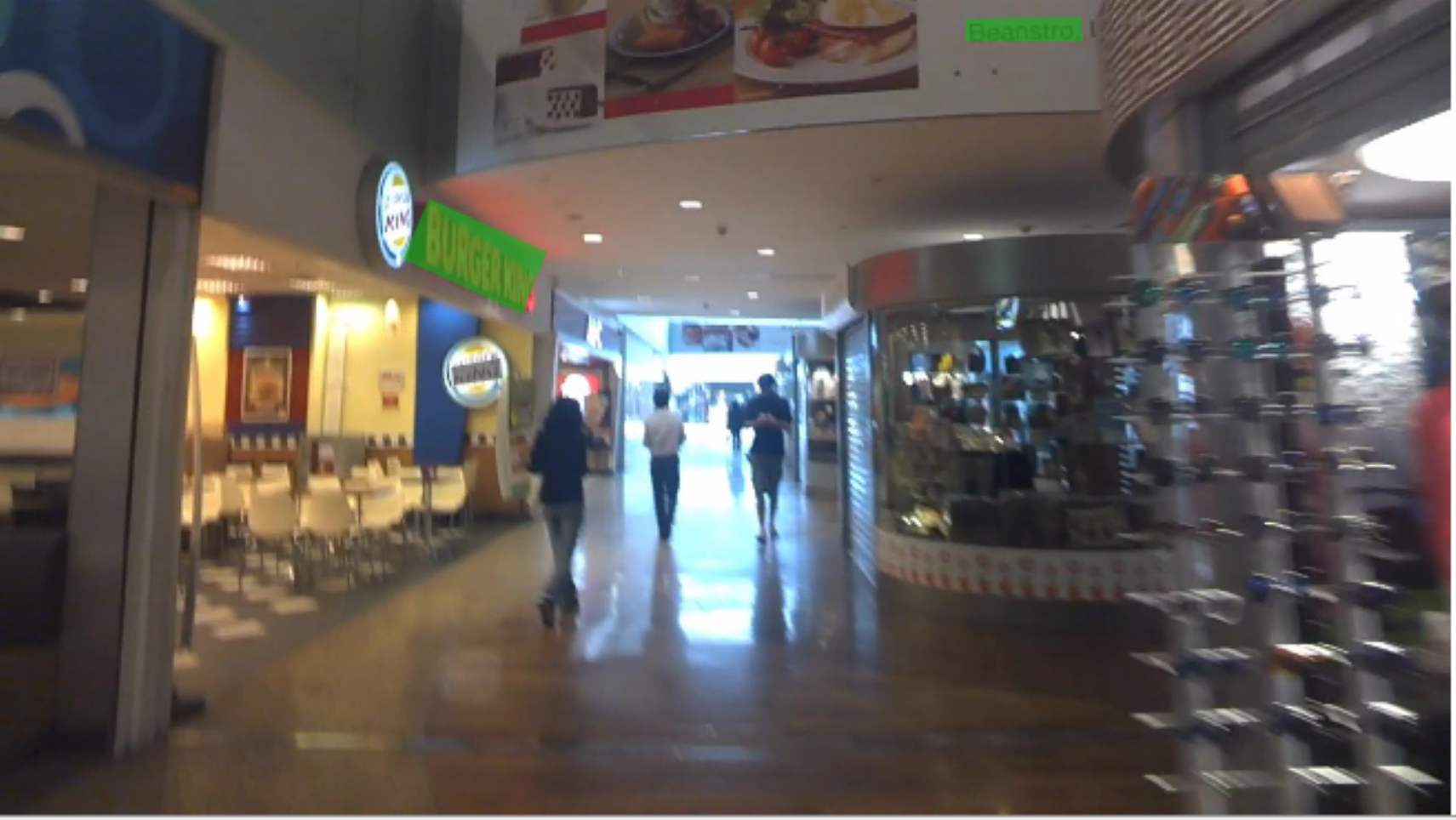}} \\
{\includegraphics[width=0.175\linewidth]{img_east/img_118.pdf}} 
& {\includegraphics[width=0.175\linewidth]{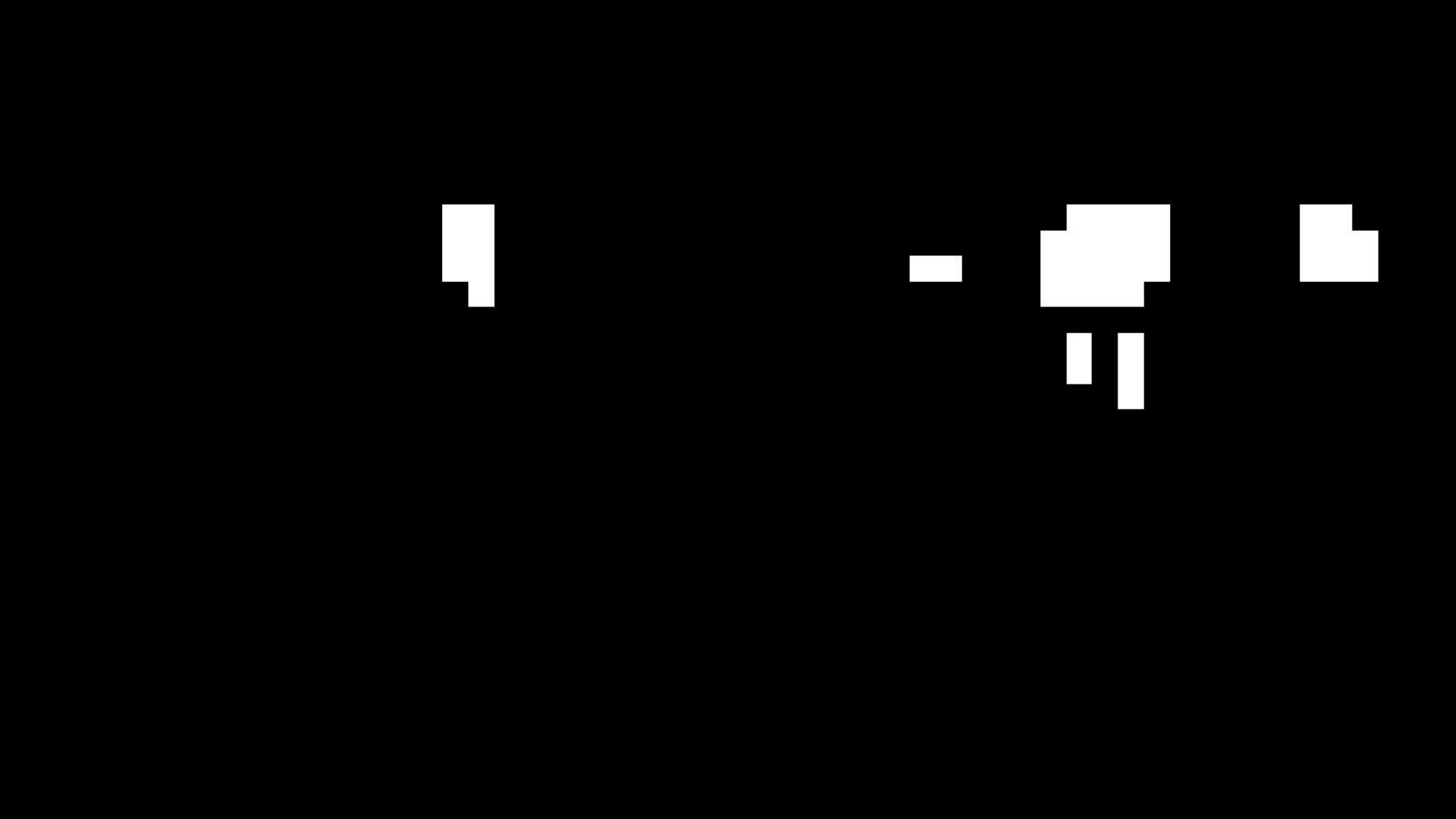}}
& {\includegraphics[width=0.175\linewidth]{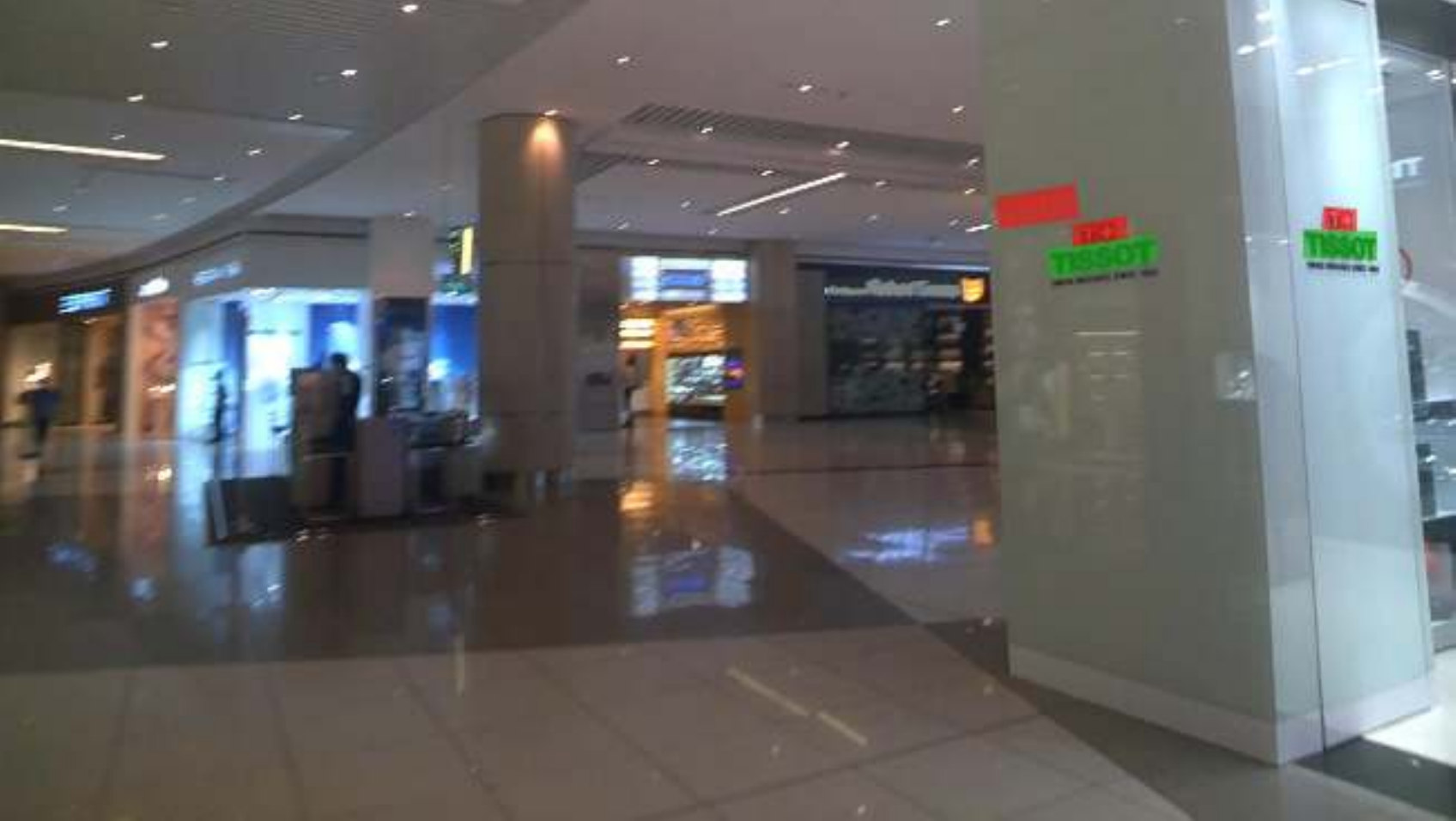}}
& {\includegraphics[width=0.175\linewidth]{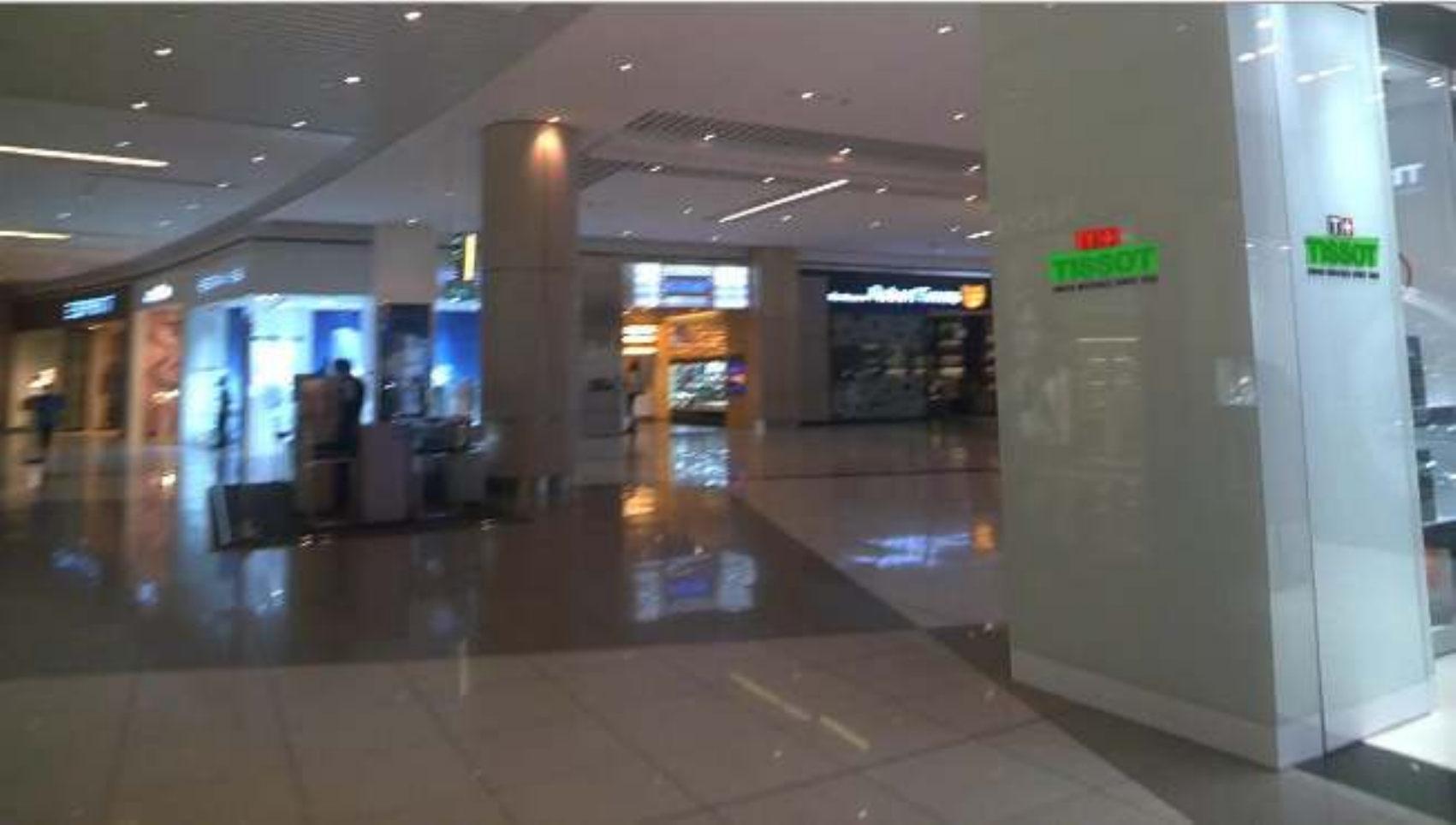}}
& {\includegraphics[width=0.175\linewidth]{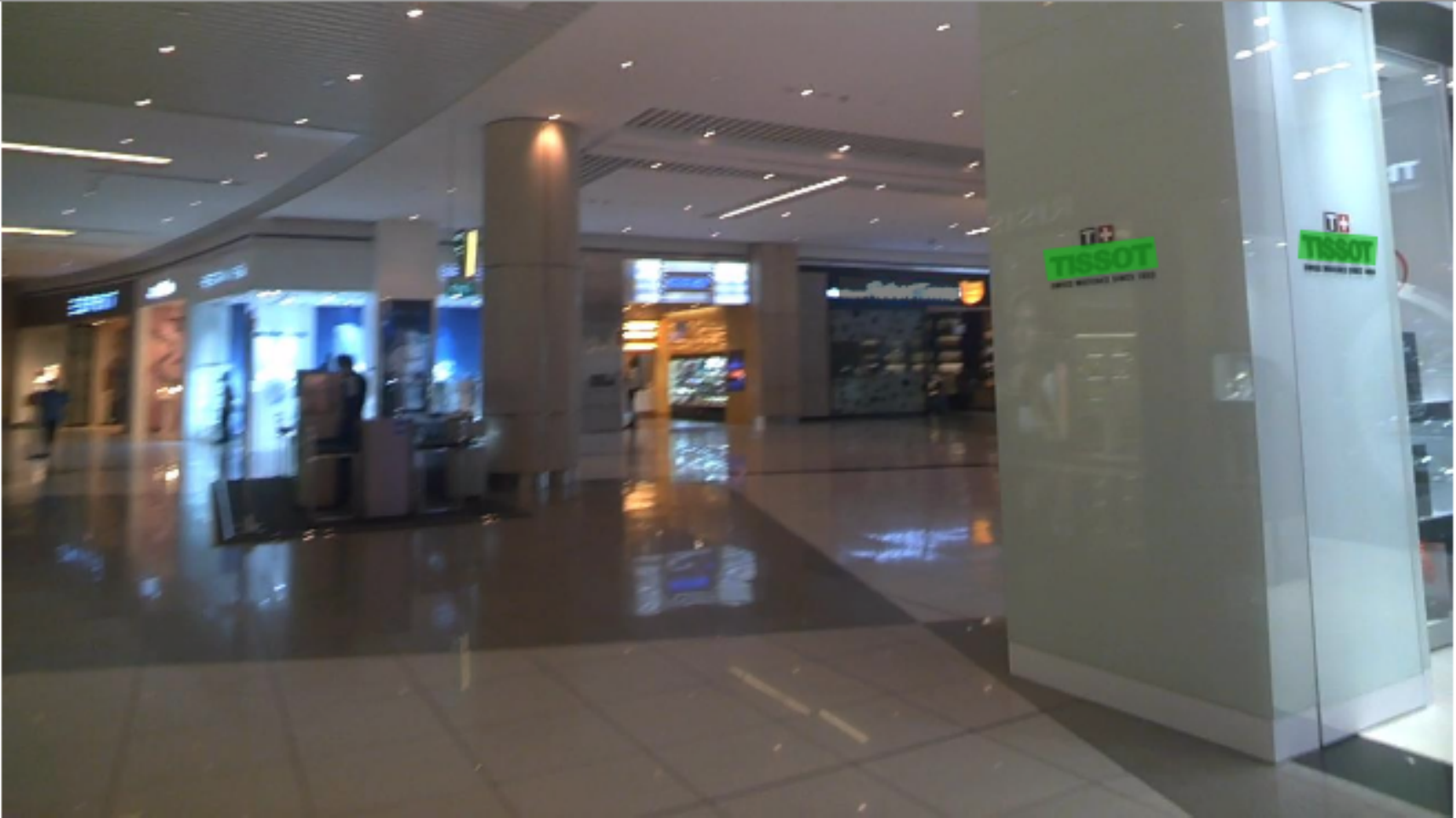}} \\
{\includegraphics[width=0.175\linewidth]{img_east/img_172.pdf}} 
& {\includegraphics[width=0.175\linewidth]{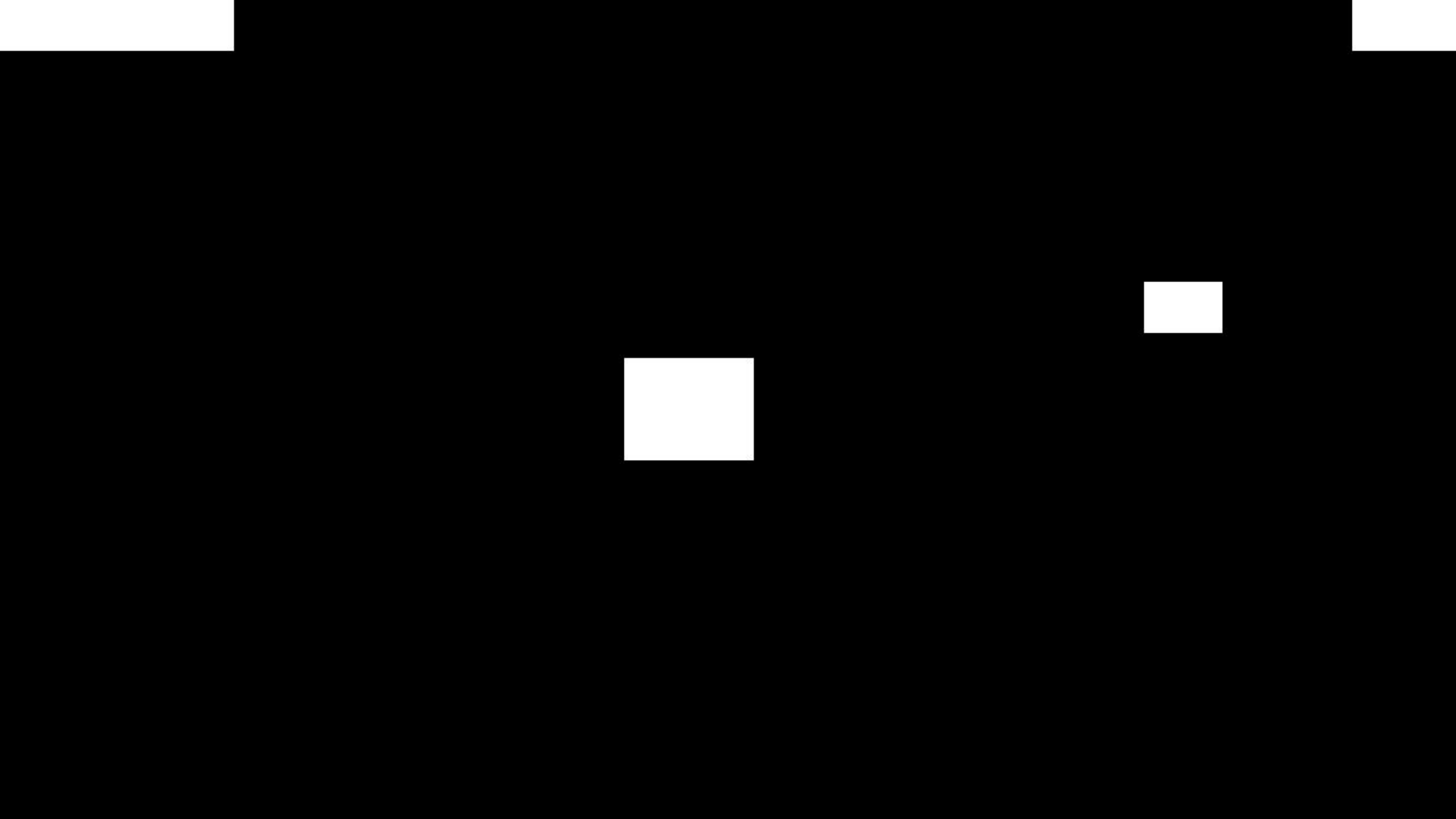}}
& {\includegraphics[width=0.175\linewidth]{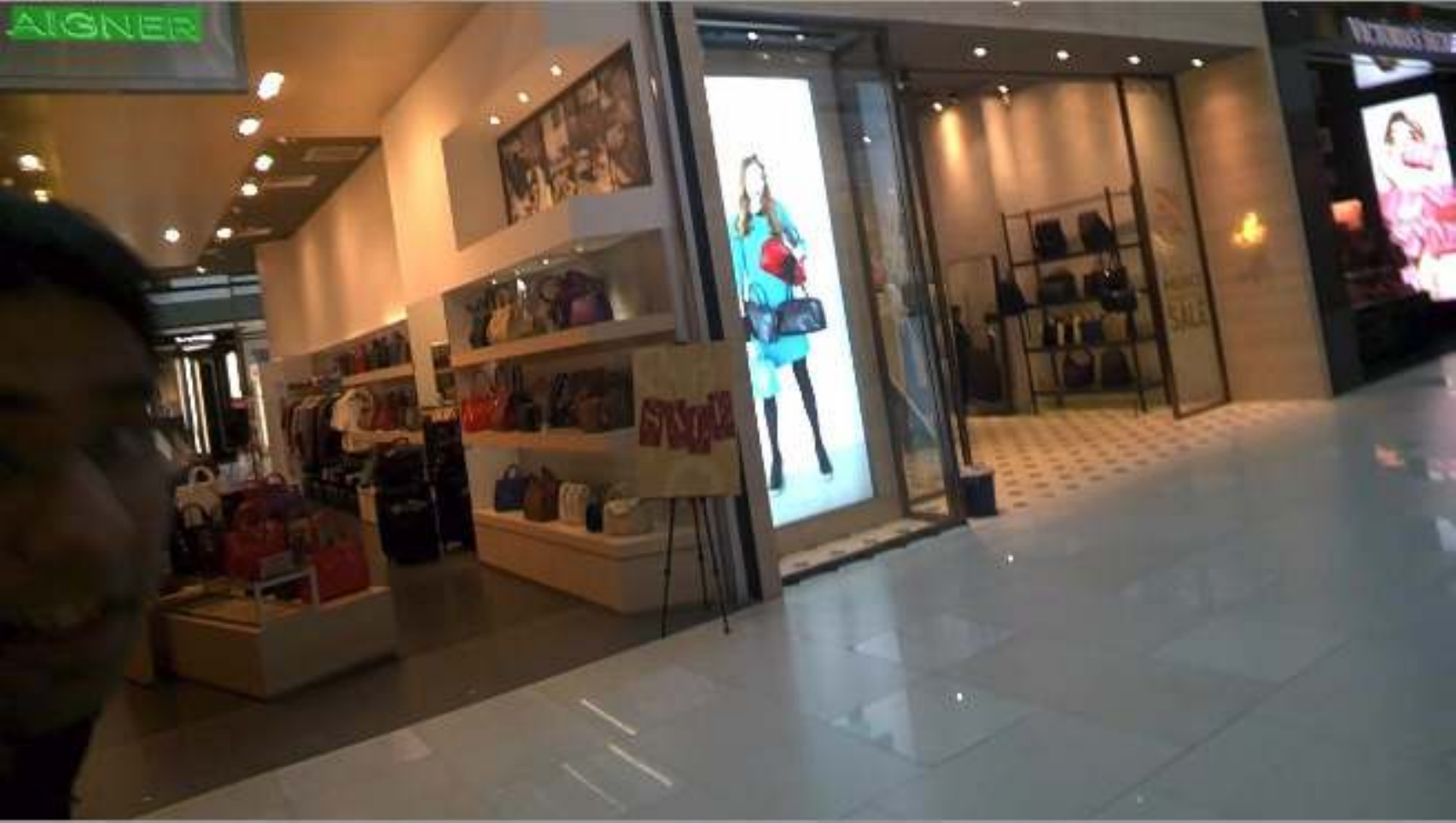}}
& {\includegraphics[width=0.175\linewidth]{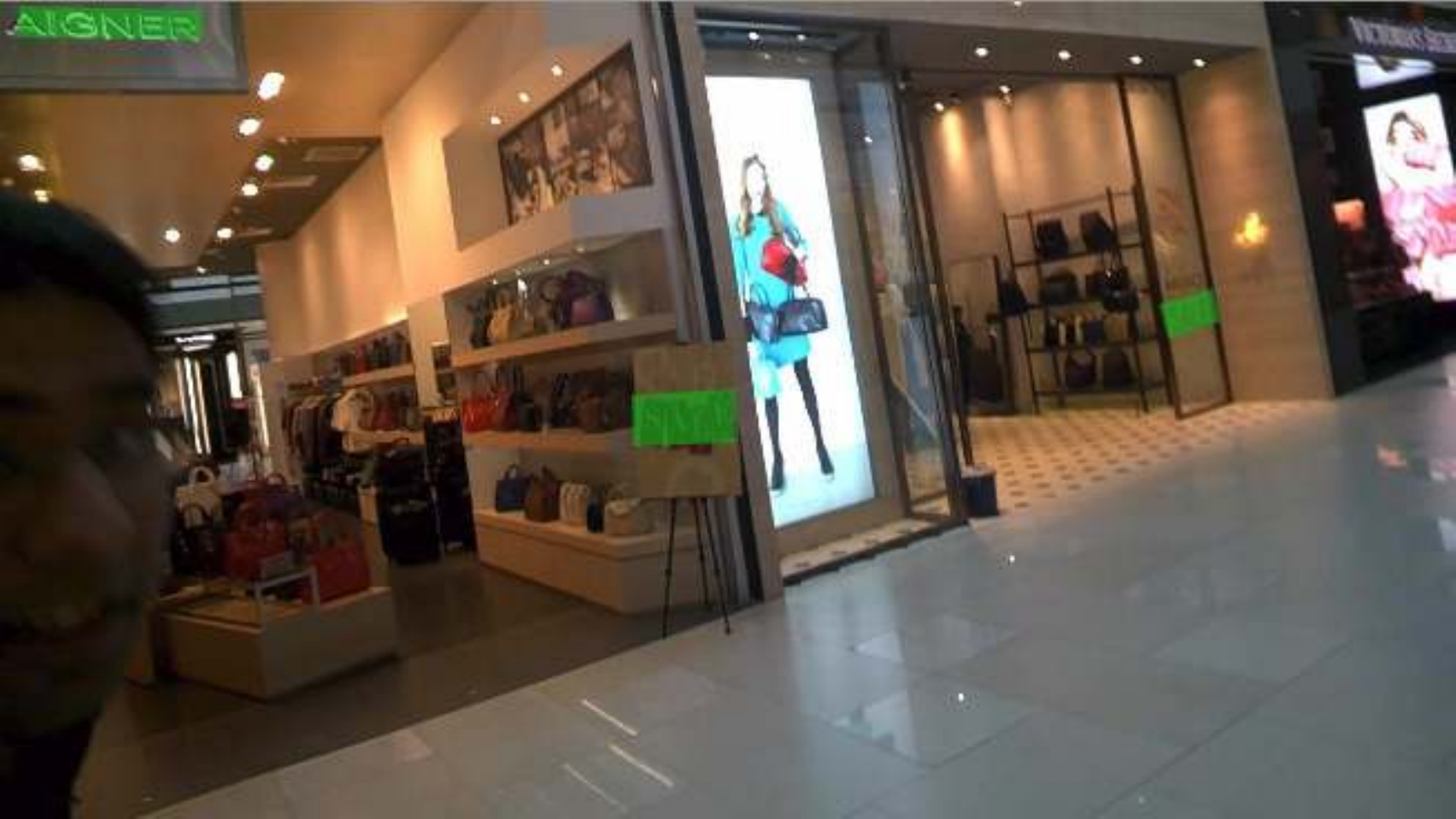}}
& {\includegraphics[width=0.175\linewidth]{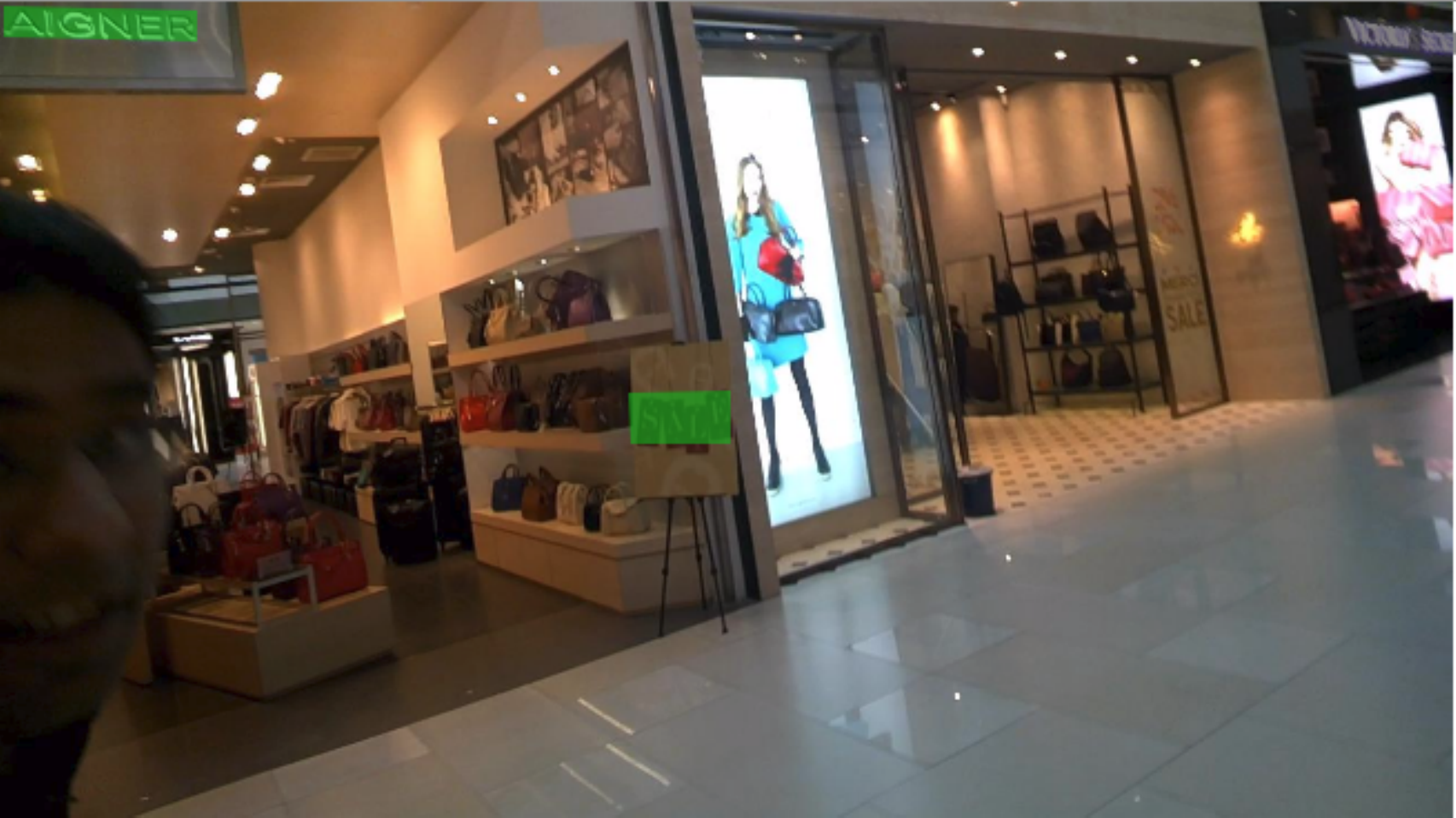}} \\
{\includegraphics[width=0.175\linewidth]{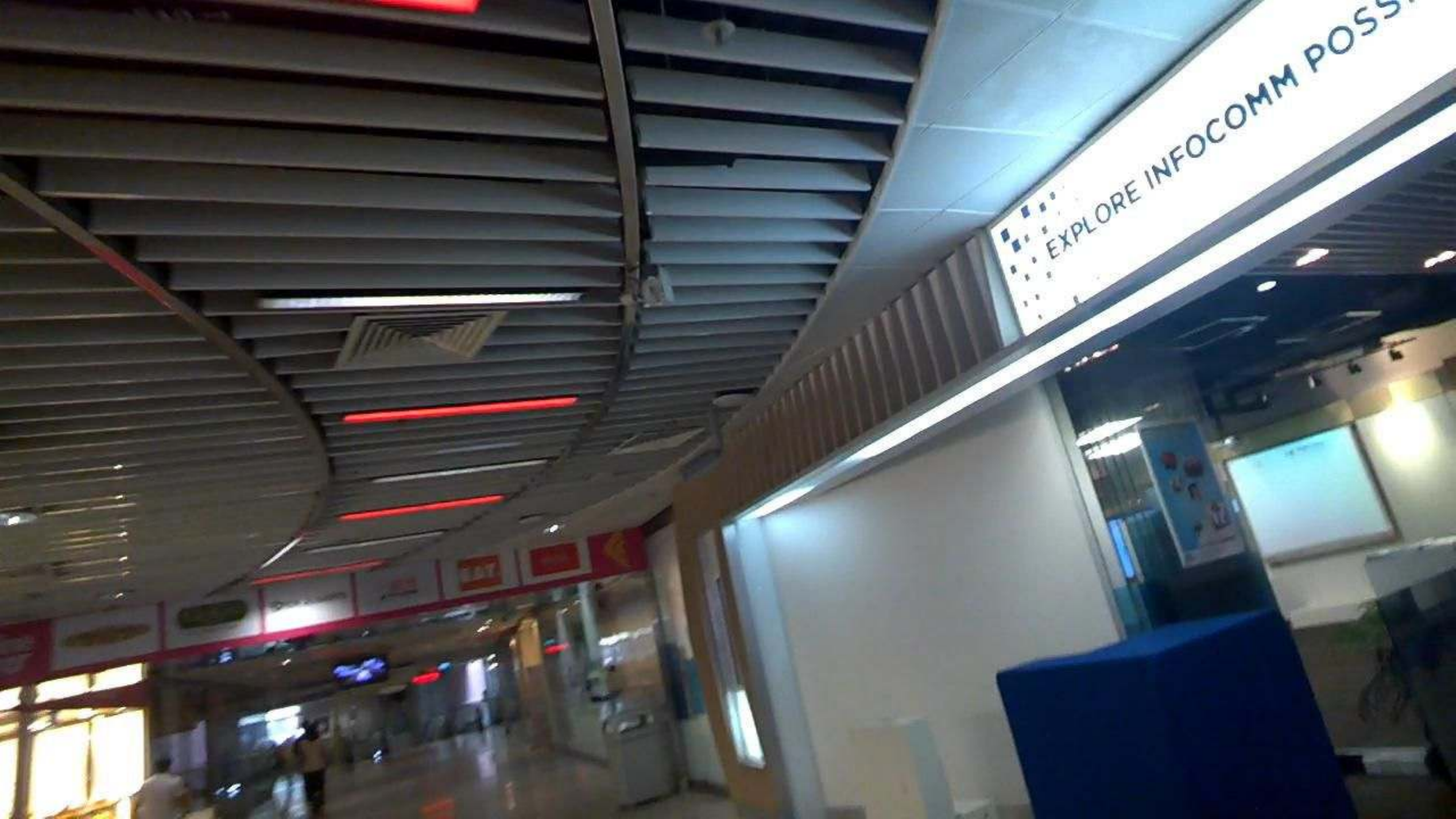}} 
& {\includegraphics[width=0.175\linewidth]{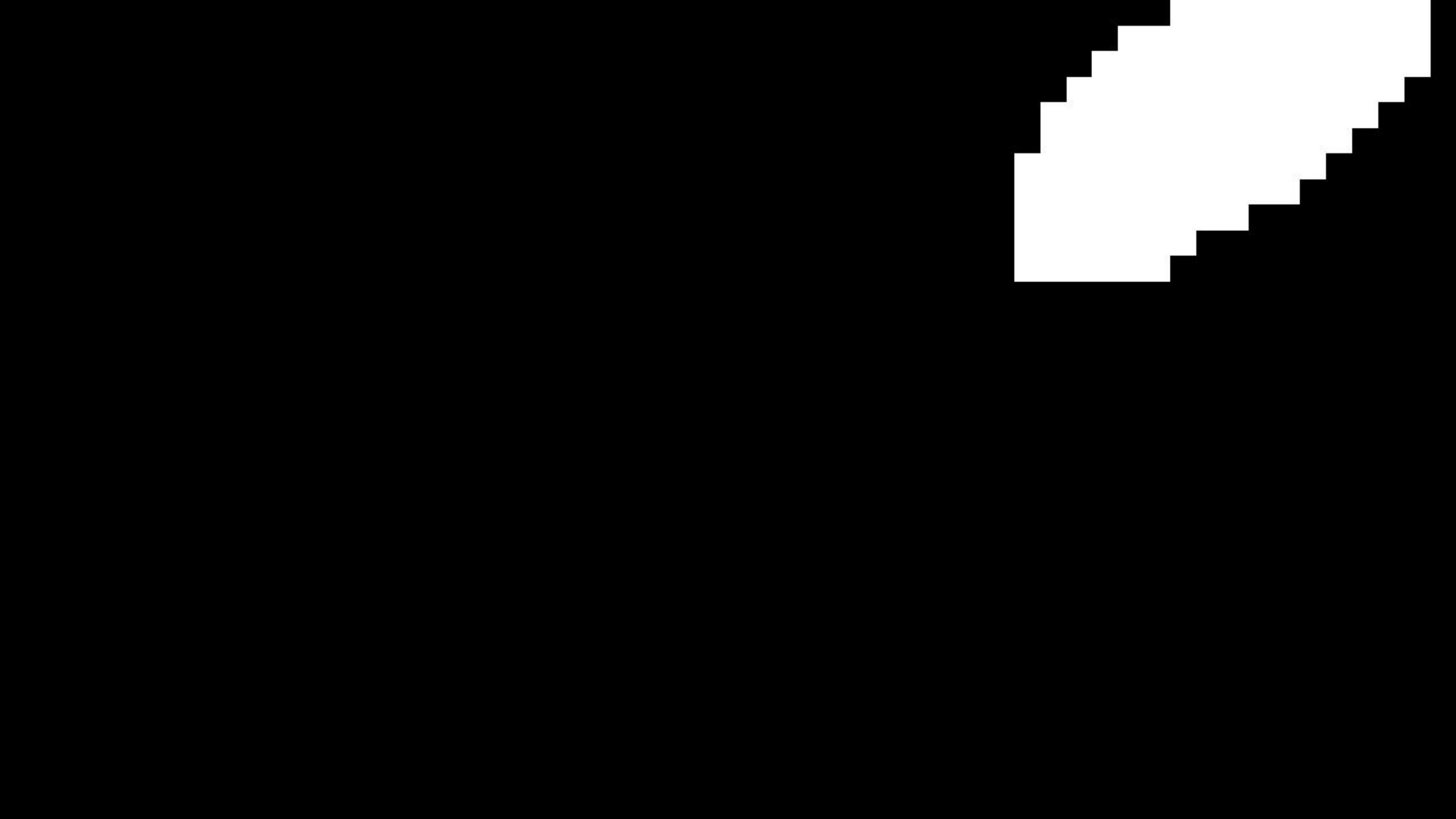}}
& {\includegraphics[width=0.175\linewidth]{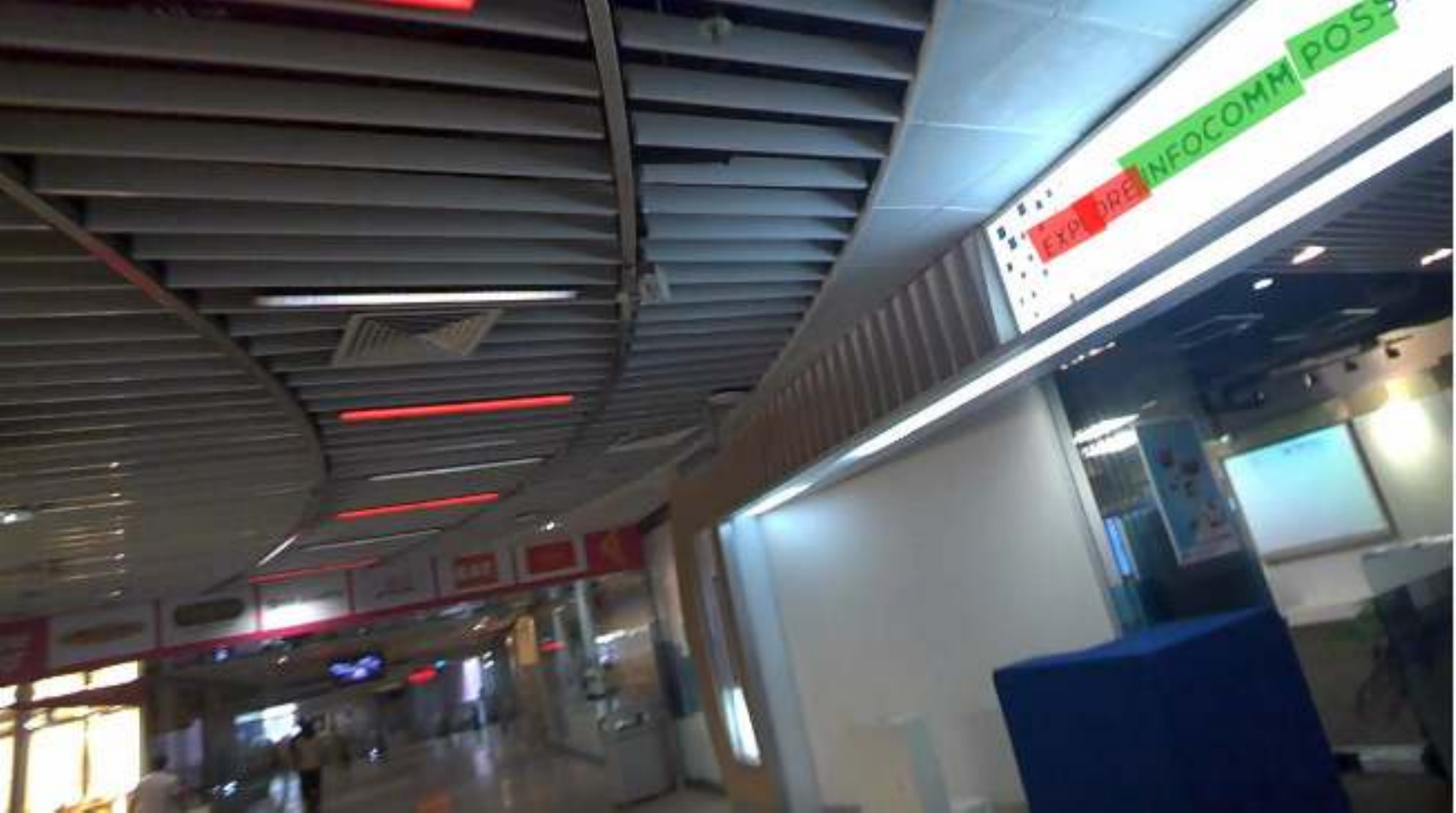}}
& {\includegraphics[width=0.175\linewidth]{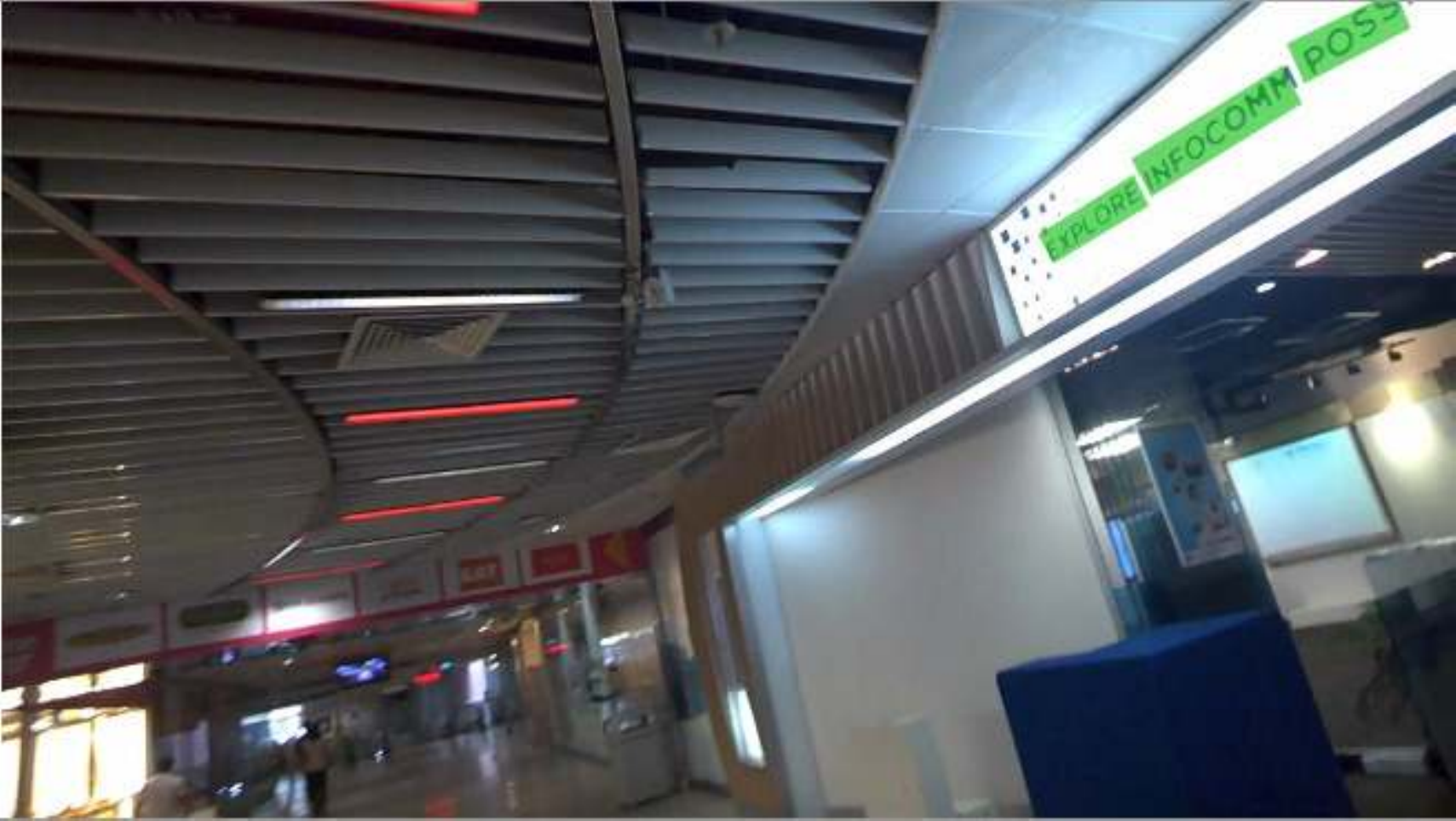}}
& {\includegraphics[width=0.175\linewidth]{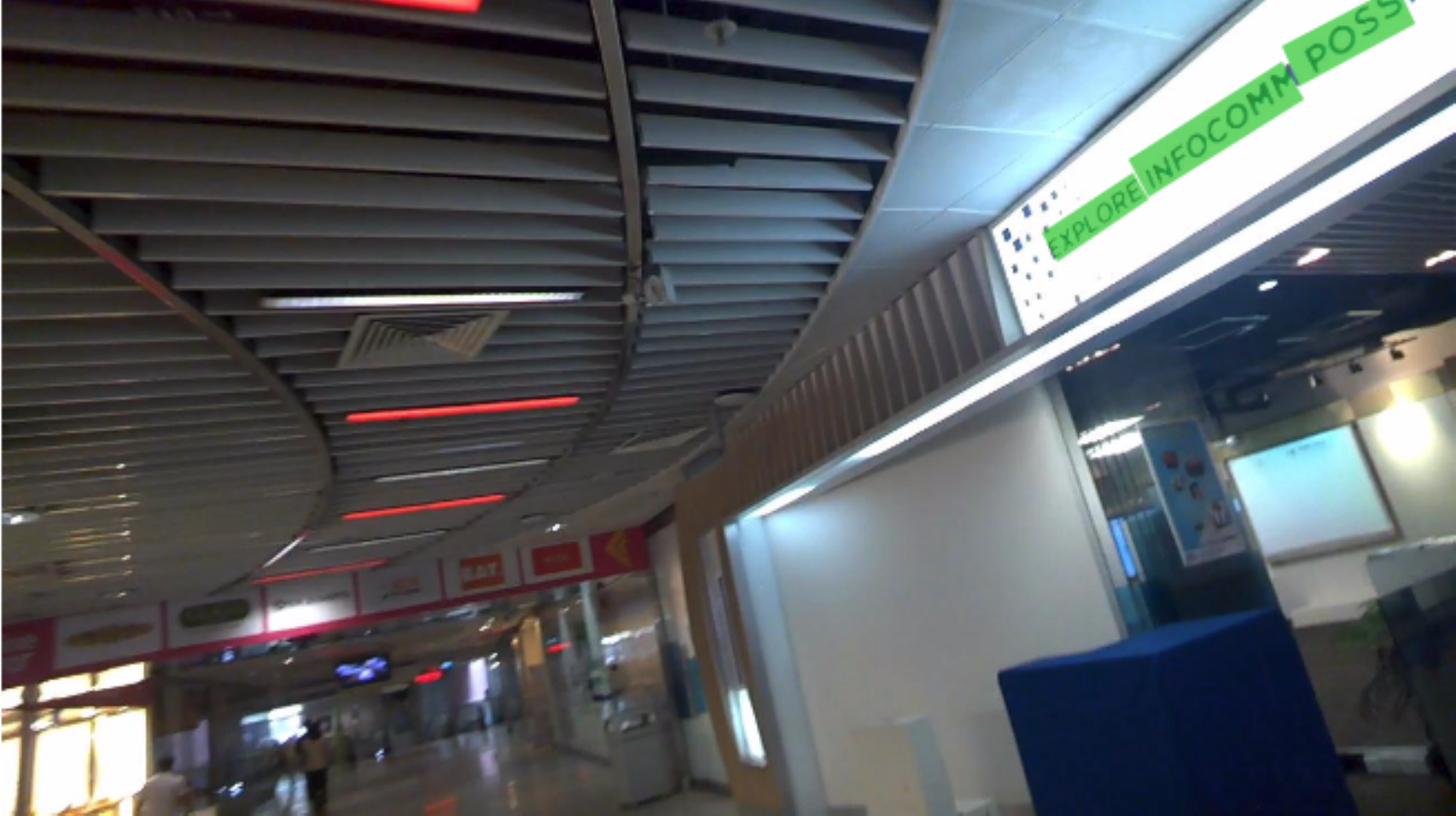}} \\
 (a) Input & (b) Predicted mask & (c) EAST & (d) Guided EAST & (e) Guided EAST+
\end{tabular}
\end{minipage}
\begin{minipage}{1\linewidth}
\caption{Comparison of qualitative result on ICDAR 2015 with Guided-CNN/Guided-CNN+ which instantiated with EAST as backbone model. false alarm and true positive are highlighted in red and green, respectively. Best viewed in color.}
\label{fig:fig_2}
\end{minipage}
\end{figure}

Several detection examples with CTPN under various challenging cases on the ICDAR 2013 dataset are shown in Figure \ref{fig:fig_1} (c). Masks predicted by our proposed guidance subnetwork are shown in Figure \ref{fig:fig_1} (b), which can filter out most backgrounds, speeding up text detector as well as improving their accuracy simultaneously. Figure \ref{fig:fig_1} (d) shows detection results predicted by Guided CTPN, and Figure \ref{fig:fig_1} (e) shows results of Guided CTPN+. Comparing the original CTPN with our Guided CTPN and Guided CTPN+ in Figure \ref{fig:fig_1}, we can see that we produce less false alarms because of the effectiveness of the proposed guidance subnetwork. More similar examples in ICDAR 2015 can be seen in Figure \ref{fig:fig_2}. 

\end{document}